\documentclass[AMA,STIX1COL]{WileyNJD-v2}
\usepackage{cleveref,rviewport,booktabs,subcaption}
\articletype{Research Article}%

\graphicspath{{../}}

\received{7 October 2022}
\revised{17 January 2023}
\accepted{19 January 2023}

\usepackage[final]{changes}

\begin{document}

\title{Probabilistic partition of unity networks for high-dimensional regression problems\protect 
}

\author[1]{Tiffany Fan*}

\author[2]{Nathaniel Trask}

\author[3]{Marta D'Elia}

\author[1,4]{Eric Darve}

\authormark{FAN \textsc{et al}}

\address[1]{\orgdiv{Institute for Computational and Mathematical Engineering}, \orgname{Stanford University}, \orgaddress{\state{California}, \country{USA}}}

\address[2]{\orgdiv{Center for Computing Research}, \orgname{Sandia National Laboratories}, \orgaddress{\state{New Mexico}, \country{USA}}}

\address[3]{\orgdiv{Data Science and Computing}, \orgname{Sandia National Laboratories}, \orgaddress{\state{California}, \country{USA}}}

\address[4]{\orgdiv{Department of Mechanical Engineering}, \orgname{Stanford University}, \orgaddress{\state{California}, \country{USA}}}

\corres{*Tiffany Fan, 475 Via Ortega, Jen-Hsun Huang Engineering Center, Stanford, CA 94305, USA. \email{tiffan@stanford.edu}}


\abstract[Summary]{
We explore the probabilistic partition of unity network (PPOU-Net) model in the context of high-dimensional regression
problems and propose a general framework focusing on adaptive dimensionality reduction. With the proposed framework, the target function is approximated by a mixture of experts model on a low-dimensional manifold, where each cluster is associated with a local fixed-degree polynomial. We present a training strategy that leverages the expectation maximization (EM) algorithm. During the training, we alternate between (i) applying gradient descent to update the DNN coefficients; and (ii) using closed-form formulae derived from the EM algorithm to update the mixture of experts model parameters. Under the probabilistic formulation, step (ii) admits the form of embarrassingly parallelizable weighted least-squares solves. The PPOU-Nets consistently outperform the baseline fully-connected neural networks of comparable sizes in numerical experiments of various data dimensions. We also explore the proposed model in applications of quantum computing, where the PPOU-Nets act as surrogate models for cost landscapes associated with variational quantum circuits.}

\keywords{deep learning, dimensionality reduction, mixture of experts, nonparametric regression}


\maketitle


\section{Introduction}\label{sec:introduction}

Using deep learning methods to approximate continuous functions and operators is a critical problem in machine learning theory and high-fidelity real-time engineering simulation applications. Recent discoveries in neural network approximation theory motivate novel strategies that combine deep neural networks (DNNs) with basis functions from classical approximation theory. \cite{yarotsky2017error,yarotsky2018optimal,opschoor2020deep} 
Data-driven learning has also demonstrated great potential to facilitate efficient engineering simulations. \cite{kirchdoerfer2016data,sirignano2018dgm,freund2019dpm,xu2020physics,fan2020solving,xu2022physics} However, many tools are designed for specific tasks and cannot be easily generalized to other engineering systems. \cite{holland2019field} Moreover, although uncertainty quantification is required in many applications, it may not be naturally produced by the approximation scheme itself. \cite{trask2021probabilistic} Instead, this work explores a general framework based on combining DNNs with polynomial approximation schemes that provides confidence regions in addition to point estimations.

Neural networks have achieved great success in solving classification problems, where the evaluation metrics are often defined as the fraction of training or testing examples that are classified correctly. \cite{michie1994machine, kotsiantis2007supervised} Applications range from object detection for autonomous driving \cite{stallkamp2012man}, virtual screening in drug discovery \cite{adeshina2020machine},  seizure type classification \cite{saputro2019seizure}, to the prediction of error-prone regions in fluid dynamics simulations. \cite{ling2015evaluation} In solving these problems, DNNs learn a partition of the data domain based on similarities in the training examples. Despite the success in classification tasks, deep learning models are subject to the curse of dimensionality: the DNN models do not exhibit the traditionally expected convergence with increasing degrees of freedom. \cite{verleysen2005curse, poggio2017and, bach2017breaking} Solving regression problems with high accuracy (e.g., > 99\%), a requirement for most scientific computing tasks,  remains a challenge for many deep learning models.

We consider a general high-dimensional regression problem with $N$ training examples $\{\mathbf{x}^{(n)}, y^{(n)} \}_{n=1}^N$, where $\mathbf{x}^{(n)} \in \Omega \subset \mathbb{R}^d$ and  $ y^{(n)}\in \mathbb{R}$. The goal is to model the mapping from $\mathbf{x}$ to $y$ in a Banach space $V$ and to discover potential lower-dimensional latent structures of the data. 

Recent discoveries in neural network approximation theory provide theoretical motivation for DNNs that break the curse of dimensionality.\cite{poggio2017and, bach2017breaking} Yarotsky proved the existence of DNNs that achieve spectral convergence with respect to their width and depth. \cite{yarotsky2017error,yarotsky2018optimal} He et al.\ established the connection between shallow networks with ReLU activation functions and linear finite element methods (FEM).\cite{he2018relu} Opschoor et al.\ demonstrated the existence of weights and biases values for DNNs to emulate a wide range of basis functions, and the approximation rates from ReLU-based DNNs were shown to closely match the best available approximation rates by piecewise polynomial spline functions.\cite{opschoor2020deep} Cyr et al.\ introduced an adaptive basis viewpoint of neural network training and initialization, proposing the LSGD algorithm and the Box initialization procedure that facilitate robust convergence as a function of both width and depth for PDE applications using physics-informed neural networks.\cite{cyr2020robust} Lee et al.\ proposed the partition of unity network (POU-Net) which adds polynomial regression explicitly into the model architecture, in order to exploit the ability of DNNs to partition the data domain into sub-domains where least-squares polynomial fits are used to achieve $hp$-convergence.\cite{lee2021partition} Inspired by the deterministic model \cite{lee2021partition} and multigrid methods,\cite{trottenberg2000multigrid} Trask et al.\ showed that a natural extension of POU-Net, the probabilistic partition of unity network (PPOU-Net), can be interpreted as a mixture of experts (MoE) model, and proposed an expectation-maximization (EM) training strategy as well as a hierarchical architecture to accelerate and improve the conditioning of the training process.\cite{trask2021probabilistic, trask2022hierarchical}

Classical approximation methods enjoy the advantages of computational efficiency and convergence guarantee in solving local, low-dimensional regression problems, but they often struggle in high dimensions or as global approximants.\cite{boyd2001chebyshev} Examples of such classical methods include truncated expansions in orthogonal polynomials (e.g., Chebyshev polynomials, Legendre polynomials, Hermite polynomials)\cite{trefethen2019approximation,shuman2018distributed} and Fourier basis functions \cite{grafakos2008classical}, rational functions \cite{newman1979approximation}, radial basis functions \cite{buhmann2003radial}, splines \cite{wegman1983splines}, wavelets \cite{mallat1999wavelet}, kernel methods \cite{donoho1989projection}, sparse grid \cite{nobile2008sparse}, etc. 
In high dimensions, high computational cost and ill-conditioning result in limited accuracy. For example, the total number of basis functions increases exponentially in the dimension of data domain for polynomials, Fourier basis functions, and wavelets; the Euclidean distance between two random points tends to a constant in high dimensions, causing difficulties for radial basis functions and stationary kernel methods. Localized methods such as splines, despite better numerical stability, require careful choice of nodal degrees of freedom or the prior construction of a mesh, both challenging in high dimensions.\cite{boyd2001chebyshev, ibrahimoglu2016lebesgue}

\begin{figure}[htbp]
\centering
\begin{minipage}{0.5\linewidth}
\includegraphics[width=1.05\columnwidth]{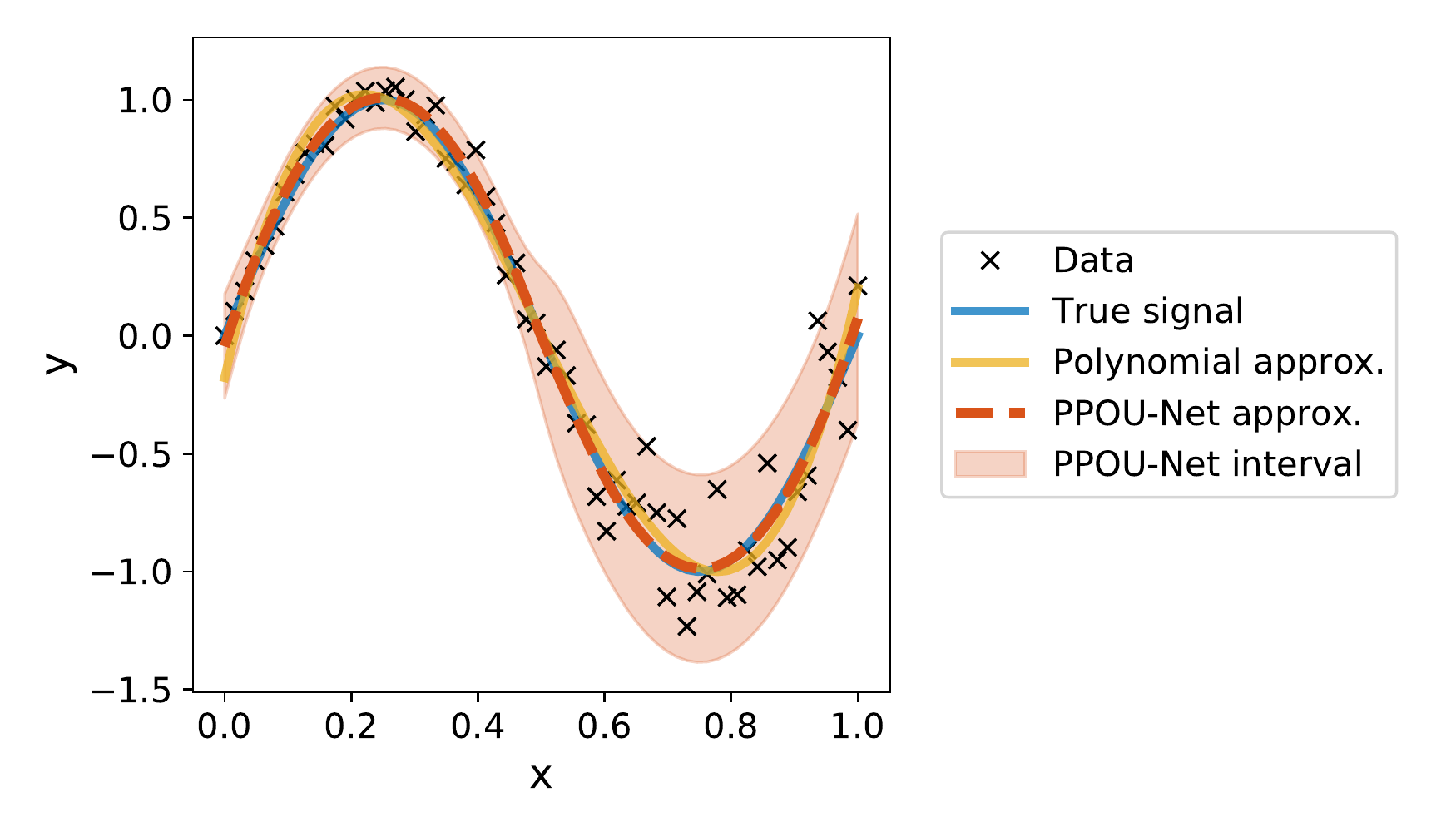}
\end{minipage}
\begin{minipage}{0.479\linewidth}
\includegraphics[width=1.05\columnwidth]{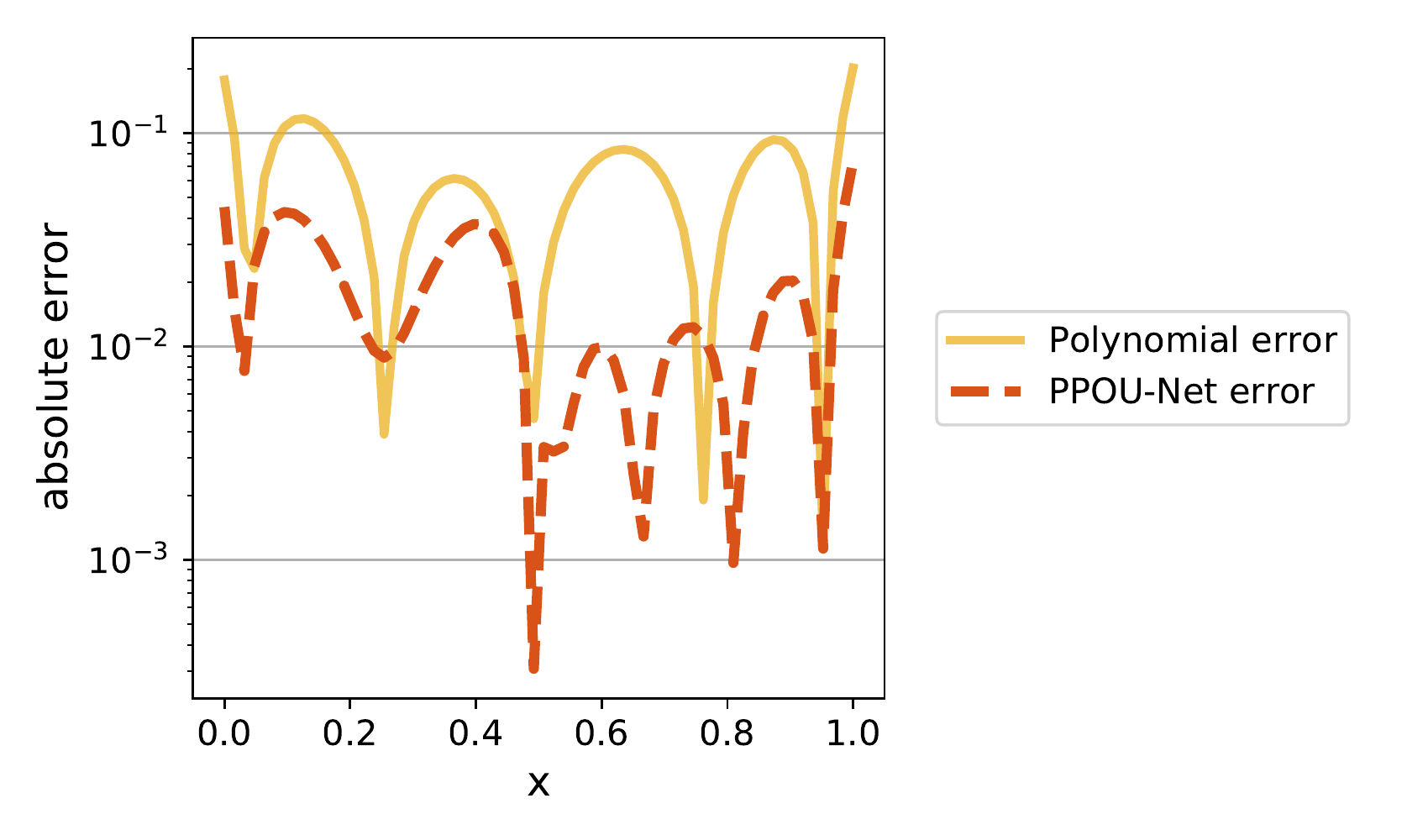} 
\end{minipage}
\caption{\label{fig:fig1} A cubic expansion in Chebyshev polynomials and a PPOU-Net fitted to a sine signal with heterogeneous noise. Both approximations are plotted against the scattered data and the true signal (left). Compared to the polynomial fit, the PPOU-Net model achieves lower absolute error (right), and quantifies the uncertainty in the data.}
\end{figure}

In this work, we extend the PPOU-Net model with a focus on dimensionality reduction for high-dimensional regression problems. The goal is to combine and leverage the advantages of neural networks and classical approximation methods: we use neural networks to perform dimensionality reduction and to partition the data domain, and use classical approximation methods to achieve efficient, robust, local convergence.  The contributions of this work are:
\begin{itemize}
\item generalizing the PPOU-Net \cite{trask2022hierarchical} to a high-dimensional hybrid regression model, consisting of an encoder, a classifier, and a chosen approximation basis, and investigating two alternative model architectures;
\item deriving an EM-motivated training strategy and proposing a heuristic loss function to facilitate robust training;
\item introducing a stabilized model for the background noise;
\item conducting numerical experiments with applications in quantum approximate optimization of various data dimensions.
\end{itemize}
When the data exhibit a low-dimensional structure, the encoder warps the input domain onto a low-dimensional latent space, and the classifier divides it into sub-regions. On each latent space sub-region, local least-square solves can efficiently compute polynomial approximations of the target function.
\Cref{fig:fig1} compares the PPOU-Net prediction (including a point estimation and a 95\% confidence region) to a polynomial fit.

We review the literature on scientific machine learning, high-dimensional uncertainty quantification, and mixture of experts models in \Cref{sec:related_work}. We recall the assumptions and training strategy of the basic PPOU-Net model in \Cref{sec:methods}, before we discuss two modifications to the basic model for high-dimensional problems and noisy observations in \Cref{sec:variants_model}. In \Cref{sec:numerical}, we present the results of numerical experiments with an extensive range of data dimensions. 

\section{Related Work}\label{sec:related_work}

\subsection{Scientific machine learning} Scientific data are often high-dimensional, multiscale, limited in amount, and affected by heterogeneous noise. \cite{baker2019workshop} Due to the high computational cost of high-fidelity numerical simulations \cite{freund2019dpm}, there have been significant efforts in leveraging machine learning to assist engineering simulations in the
past decade.  Physics-informed machine learning builds deep learning models, e.g., physics-informed neural networks (PINNs), for physical systems to solve both forward and inverse problems, where deviations from the physical constraints (e.g., conservation laws) are included in the deep learning models as penalty terms in the loss function. \cite{lagaris1998artificial, han2018solving, berg2018unified, raissi2019physics, karniadakis2021physics} Physics-constrained machine learning, on the other hand, enforces the physical relations more strictly by propagating numerical gradients through numerical PDE schemes using automatic differentiation. \cite{xu2020physics, fan2020solving, huang2020learning, xu2022physics, xu2022learning} In addition, neural networks as highly flexible functions are used as surrogate models for implicit physical relations to provide end-to-end approximate simulation outcomes \cite{tripathy2018deep, sirignano2018dgm, schwab2019deep} or as augmentations and corrections to traditional schemes. \cite{freund2019dpm}

\subsection{High-dimensional uncertainty qualification}

Dimensionality reduction plays a critical role in controlling the computational cost, avoiding overfitting, and allowing for straightforward visualization of complex engineering systems. High-dimensional uncertainty quantification using dimensionality reduction often involves the discovery of a low-dimensional latent manifold, followed by surrogate modeling in the discovered space of reduced dimension. \cite{kontolati2022survey} Dimension-reducing surrogate methods leverage ideas from Gaussian process (GP) regression \cite{damianou2013deep, calandra2016manifold}, active subspaces
\cite{constantine2015active, constantine2014computing, constantine2015exploiting}, kernel-PCA \cite{scholkopf1997kernel, mika1998kernel, hoffmann2007kernel}, polynomial chaos expansions (PCEs) \cite{lataniotis2020extending}, and other numerical schemes based on low-rank tensor decomposition. \cite{doostan2013non, konakli2016reliability, bigoni2016spectral, gorodetsky2018gradient, he2020high}

Deep learning motivates another class of methods in dimensionality reduction. Hinton and Salakhutdinov proposed autoencoders to learn a latent representation of training examples in a data-driven way. \cite{hinton2006reducing} Huang et al.\ and Li et al.\ proposed dimensionality reduction methods that use DNNs as feature maps for GP models \cite{huang2015scalable, li2020deep}. Convolutional neural networks (CNNs) \cite{khoo2021solving} and convolutional autoencoders \cite{boncoraglio2021active} were also leveraged in high-dimensional uncertainty quantification. Bayesian statistics motivated variational autoencoders (VAEs) \cite{kingma2013auto, kingma2019introduction} and Bayesian deep convolutional networks \cite{zhu2018bayesian, zhu2019physics}, providing uncertainty quantification for problems involving stochastic PDEs in the limit of small training data size.
Lu et al.\ proposed the DeepONet to approximate nonlinear operators and provide uncertainty quantification for PDE systems. \cite{lu2021learning}

\subsection{Mixture of experts and expectation maximization} 
Introduced by Jacobs et al.\ \cite{jacobs1991adaptive}, mixture of experts (MoE) is among the most popular ``combining methods" and has shown great potential in machine learning applications. \cite{masoudnia2014mixture}  A general MoE consists of a gating network and a set of expert models, where the gating network chooses a model from the set of expert models for evaluation. We refer readers to \cite{masoudnia2014mixture} by Masoudnia and Ebrahimpour for a survey of MoE models.  Jordan and Jacobs proposed the expectation maximization (EM) algorithm for training MoE models. \cite{jordan1994hierarchical, jordan1995convergence} Recent works have leveraged MoE for various deep learning tasks including manifold approximation and language processing. Schonsheck et al.\ proposed the chart auto-encoders for manifold structured data \cite{schonsheck2019chart}, where DNNs are used to learn a chart (i.e., a partition of unity) of the data manifold. Fedus et al.\ proposed the switch routing of MoE models that leads to a significant reduction of computational cost. 
\cite{fedus2021switch}  The PPOU-Net \cite{trask2022hierarchical} can be interpreted as an MoE model whose gating function is parameterized with a DNN.  The partitioning from MoE allows for local and reliable approximations by low-degree polynomials, similar to the classical finite element method, but the PPOU-Nets do not require the human-in-the-loop construction of a mesh.

\section{Basic PPOU-Net model} \label{sec:methods}

\subsection{Deterministic POU-Net model}

We consider a general regression problem with scattered observations $\{\mathbf{x}^{(n)}, y^{(n)} \}_{n=1}^N$, where $\mathbf{x}^{(n)} \in \Omega \subset \mathbb{R}^d$ and  $ y^{(n)}\in \mathbb{R}$. We choose a set of basis polynomials $\{p_k(\mathbf{x})\}_{k=1}^K$ that are supported on $\Omega$ and span a Banach space $V$. Introduced by Lee et al.\ \cite{lee2021partition}, the POU-Net model is a hybrid DNN/polynomial method that approximates the target mapping with a mixture of polynomials. Fixing an input $\mathbf{x}^{(n)}$, the model output is given by a weighted average of $J$ polynomials $\{\mu_j\}_{j=1}^J$ 
\begin{equation*}
    y_\text{POU}(\mathbf{x}^{(n)})
    = \left< \phi ( \mathbf{x}^{(n)};\theta ),\,\mu ( \mathbf{x}^{(n)}) \right>
    = \sum_{j} \phi_j(\mathbf{x}^{(n)};\theta)\, \mu_j(\mathbf{x}^{(n)}) = \sum_{j} \phi_j(\mathbf{x}^{(n)};\theta) \sum_{k} c_{j, k} p_k(\mathbf{x}^{(n)})
\end{equation*}
where the weights $\{ \phi_j(\mathbf{x}^{(n)};\theta) \}_{j=1}^J$ are nonnegative and $\sum_{j=1}^J \phi_j(\mathbf{x}^{(n)};\theta)=1$. The mapping $\phi: \mathbb{R}^d \to \mathbb{R}_+^J $ is parameterized as a DNN with parameters $\theta$.
Each polynomial is computed as an expansion in the chosen basis $\{p_k\}_{k=1}^K$, where the coefficients for $\mu_j$ are $\{c_{j, k}\}_{k=1}^K.$  Lee et al.\ \cite{lee2021partition} used the two-phase LSGD algorithm \cite{cyr2020robust} to train POU-Nets, which alternatively updates $\{c_{j, k}\}_{k=1}^K$ using least-squares solves and $\theta$ using gradient descent. The algorithm requires a scheduled regularization of the least-squares solves in the first phase, where the regularization hyperparameter decreases geometrically as the training progresses. \added{Each training iteration of the deterministic POU-Net requires solving a least-squares system with matrix size $N\times JK $. }

\subsection{Probabilistic POU-Net model}
Trask et al.\ reformulated the POU-Nets from a probabilistic perspective and showed that it is equivalently a MoE model where each expert is associated with a polynomial. \cite{trask2021probabilistic, trask2022hierarchical}
Given an input $\mathbf{x}^{(n)}$, we introduce a categorical latent variable $z^{(n)}$ supported on the finite set $\{1,\ldots, J\}$. The output $Y(\mathbf{x}^{(n)})$ is modeled as
\begin{equation}
    \begin{aligned}
    z^{(n)} & \sim \  \operatorname{Cat}\left(J, \ \phi \left( \mathbf{x}^{(n)};\theta \right) \right)\\
    Y(\mathbf{x}^{(n)}) \mid z^{(n)} & \sim\ \mathcal{N}\left( \mu_{z^{(n)}} \left( \mathbf{x}^{(n)} \right), \, \sigma^2_{z^{(n)}} \right).
    \end{aligned} \label{eq:assumptions1}
\end{equation} 
Equivalently, the output $Y(\mathbf{x}^{(n)})$ follows a Gaussian mixture distribution with $J$ components, where the cluster weights $\phi ( \mathbf{x}^{(n)}; \theta ) = \{ \phi_j(\mathbf{x};\theta) \}_{j=1}^J $ are a partition of unity parameterized by $\theta$. The closed-form expressions for the expectation and variance of the model output are given by
\begin{align*}
\mathbb E [Y(\mathbf{x}^{(n)}) ]
    &= \sum_{j} \phi_j\left(\mathbf{x}^{(n)};\theta \right)\, \mu_j \left(\mathbf{x}^{(n)} \right) = \sum_{j} \phi_j \left(\mathbf{x}^{(n)};\theta \right) \sum_{k} c_{j, k} p_k \left(\mathbf{x}^{(n)} \right)\\
     \operatorname{Var} [Y(\mathbf{x}^{(n)}) ]
    &= \sum_{j} \phi_j\left(\mathbf{x}^{(n)};\theta\right) \left ( \sigma_j^2   +  \mu_j\left(\mathbf{x}^{(n)}\right)  ^2 \right) -  \Big( \sum_{j} \phi_j\left(\mathbf{x}^{(n)};\theta\right)\, \mu_j\left(\mathbf{x}^{(n)}\right)  \Big)^2.
\end{align*}
Leveraging the ideas of EM \cite{dempster1977maximum,blei2017variational} and the LSGD algorithm \cite{cyr2020robust}, Trask et al.\ proposed a training strategy for PPOU-Nets \cite{trask2021probabilistic} and obtained closed-form update formulae under the probabilistic assumptions. The training strategy is summarized in \Cref{alg:em} and the derivation is included in \Cref{sec:basic_ppou_derivation}. \added{Each training iteration of the PPOU-Net solves $J$ decoupled least-squares systems with matrix size $N\times K $. The least squares solves are embarrassingly parallel and map well onto accelerated GPUs \cite{trask2022hierarchical}.  The introduction of the EM step reduces the computational time of the least-squares solves from $\mathcal{O}(NJ^2K^2)$ to in $\mathcal{O}(NK^2)$. }

\section{Dimensionality-reducing PPOU-Nets}\label{sec:variants_model}
\subsection{Dimensionality reduction with PPOU-Nets}

\begin{figure}[htbp]
    \centering
    \begin{minipage}[l]{0.22\textwidth}
        PPOU-Net, serial
    \end{minipage}
    \begin{minipage}[c]{0.65\textwidth}
        \includegraphics[width=\columnwidth]{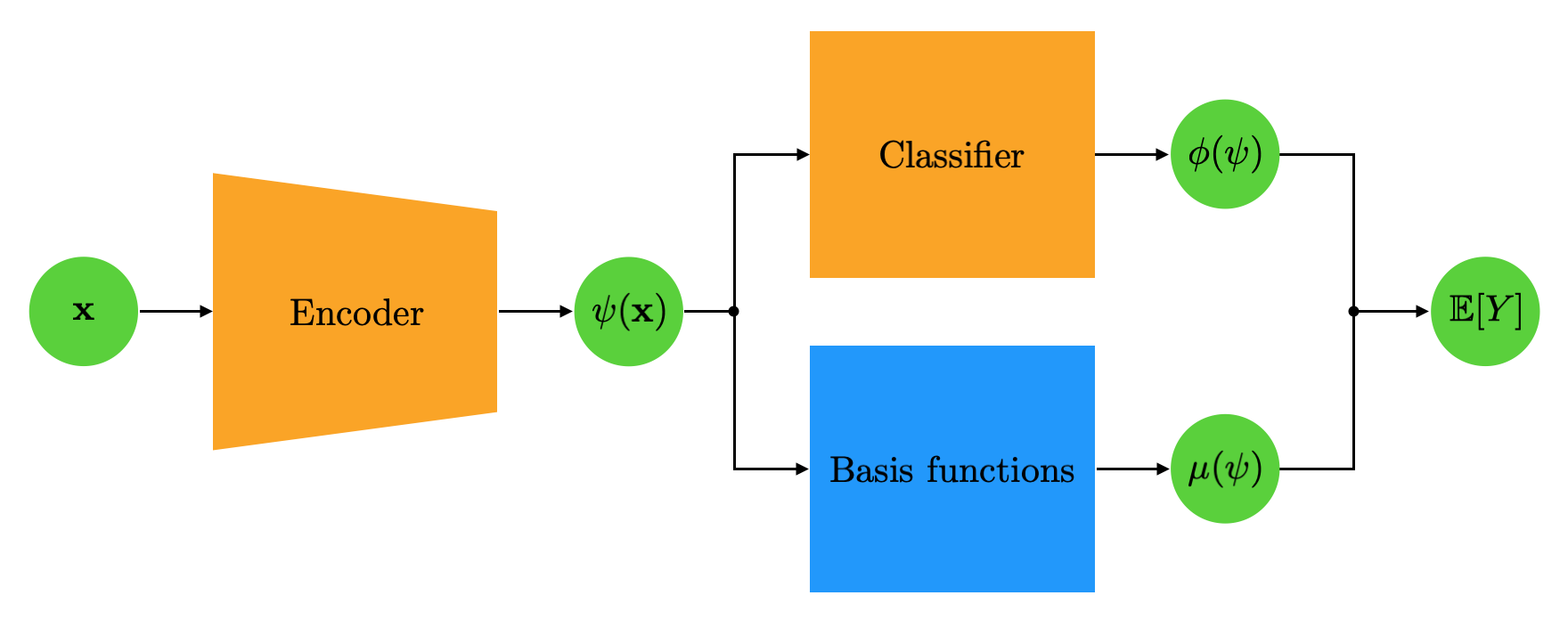}
    \end{minipage}\\
    \begin{minipage}[l]{0.22\textwidth}
        PPOU-Net, parallel
    \end{minipage}
    \begin{minipage}[c]{0.65\textwidth}
        \includegraphics[width=\columnwidth]{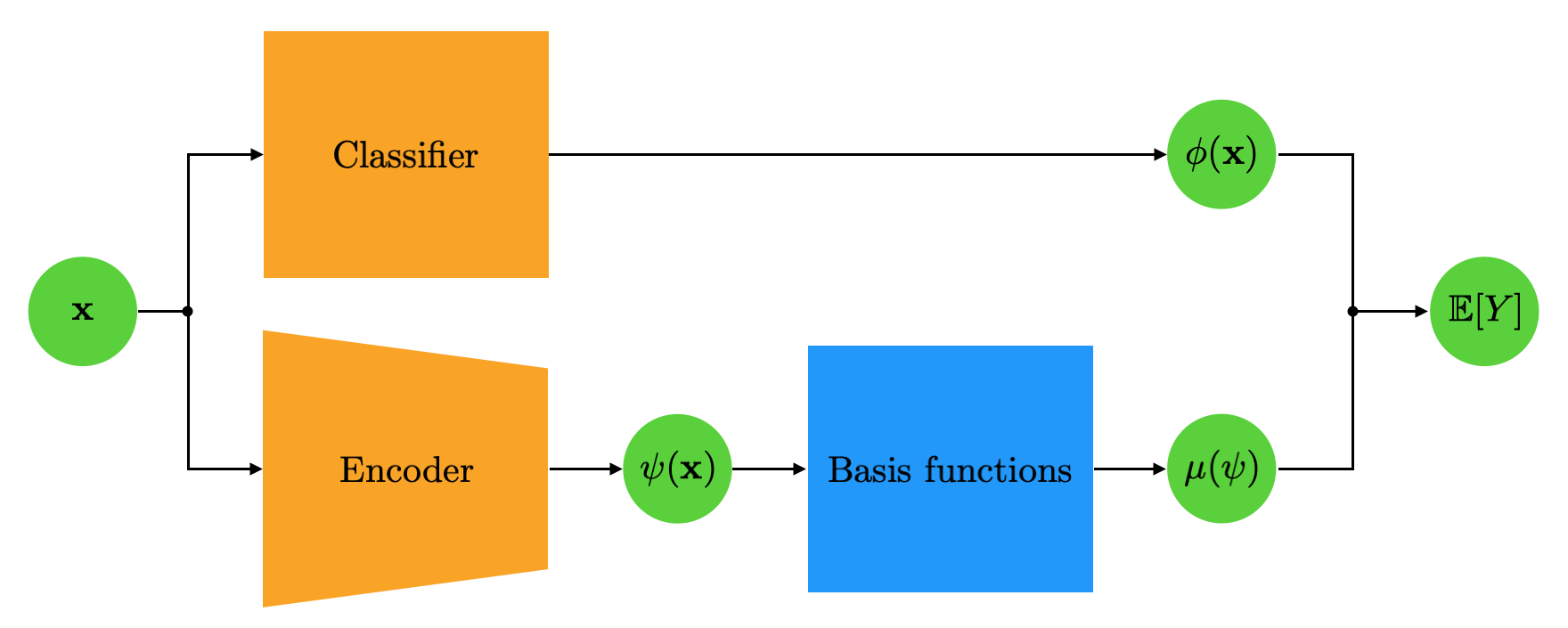}
    \end{minipage}
    \caption{Alternative architectures of dimensionality-reducing PPOU-Net. Due to different positions of the classifier, the serial PPOU-Net provides better visualization and interpretability, while the parallel PPOU-Net is more flexible. }
    \label{fig:serial_parallel}
\end{figure}

In this subsection, we augment the basic PPOU-Net model with an encoder $\psi$, with its collection of weights and biases denoted as $\theta_\psi$. In addition, we note that the DNN component of the basic PPOU-Net can be generalized to an arbitrary classifier. We consider two model architectures, where we connect the encoder with the classifier (PPOU-Net) in serial or in parallel (see \Cref{fig:serial_parallel}). We denote the weights an biases of the classifier $\phi$ as $\theta_\phi$ and let $\theta=\theta_\phi \cup \theta_\psi$. The goal of the model augmentation is to search for a low-dimensional representation of the input data and to improve the efficiency of the polynomial fits.

The key difference of the two architectures is the classifier input: the classifier operates on the raw data in the parallel architecture, and on the encoded data in the serial architecture. As a result, the parallel model is designed with more flexibility, while the serial model provides better visualization and interpretability of the latent space. For the numerical experiments we use the serial PPOU-Net in \Cref{subsec:numerical_swiss_and_trefoil} and \Cref{subsec:numerical_quantum}, and we use both architectures in \Cref{subsec:numerical_rings}. The simulated data in \Cref{subsec:numerical_sine} are low-dimensional, so we simplify the encoder to an identity mapping; this reduces both architectures to the basic PPOU-Net.

\paragraph{Model assumptions}
As a natural extension to \Cref{eq:assumptions1}, the assumptions of the MoE model for high-dimensional data are
\begin{align*}
    z^{(n)} & \sim \operatorname{Cat}\left(J, \ \phi \left( \mathbf{x}^{(n)}; \theta_\phi \right) \right)\\
    Y(\mathbf{x}^{(n)}) \mid z^{(n)} 
    & \sim \mathcal{N}\left( \mu_{z^{(n)}} \left( \psi \left( \mathbf{x}^{(n)} ; \theta_\psi \right)\right), \, \sigma^2_{z^{(n)}} \right)
\end{align*}
Fixing an input $\mathbf{x}^{(n)}$,  the $j$-th Gaussian component is centered at $\mu_j\left(\left( \psi( \mathbf{x}^{(n)} ; \theta_\psi)\right)\right)$ and has variance $\sigma_j^2>0$, for each $j=1, \ldots, J$. We can reparameterize the mixture model as
\begin{equation*} \label{eq:gaussian_mixture3}
\begin{aligned}
    Y
(\mathbf{x}^{(n)}) &=\sum_{j} \phi_j(\mathbf{x}^{(n)};\theta_\phi)\left( \mu_j\left( \psi(\mathbf{x}^{(n)}; \theta_\psi) \right)+ \epsilon_j \right) =  \sum_{j} \phi_j(\mathbf{x}^{(n)}; \theta_\phi) \Big(\sum_{k} c_{j, k} p_k\left( \psi(\mathbf{x}^{(n)}; \theta_\psi)\right) + \epsilon_j \Big) \end{aligned}
\end{equation*}
where $\epsilon_j^2\sim \mathcal{N}(0, \sigma_j^2)$ for each $j=1,\ldots,J$. The closed-form expressions for the expectation and variance of the $n$-th model output are given by
\begin{equation*} \label{eq:mean_and_variance3}
\begin{aligned}
\mathbb E [Y(\mathbf{x}^{(n)})] 
    &= \sum_{j} \phi_j(\mathbf{x}^{(n)};\theta_\phi)\, \mu_j \left( \psi(\mathbf{x}^{(n)}; \theta_\psi) \right) = \sum_{j} \phi_j(\mathbf{x}^{(n)};\theta_\phi) \sum_{k} c_{j, k} p_k(  \psi(\mathbf{x}^{(n)}; \theta_\psi) )\\
     \operatorname{Var} [Y(\mathbf{x}^{(n)}) ]
    &= \sum_{j} \phi_j(\mathbf{x}^{(n)};\theta_\phi) \Big( \sigma_j^2   +  \mu_j ( \psi(\mathbf{x}^{(n)}; \theta_\psi) )^2 \Big) -  
    \Big( \sum_{j} \phi_j(\mathbf{x}^{(n)};\theta_\phi)\, \mu_j \big( \psi(\mathbf{x}^{(n)}; \theta_\psi) \big) \Big)^2.
    \end{aligned}
\end{equation*}

\paragraph{Expectation-maximization training strategy} Motivated by \cite{trask2021probabilistic}, we derive an EM training strategy in the context of high-dimensional regression.
The log-likelihood of the data $\{\mathbf{x}^{(n)}, y^{(n)}\}_{n=1}^N$ can be expressed in terms of the latent variable model
\begin{equation} \label{eq:log_likelihood3}
\log p( \{ \mathbf{x}^{(n)}, y^{(n)} \}_n ; \theta, c, \sigma^2) = \sum_{n} \log p( \mathbf{x}^{(n)}, y^{(n)} ; \theta, c, \sigma^2)
=  \sum_{n} \log  \sum_{z^{(n)}} p( \mathbf{x}^{(n)}, y^{(n)}, z^{(n)} ; \theta, c, \sigma^2).
\end{equation}
For any valid distributions $ \{ w^{(n)} (z^{(n)}) \}_{n=1}^N$ where the positive probability mass values are given by $w_j^{(n)}:= p_{w^{(n)}} (z^{(n)}=j) > 0$, by Jensen's Inequality,  the log-likelihood has an evidence lower bound (ELBO)
\begin{equation}
\begin{aligned}
 \eqref{eq:log_likelihood3} \geq \, &  \ell_{\text{ELBO}} 
 ( \theta_\phi, \theta_\psi, c, \sigma^2; \{\mathbf{x}^{(n)}, y^{(n)}, w^{(n)} \}_n) \\ &=   \sum_{n} \sum_{z^{(n)}} w^{(n)}(z^{(n)})   \log \frac{ p( \mathbf{x}^{(n)}, y^{(n)}, z^{(n)} ; \theta_\phi, \theta_\psi, c, \sigma^2) }{  w^{(n)}(z^{(n)})  }\\
 &= \sum_{n} \sum_{j} w^{(n)}_j  \left( \log \phi_j( \mathbf{x}^{(n)};\theta_\phi ) + \log \mathcal{N} \left( y^{(n)} \mid \mu_j  \left(  \psi(\mathbf{x}^{(n)}; \theta_\psi ); c \right) , \sigma^2_j \right)\right) +C.
 \label{eq:elbo3}
 \end{aligned}
\end{equation}
Taking the partial derivative of the ELBO \eqref{eq:elbo3} with respect to $c_{j,k}$ and setting it to zero, 
\begin{equation*}
    \begin{aligned}
    \frac{\partial \ell_{\text{ELBO}}}{\partial c_{j,k}}  &=  \sum_{n }  w_j^{(n)} \frac{y^{(n)} - \mu_j  \left(  \psi(\mathbf{x}^{(n)}; \theta_\psi )\right) }{  \sigma_j^2} \  p_k(x^{(n)}) \\&= \sum_{n } \frac{w_j^{(n)}}{\sigma_j^2}
    \Big( y^{(n)} p_k(\psi(\mathbf{x}^{(n)}; \theta_\psi ))
    - \sum_{l} c_{j, l} p_l(\psi(\mathbf{x}^{(n)}; \theta_\psi )p_k(\psi(\mathbf{x}^{(n)}; \theta_\psi ) \Big) =0
    \end{aligned}
\end{equation*} 
we derive that the optimal value of $\mathbf{c}_j= \{ c_{j,k}\}_ {k}$ satisfies the normal equation 
$$ P_j^\top W_j P_j \mathbf{c}_j = P_j^\top W_j\, \mathbf{y}, $$  
where $P_j$ has entries $[P_j]_{n, k} = p_k\left( \psi( \mathbf{x}^{(n)}; \theta_\psi) \right) ,  W_j =\operatorname{diag}( \{w^{(n)}_{j} \}_{n=1}^N  )$, and $\mathbf{y}=\{y^{{(n)}} \}_{n=1}^N$. Hence, fixing $\{\theta_\phi, \theta_\psi, \{w^{(n)}\}_n\}$, the polynomial coefficients $\mathbf{c}_j$ are a solution to the weighted least squares problem in the latent space
$$
    \mathbf{c}_j = \text{argmin}_{\{c_{j, l}\}} \sum_{n} w_j^{(n)} \Big( y^{(n)} - \sum_{l} c_{j,l} p_l\left( \psi( \mathbf{x}^{(n)}; \theta_\psi) \right) \Big)^2.
$$
Similar to Gaussian mixture models, maximizing the ELBO \eqref{eq:elbo3} with respect to $\sigma^2_j$ yields
$$
    \sigma_j^2 = \frac{\sum_{n=1}^N w_j^{(n)} \left( y^{(n)} - \mu_j   \left(  \psi(\mathbf{x}^{(n)}; \theta_\psi ) \right)  \right)^2 
    }{\sum_{n=1}^N w_j^{(n)}}.
$$
\added{For dimension-reducing PPOU-Nets, as the polynomial basis is defined in the latent space, the total number of basis polynomials $K$ grows with the latent dimension $d'$ rather than the input dimension $d$. The dimension reduction of the polynomial basis, in addition to the GPU-accelerated least-squares solves, contributes significantly to the efficiency of the PPOU-Net model.}

\paragraph{The EM loss function} Following the EM algorithm, we update the weights and biases $\theta$ using gradient ascent with respect to the ELBO \eqref{eq:elbo3}, minimizing 
\begin{equation*} \begin{aligned}\label{eq:loss3}
        L(\theta) &= - \sum_{n} \sum_{j} w^{(n)}_j  \left( \log \phi_j( \mathbf{x}^{(n)};\theta_\phi ) + \log \mathcal{N} \left( y^{(n)} \mid \mu_j  \left(  \psi(\mathbf{x}^{(n)}; \theta_\psi ); c \right) , \sigma^2_j \right)\right) \\
        &= - \sum_{n} \sum_{j} w^{(n)}_j  \log \phi_j( \mathbf{x}^{(n)};\theta_\phi ) 
        -\sum_{n} \sum_{j} w^{(n)}_j   \log \mathcal{N} \left( y^{(n)} \mid \mu_j  \left(  \psi(\mathbf{x}^{(n)}; \theta_\psi ); c \right) , \sigma^2_j  \right).
\end{aligned}
\end{equation*}
We can write $L(\theta) = L_1 (\theta_\phi) + L_2 (\theta_\psi)$, where 
$$ L_1 (\theta_\phi) = - \sum_{n} \sum_{j} w^{(n)}_j  \log \phi_j( \mathbf{x}^{(n)};\theta_\phi ) $$
and 
\begin{align*}
    L_2 (\theta_\psi) &=  -\sum_{n} \sum_{j} w^{(n)}_j   \log \mathcal{N} \Big( y^{(n)} \mid \mu_j  \big(  \psi(\mathbf{x}^{(n)}; \theta_\psi ); c \big) , \sigma^2_j  \Big)\\
    &=  -\sum_{n} \sum_{j} w^{(n)}_j \Big(- \frac{1}{2} \Big( \frac{ y^{(n)} - \mu_j   \big(  \psi(\mathbf{x}^{(n)}; \theta_\psi ) \big) }{\sigma_j} \Big)^2   - \frac{1}{2}\log \big(2\pi \sigma_j^2\big)
    \Big)\\
    &= \sum_{j}  \frac{ 1 }{2\sigma^2_j}  \sum_{n} w^{(n)}_j   \Big( y^{(n)} - \mu_j   \left(  \psi(\mathbf{x}^{(n)}; \theta_\psi ) \right) \Big)^2 + \,C.
\end{align*}  
We note that $L_1$ is the cross entropy between the distributions of the classifier outputs $\phi_j( \mathbf{x}^{(n)};\theta_\phi )$ and the current posterior $w^{(n)}_j$. Thus, minimizing $L_1$ encourages the classifier to emulate the current posterior distribution from the E-step for each training example. In addition, $L_2$ is a weighted sum of the squared differences between the target output and the polynomials. A larger posterior weight $w^{(n)}_j$ and a smaller cluster variance $\sigma^2_j$ gives more weight to the difference between $y^{(n)}$ and $\mu_j(\psi(x^{(n)}))$.

\paragraph{An alternative loss function}

Motivated by the deterministic POU-Net, which minimizes the sum of squared errors between the observed output and the overall model expectation,  we introduce the alternative loss function where we combine the EM loss term $L_1(\theta_\phi)$ with the overall prediction error $\tilde{L}_2 (\theta_\psi)= \sum_n \left( y^{(n)} - \sum_j \phi_j(\mathbf{x}^{(n)}; \theta_\phi) \, \mu_j(\psi(\mathbf{x}^{(n)}; \theta_\psi))\right)^2$ to promote model accuracy
\begin{equation*}
        \tilde{L} (\theta) = - \sum_n \sum_j w^{(n)}_j \log \phi_j (\mathbf{x}^{(n)}; \theta_\phi ) 
        +  \sum_n   \Big(  y^{(n)} - \sum_j \phi_j \left(\mathbf{x}^{(n)}; \theta_\phi \right)\, \mu_j\left(  \psi(\mathbf{x}^{(n)}; \theta_\psi ) \right)  \Big) ^2.
\end{equation*}
Minimizing $L_1$ forces the classifier outcomes towards the posterior, while minimizing $L_2$ guarantees that the overall model prediction is close to the target output. When the posterior distribution assigns probability $1$ to one cluster and probability $0$ to the other clusters, $L_2$ and $\tilde{L}_2$ are equivalent up to a scalar factor. In some numerical experiments, we observe that the new loss function $\tilde L$ improves the conditioning of the training process.

We use a gradient descent optimizer, e.g., Adam, to update $\theta_\phi$ and $\theta_\psi$ in every M-step. The EM loss function is used in the numerical experiments in \Cref{subsec:numerical_swiss_and_trefoil}, \Cref{subsec:numerical_rings}, and \Cref{subsec:numerical_quantum}. The alternative loss function is used in \Cref{subsec:numerical_sine}. We summarize the training process for a dimensionality-reducing PPOU-Net in \Cref{alg:em_with_encoder}.

\begin{algorithm}[tbh!]
\caption{EM training loop for dimension-reducing PPOU-Nets}\label{alg:em_with_encoder}
\begin{algorithmic}
\State Repeat until convergence \{ \\
\begin{enumerate}
    \item (E-step) For $n=1,\ldots,N$ and $j=1,\ldots,J$, compute
    \begin{align*}
        w_j^{(n)} \gets &\   \frac{1}{\sigma_j} \exp 
        \Big( -\frac{ \left(y^{(n)} - \mu_j   \left(  \psi(\mathbf{x}^{(n)}; \theta_\psi ) \right) \right)^2 }{2\sigma_j^2} 
        \Big) \phi_j \left(\mathbf{x}^{(n)};\theta_\phi \right)\\
        w_j^{(n)} \gets & \  \frac{w_j^{(n)}}{\sum_{l=1}^J w_l^{(n)} } 
    \end{align*}
    \item (M-step) Update the DNN parameters $\theta= \theta_\phi \cup \theta_\psi$ using gradient descent, under the loss function
    \begin{equation*}
        L(\theta) = - \sum_{n=1}^N \sum_{j=1}^J w_j^{(n)} \log \phi_j \left(\mathbf{x}^{(n)}; \theta_\phi \right) +   \sum_{j}  \frac{ 1 }{2\sigma^2_j}  \sum_{n} w^{(n)}_j   \left( y^{(n)} - \mu_j   \left(  \psi(\mathbf{x}^{(n)}; \theta_\psi ) \right) \right)^2
\end{equation*}
    For $j=1,\ldots,J$, solve the weighted least-squares problem
    \begin{align*}
        \mu_j \gets \operatorname{argmin}_{g \in \text{span} \{p_1, \ldots, p_K \}} \sum_{n=1}^N w_j^{(n)} \left( y^{(n)} - g  \left(  \psi(\mathbf{x}^{(n)}; \theta_\psi ) \right)   \right)^2
    \end{align*}
    and update the variance term \begin{align*}
    \sigma_j^2 \gets \  \frac{\sum_{n=1}^N w_j^{(n)} \left( y^{(n)} - \mu_j   \left(  \psi(\mathbf{x}^{(n)}; \theta_\psi ) \right)  \right)^2 
    }{\sum_{n=1}^N w_j^{(n)}}
    \end{align*}
\end{enumerate}
\State \}
\end{algorithmic}
\end{algorithm}

\subsection{Background noise}\label{subsec:background noise}
In practice, measurements from scientific and engineering experiments are often subject to background noise or machine error. We can modify the model assumptions in \Cref{eq:assumptions1} to explicitly account for the background noise throughout the EM updates. \cite{melchior2018filling} 
The modified assumptions are
\begin{align*}
    z^{(n)} & \sim \operatorname{Cat}(J, \ \phi ( \mathbf{x}^{(n)};\theta ) )\\
    Y (\mathbf{x}^{(n)}) \mid z^{(n)}
    & \sim \mathcal{N} ( q_{z^{(n)}} ( \mathbf{x}^{(n)} ), \, \sigma^2_{z^{(n)}} ) + \mathcal{N}(0, \, \sigma^2_{0})
\end{align*}
where the variance of the background noise, denoted $\sigma^2_{0}$, is constant for all training samples. The background noise is independent from the \added{trainable} variance of the mixture model. The overall model output can be reparameterized as
\begin{equation*}  
    Y(\mathbf{x}^{(n)}) =\sum_{j} \phi_j(\mathbf{x}^{(n)};\theta)\left( \mu_j(\mathbf{x}^{(n)})+ \epsilon_j  \right) + \epsilon_0 = \sum_{j} \phi_j(\mathbf{x}^{(n)}; \theta) \Big(\sum_{k} c_{j, k} p_k(\mathbf{x}^{(n)}) + \tilde{\epsilon}_j \Big) 
\end{equation*}  where $\tilde{\epsilon}_j
\sim \mathcal{N}(0, \sigma^2_j + \sigma^2_0)$ for each $j=1,\ldots, J$.
It follows that 
\begin{equation*}
\begin{aligned}
\mathbb E [Y(\mathbf{x}^{(n)}) ]
    &= \sum_{j} \phi_j(\mathbf{x}^{(n)};\theta)\, \mu_j(\mathbf{x}^{(n)}) = \sum_{j} \phi_j(\mathbf{x}^{(n)};\theta) \sum_{k} c_{j, k} p_k(\mathbf{x}^{(n)})\\
     \operatorname{Var} [Y(\mathbf{x}^{(n)}) ]
    &= \sum_{j} \phi_j(\mathbf{x}^{(n)};\theta) \left ( \sigma_j^2   + \sigma_0^2 + \mu_j(\mathbf{x}^{(n)})  ^2 \right) -  \Big( \sum_{j} \phi_j(\mathbf{x}^{(n)};\theta)\, \mu_j(\mathbf{x}^{(n)})  \Big)^2.
    \end{aligned}
\end{equation*} 

\begin{algorithm}[tbh!]
\caption{EM training loop for PPOU-Nets under background noise}\label{alg:em_noise}
\begin{algorithmic}
\State Repeat until convergence \{ \\
\begin{enumerate}
    \item (E-step) For $n=1,\ldots,N$ and $j=1,\ldots,J$, compute
    \begin{align*}
        w_j^{(n)} \gets &\   \frac{1}{\sigma_j} \exp \Big( -\frac{ \left(y^{(n)} - \mu_j(\mathbf{x}^{(n)} ) \right)^2 }{2\sigma_j^2} \Big)  \phi_j(\mathbf{x}^{(n)};\theta)\\
        w_j^{(n)} \gets & \  \frac{w_j^{(n)}}{\sum_{l=1}^J w_l^{(n)} } \\
        b_j^{(n)} \gets & \  \mu_j(\mathbf{x}^{(n)} ) + \frac{\sigma_j^2}{\sigma_j^2+\sigma_0^2} \left(y^{(n)} - \mu_j(\mathbf{x}^{(n)} )\right)\\
        B_j \ \gets & \   \sigma_j^2 - \frac{\sigma_j^4}{\sigma_j^2+\sigma_0^2  }
    \end{align*}
    \item (M-step) Update the DNN parameters $\theta$ using gradient descent, under the loss function
    \begin{equation*}
        L(\theta) = - \sum_{n=1}^N \sum_{j=1}^J w_j^{(n)} \log \phi_j (\mathbf{x}^{(n)}; \theta  ) +   \sum_{n=1}^{N}   \Big(  y^{(n)} - \sum_{j=1}^J \phi_j  (\mathbf{x}^{(n)}; \theta )\, \mu_j  (\mathbf{x}^{(n)}  )  \Big) ^2
\end{equation*}
    For $j=1,\ldots,J$, solve the weighted least-squares problem
    \begin{align*}
        \mu_j \gets \operatorname{argmin}_{g \in \text{span} \{p_1, \ldots, p_K \}} \sum_{n=1}^N w_j^{(n)} \left( y^{(n)} - g \left(\mathbf{x}^{(n)}\right)   \right)^2
    \end{align*}
    and update the variance term \begin{align*}
    \sigma_j^2 \gets \  \frac{\sum_{n=1}^N w_j^{(n)}\Big(\big( b_j^{(n)} - \mu_j  ( \mathbf{x}^{(n)} )  \big)^2  + B_j\Big)
    }{\sum_{n=1}^N w_j^{(n)}}
    \end{align*}
    
\end{enumerate}
\State \}
\end{algorithmic}
\end{algorithm}
\clearpage
\added{
The background noise also functions as a lower bound for the overall variance and improves the model stability by preventing a divide-by-zero error in the EM updates. In this work, we assume that the background noise level $\sigma^2_{0}$ is constant across all data examples; however, this can be generalized to any space-dependent noise level $\sigma^2_{0}(\mathbf{x})$ based on prior knowledge of the system.}

We summarize the EM-inspired training strategy under background noise in \Cref{alg:em_noise}, where we use results derived in Section 2.2 of \cite{melchior2018filling}. Similar to the basic PPOU-Nets, the background noise PPOU-Nets can be extended to high-dimensional data. The background noise PPOU-Nets are used for experiments in \Cref{subsec:numerical_sine} and the first task in \Cref{subsec:numerical_quantum}, where the data are subject to heterogeneous noise.
\section{Numerical experiments}  \label{sec:numerical}

\subsection{Regression of smooth functions with spatially heterogeneous noise}\label{subsec:numerical_sine}

We first illustrate the ability of PPOU-Nets to approximate the uncertainty in the data, i.e., to learn confidence regions in addition to point estimations. We consider a sinusoidal wave on $x\in [0, 1]$ and define a normally distributed noise term $\epsilon(x)$. The noise term is centered at zero and its standard deviation increases linearly with $x$ under different multiplicative noise levels $\alpha$:
\[
    y = \sin(2\pi x) + \epsilon(x), \quad 
    \epsilon(x) \sim \mathcal{N}\big(0, (\alpha x)^2\big) 
\]
where we take $\alpha \in \{0.1, 0.2, 0.5\}$. For each noise level, the training data input $x$ consists of $N=1024$ evenly spaced points on $[0, 1]$. We \added{train} a PPOU-Net consisting of a Box-initialized ResNet with depth 4, width 8, and 4 partitions with quadratic functions \added{to learn the mapping from input $x$ to output $y$. }

\begin{figure}[th!]
\begin{minipage}{0.28\linewidth}
\centering
~~~~~~$\alpha=0.1$
\end{minipage}
\begin{minipage}{0.28\linewidth}
\centering
~~~~~~$\alpha=0.2$
\end{minipage}
\begin{minipage}{0.28\linewidth}
\centering
~~~~~~$\alpha=0.5$
\end{minipage}\\\vspace{0.1cm}
\begin{minipage}{0.28\linewidth}
\centering
\includegraphics[width=\columnwidth]{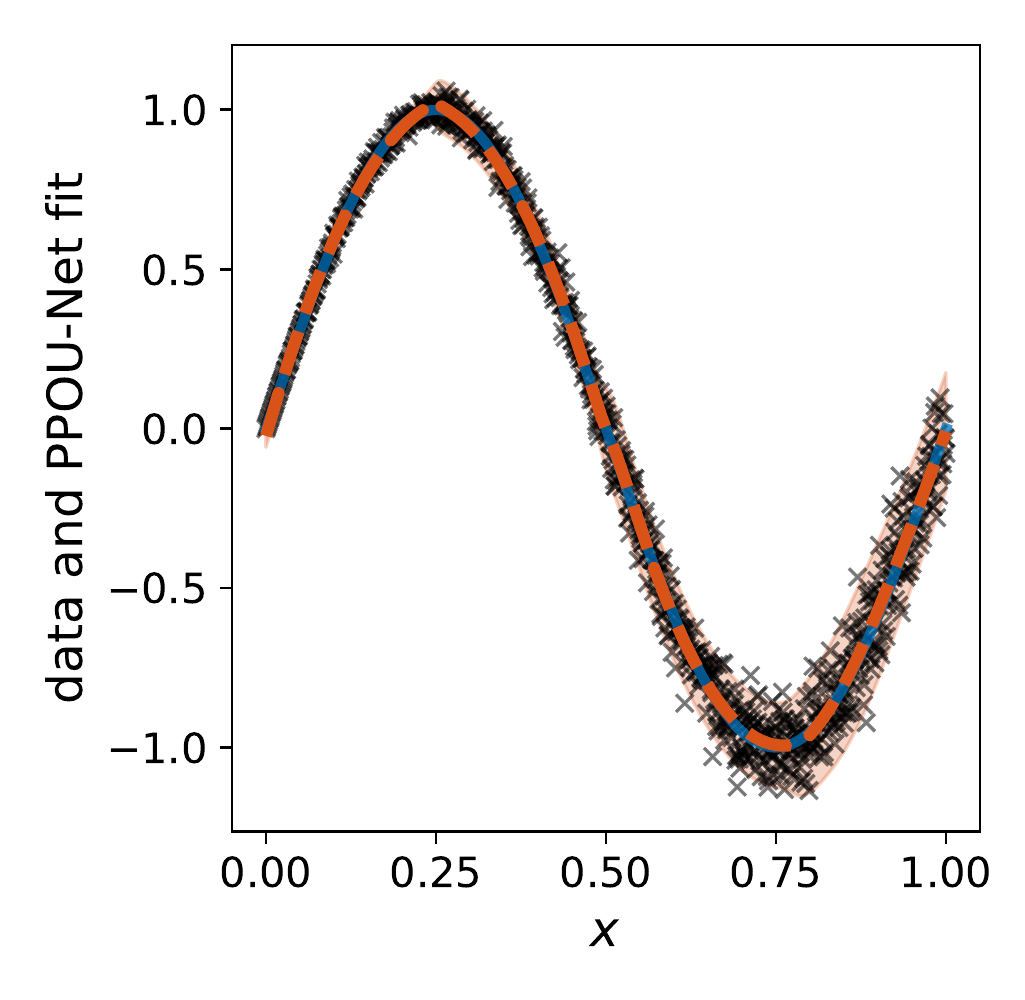}
\end{minipage}
\begin{minipage}{0.28\linewidth}
\centering
\includegraphics[width=\columnwidth]{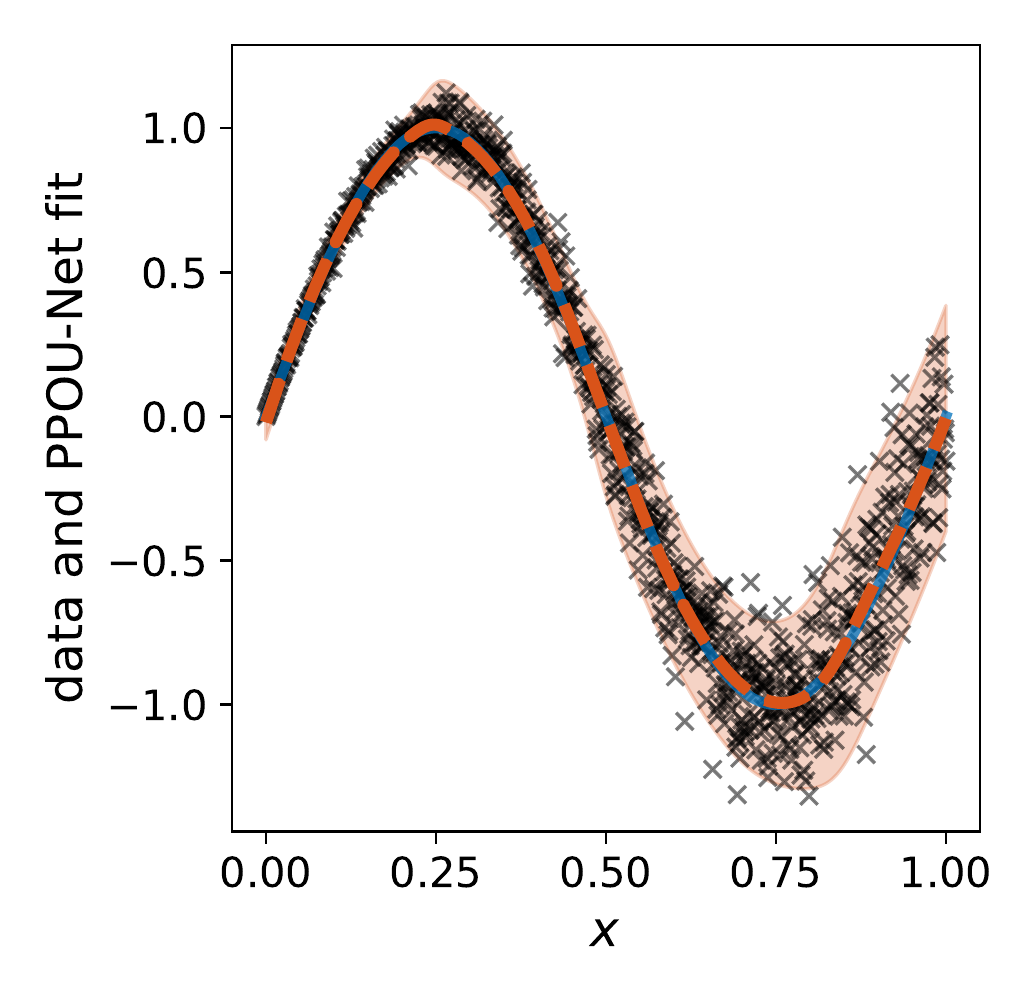}
\end{minipage}
\begin{minipage}{0.455\linewidth}
\centering
\includegraphics[width=\columnwidth]{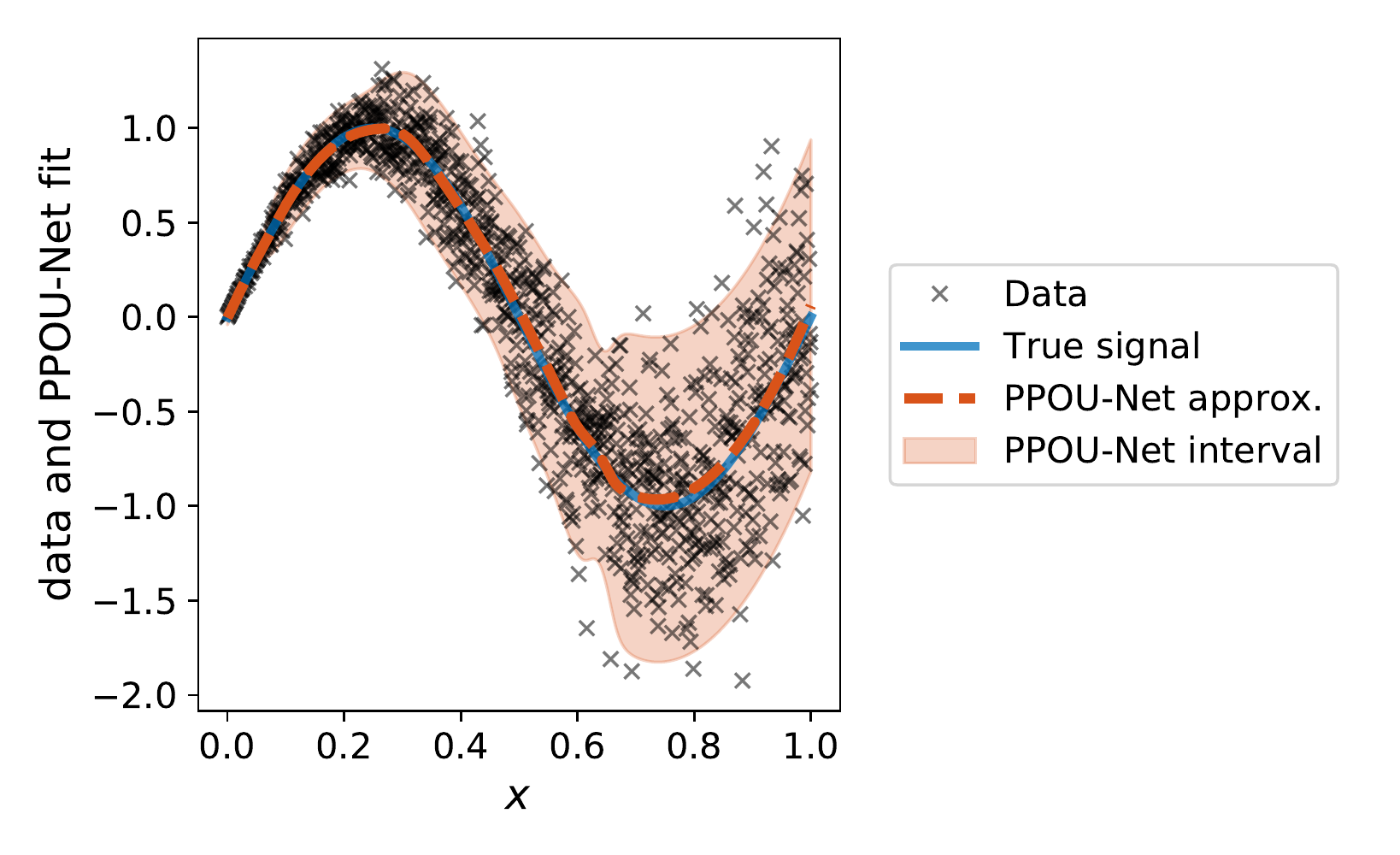}
\end{minipage}
\begin{minipage}{0.28\linewidth}
\centering
\includegraphics[width=1.01\columnwidth]{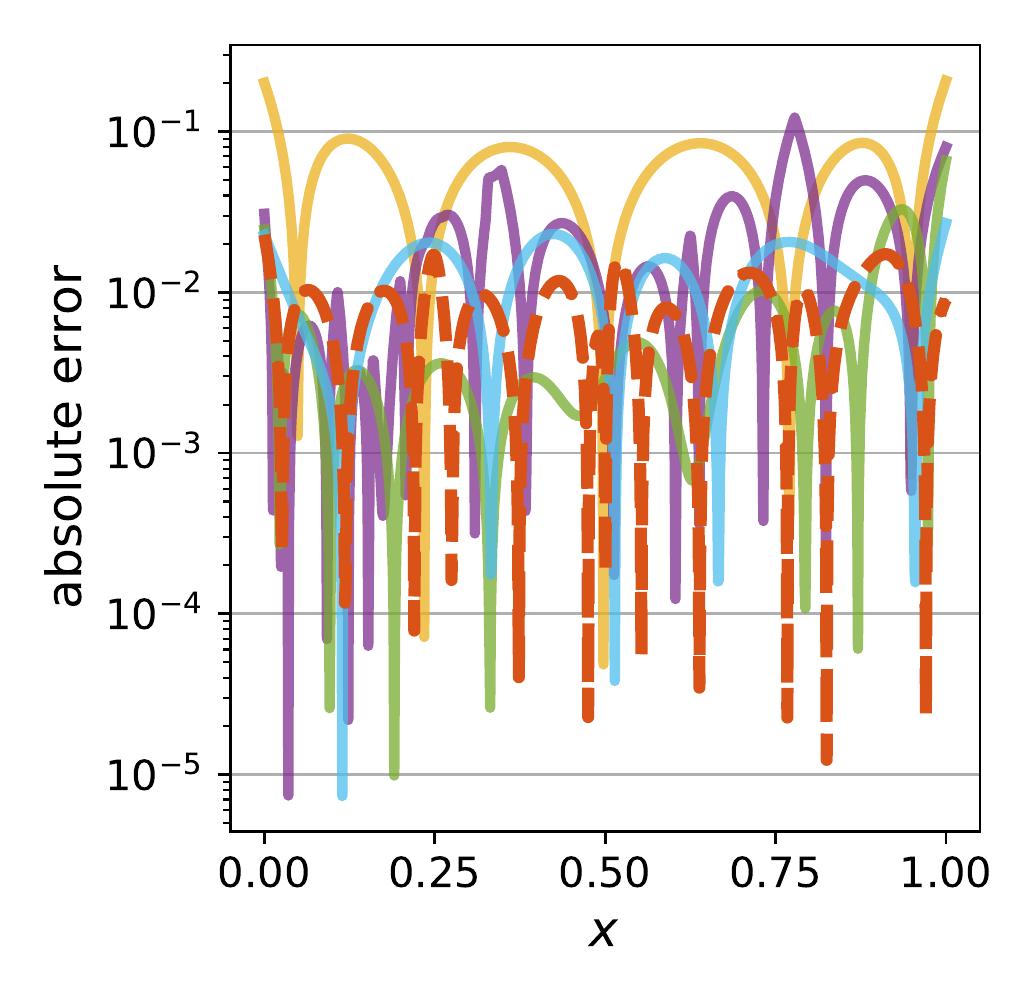}
\end{minipage}
\begin{minipage}{0.28\linewidth}
\centering
\includegraphics[width=1.01\columnwidth]{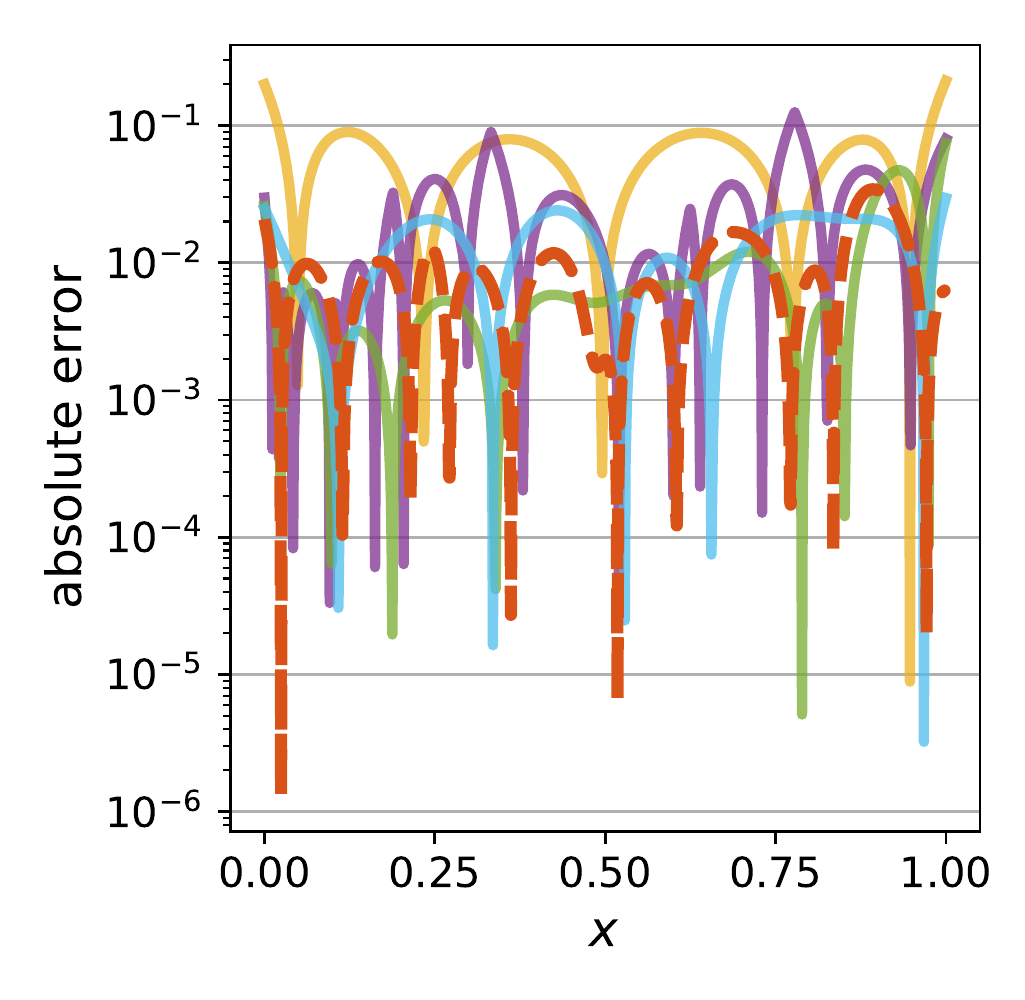}
\end{minipage}~
\begin{minipage}{0.44\linewidth}
\centering
\includegraphics[width=1.0\columnwidth]{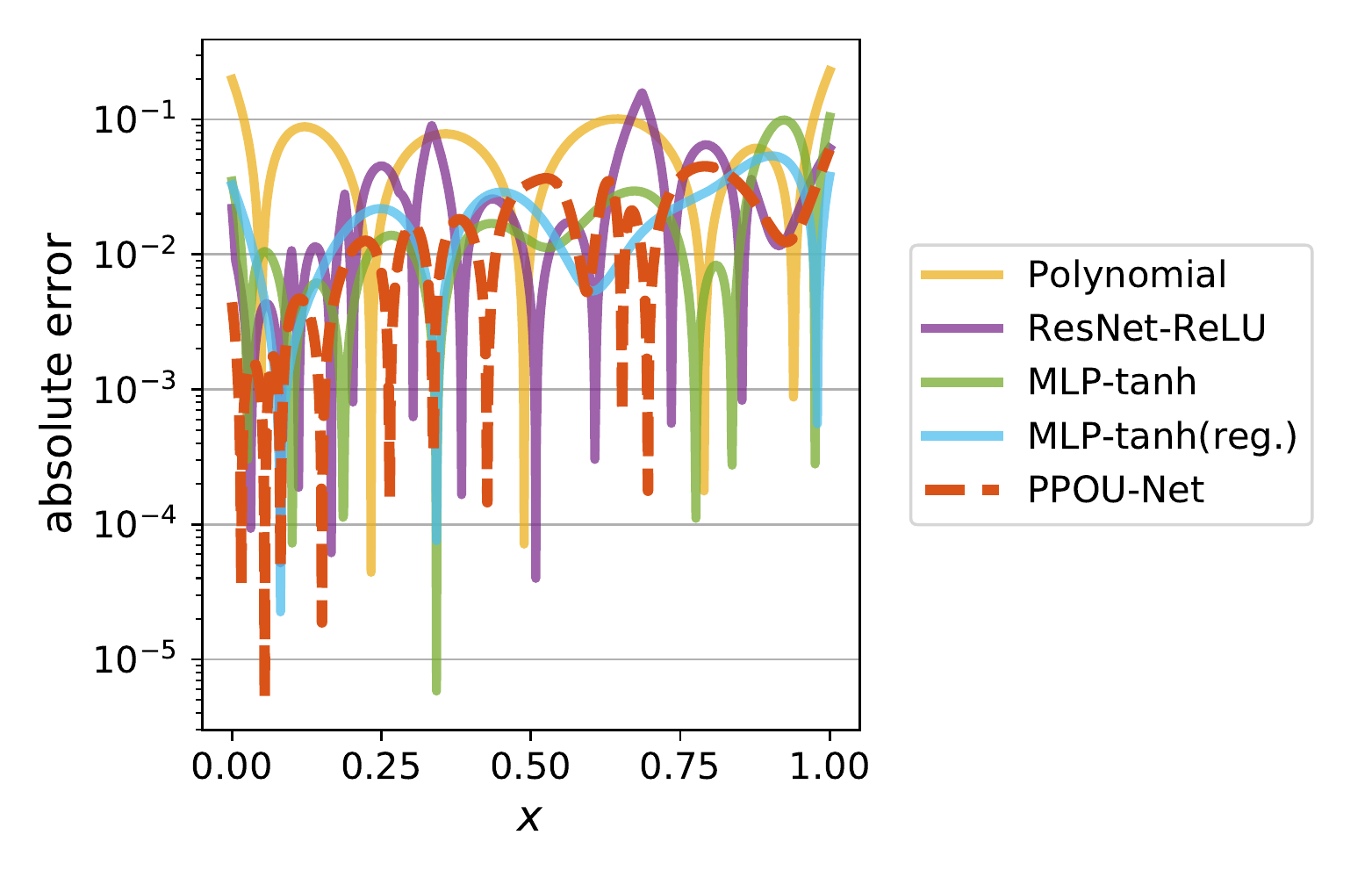}
\end{minipage}

\caption{\label{fig:noise} Examples of PPOU-Net approximations trained with data of various noise levels, compared to cubic expansions in Chebyshev polynomials, \added{ReLU-based ResNets, and tanh-based MLPs}. \added{The PPOU-Net} approximation and the 95\% confidence interval are plotted against the noisy data and the true signal on the top row; the absolute errors of all approximation methods are plotted on the bottom row. \added{At all noise levels, the PPOU-Net achieves lower approximation error than the Chebyshev polynomial and the ResNet; on sub-intervals with higher noise, the PPOU-Net achieves lower approximation error than the MLPs. In addition,} the PPOU-Net provides an accurate estimation of the heterogeneous uncertainty.}
\end{figure}
\clearpage

\begin{figure}[th!]
\centering
\includegraphics[width=0.8\columnwidth]{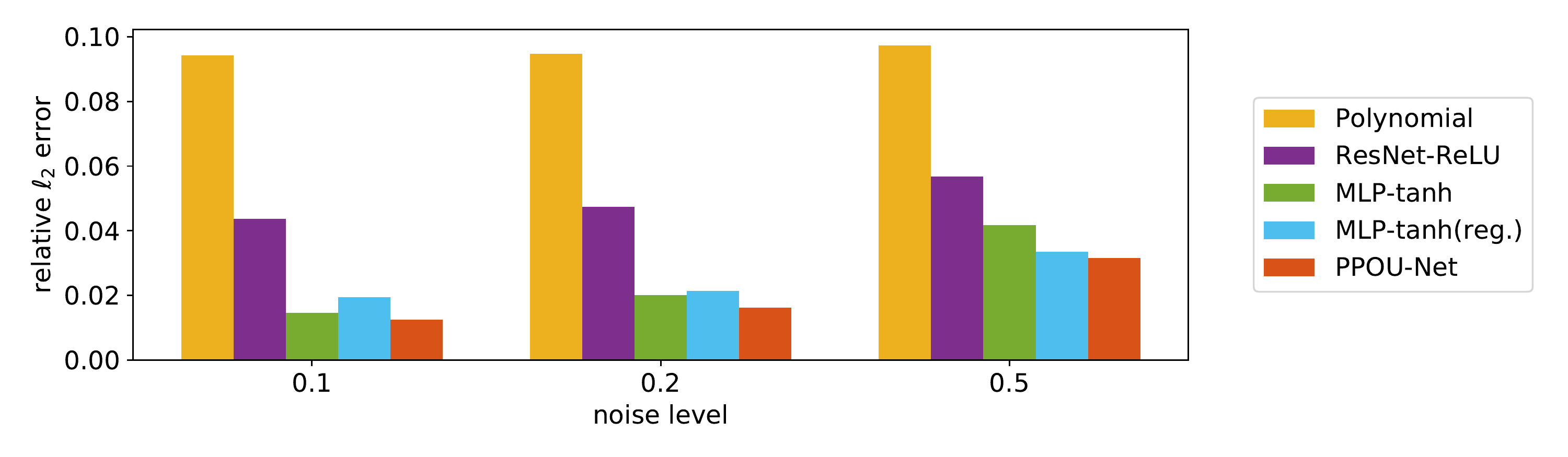}
\caption{\label{fig:noise_barplot} \added{A comparison of relative $\ell_2$ error of PPOU-Net against baseline methods, where the reference is the true signal. The PPOU-Net exhibits robust performance and achieves lowest error at all noise levels.}}
\end{figure}
\added{We consider baseline methods from both classical and deep learning literature, including (i) cubic expansion in Chebyshev polynomials; (ii) ResNet with ReLU activation functions; (iii) MLP with tanh activation functions; and (iv) MLP with tanh activation functions and $\ell_2$ regularization.} In \Cref{fig:noise}, we observe that the 95\% confidence intervals capture most of the noisy data, thus the PPOU-Net accurately predicts the noise levels in the data. In addition, the PPOU-Net achieves an order of magnitude lower error than the baseline cubic polynomial fit \added{and the ReLU-based ResNet} as we compare point estimations against the true signal. \added{On sub-intervals with higher noise levels, the PPOU-Net achieves lower approximation error than the tanh-based MLP without regularization. The addition of $\ell_2$ regularization improves the point estimation of MLP at high noise levels. Overall, as we show in \Cref{fig:noise_barplot},  the PPOU-Net exhibits robust performance and achieves the lowest $\ell_2$ error at all noise levels.  }

\subsection{Dimensionality reduction with non-smooth functions}\label{subsec:numerical_swiss_and_trefoil}
In this section, we explore the power of PPOU-Nets in approximating non-smooth functions by piecewise polynomials and in performing model order reduction to data manifolds with complex geometries. 
\subsubsection{Trefoil knot}
\begin{figure}[h!]
    \centering
    \begin{minipage}{0.33\linewidth}
    \centering
    data ~~~~
    \end{minipage} ~~~~~
    \begin{minipage}{0.32\linewidth}
    \centering
    ~~~~ partitions 
    \end{minipage}
    \begin{minipage}{0.32\linewidth}
    \centering
    ~~~~ approximation 
    \end{minipage}
    \begin{minipage}{0.33\linewidth}
    \centering
    \includegraphics[width=1.0\columnwidth]{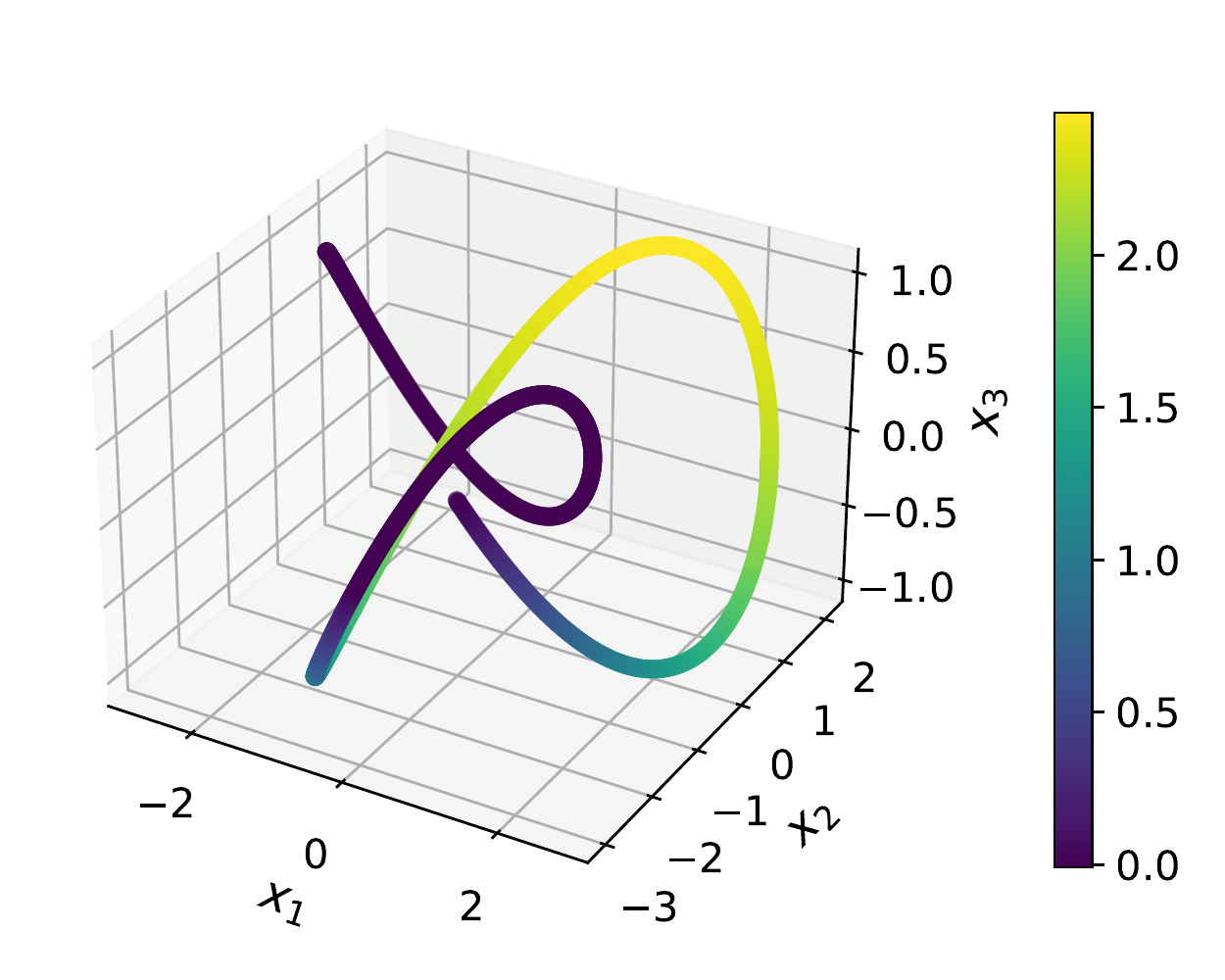}
    \end{minipage}
    \begin{minipage}{0.33\linewidth}
    ~~~\includegraphics[width=0.98\columnwidth]{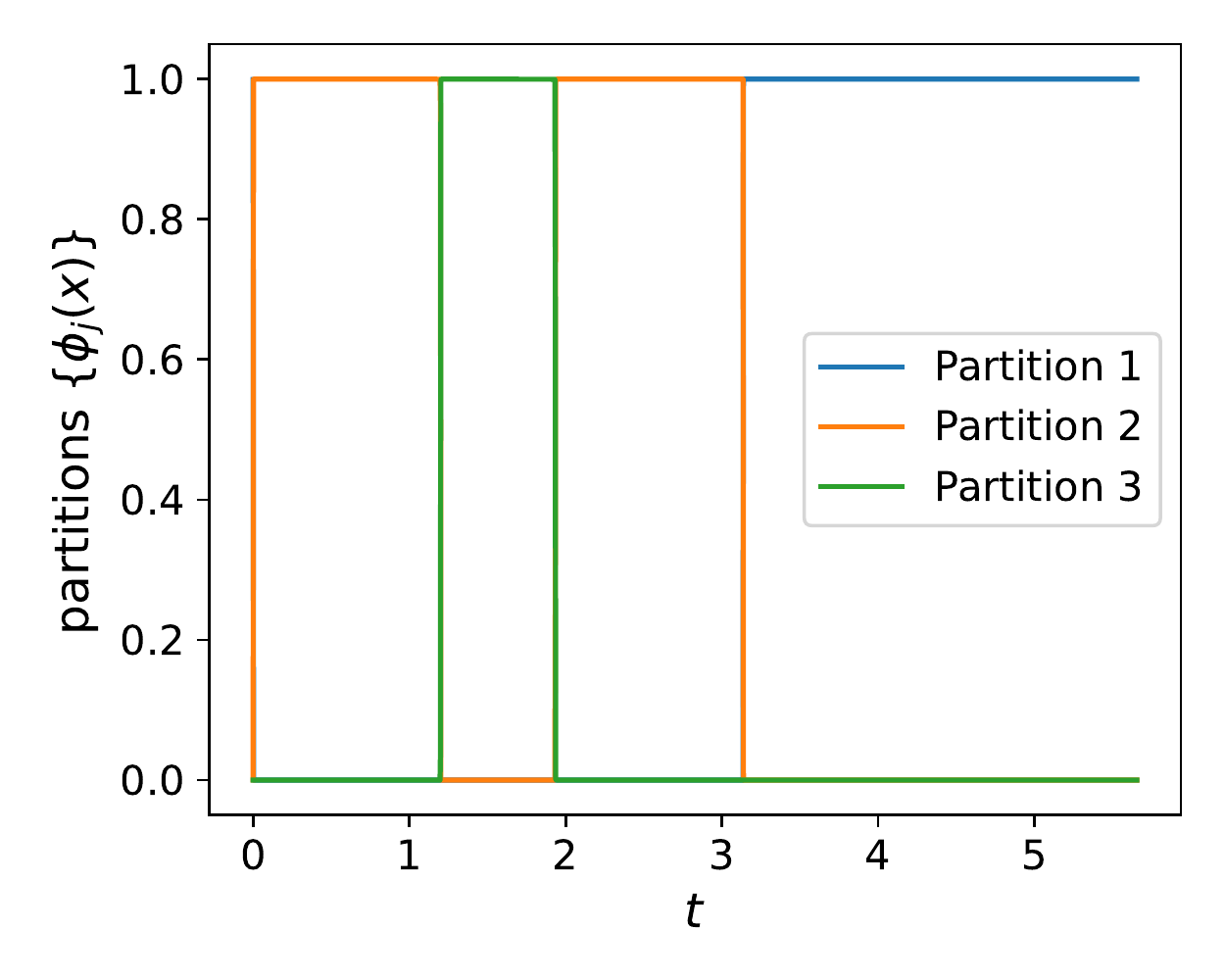}
    \end{minipage}
    \begin{minipage}{0.33\linewidth}
    \centering
    ~~\includegraphics[width=0.98\columnwidth]{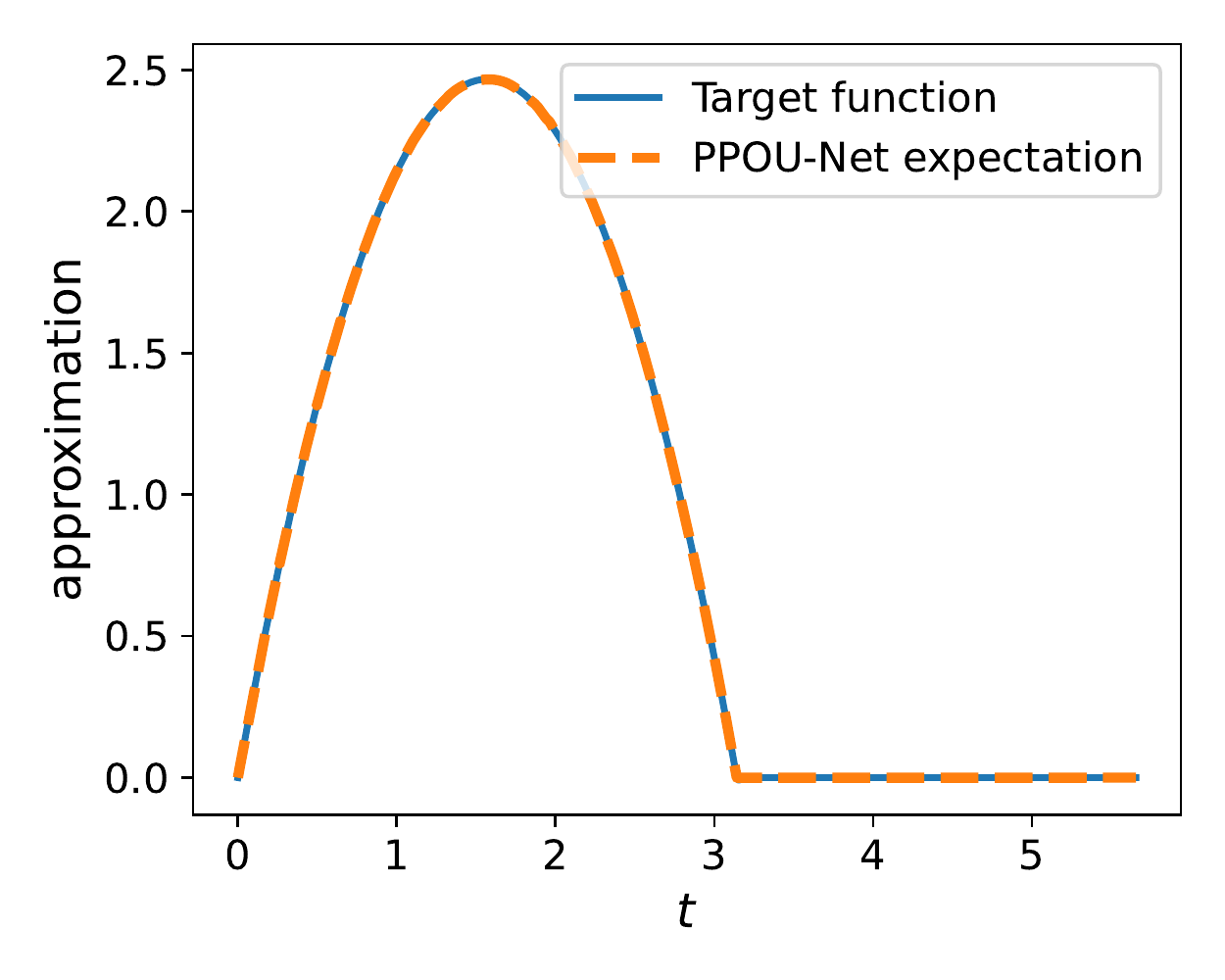}
    \end{minipage} 
    \begin{minipage}{0.33\linewidth}
    \hspace{4.12cm}
    \end{minipage} ~~~
    \begin{minipage}{0.32\linewidth}
    \centering
    ~~~~ error 
    \end{minipage}
    \begin{minipage}{0.32\linewidth}
    \centering
    ~~~~ error distribution
    \end{minipage}
    \begin{minipage}{0.33\linewidth}
    \hspace{-0.5cm}\centering
    \end{minipage}
    \begin{minipage}{0.33\linewidth}
    \centering
    \includegraphics[width=1.015\columnwidth]{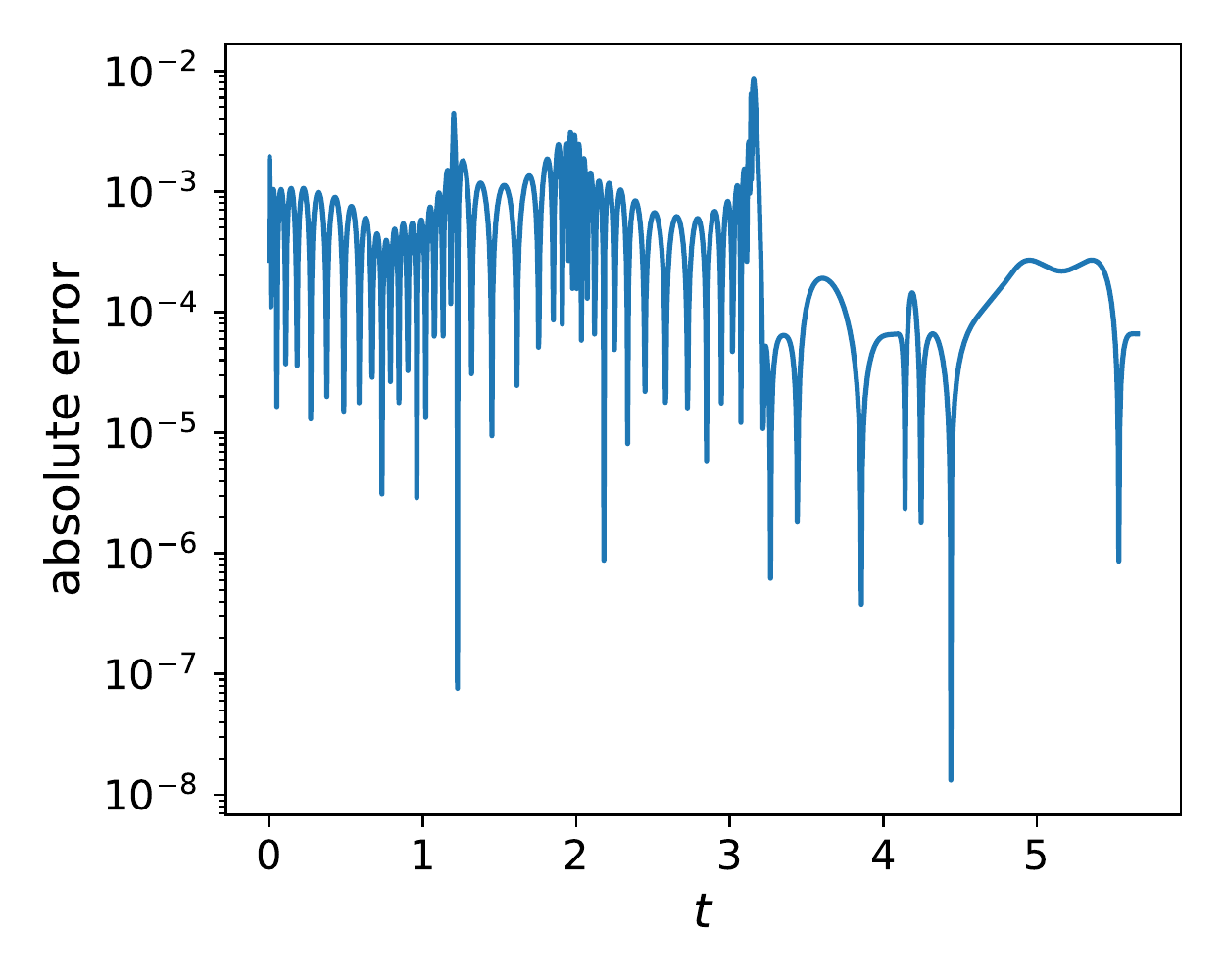}
    \end{minipage}
    \begin{minipage}{0.33\linewidth}
    \centering
    \includegraphics[width=1.005\columnwidth]{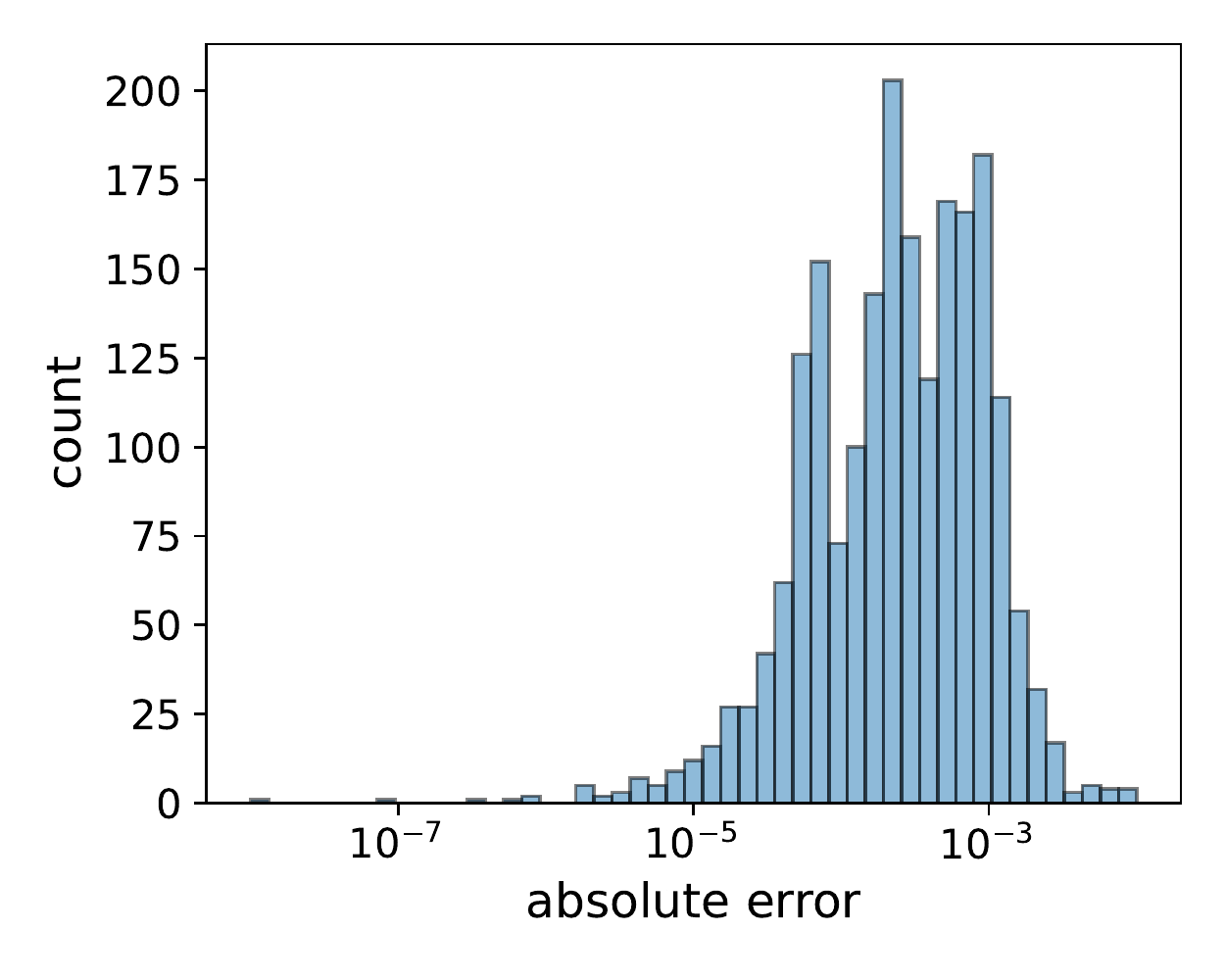}
    \end{minipage}
    \caption{\label{fig:trefoil_90} Example of a PPOU-Net that encodes data from an open 3D trefoil knot (top left) into a 1D latent space. As functions of $t$, we plot the weights of the partitions (top center) and compare the target function to the model prediction (top right). We visualize the absolute errors (bottom center) in the PPOU-Net approximation and include a histogram of the errors (bottom right). The PPOU-Net identifies the unsmoothness in the training data and partitions the data manifold at the boundary so that the data within each partition can be closely represented by a polynomial, providing an accurate approximation across the domain. }
\end{figure}

We first generate scattered data on a trefoil knot from the parameterization (see the first plot in \Cref{fig:trefoil_90}):
\begin{equation*} \label{eq:trefoil}
\begin{aligned}
x  = \, ( \sin t + 2 \sin 2 t, \ \cos t-2 \cos 2 t, \ -\sin 3 t ) 
, \quad
 y   =  \pi t - t^2 +  \frac{t^2 - \pi t}{ 1 +  \exp{ (- 100 \, t ) } } 
\end{aligned}
\end{equation*}
where $t$ is evaluated at $N=2048$ evenly spaced points on $[0, 1.8\pi]$.  

We train a PPOU-Net that maps the data onto a 1D latent space and approximates the target function using quadratic functions. \added{The model is trained with input $x\in \mathbb{R}^3$ and output $y\in \mathbb{R}$.} The PPOU-Net contains an encoder with depth 4 and width 16, a Box-initialized ResNet classifier with depth 12 and width 8, and 8 partitions with quadratic functions.  We plot the training metrics, the model approximation, the weights of the partitions $\{\phi_j(\mathbf{x};\theta_\phi)\}_j$ and the absolute errors in the PPOU-Net approximation as a function of $t$ in \Cref{fig:trefoil_90}, as well as a histogram of the absolute errors.  We observe that the model has relatively high approximation errors on the ends of the trefoil curve, and on the sub-regions where the target function is non-smooth as a function of $t$, while achieving an overall accurate approximation.

\subsubsection{Swissroll}
\begin{figure}[htbp!]
\centering
    \begin{minipage}{0.33\linewidth}
    \centering
    data~~~~~~~~~
    \end{minipage}
    \begin{minipage}{0.33\linewidth}
    \centering
    target
    \end{minipage}
    \begin{minipage}{0.33\linewidth}
    \centering
    partitions
    \end{minipage}
    \begin{minipage}{0.33\linewidth}
    \centering
    \includegraphics[width=1.01\columnwidth, rviewport=0.02 0 0.98 1, clip]
    {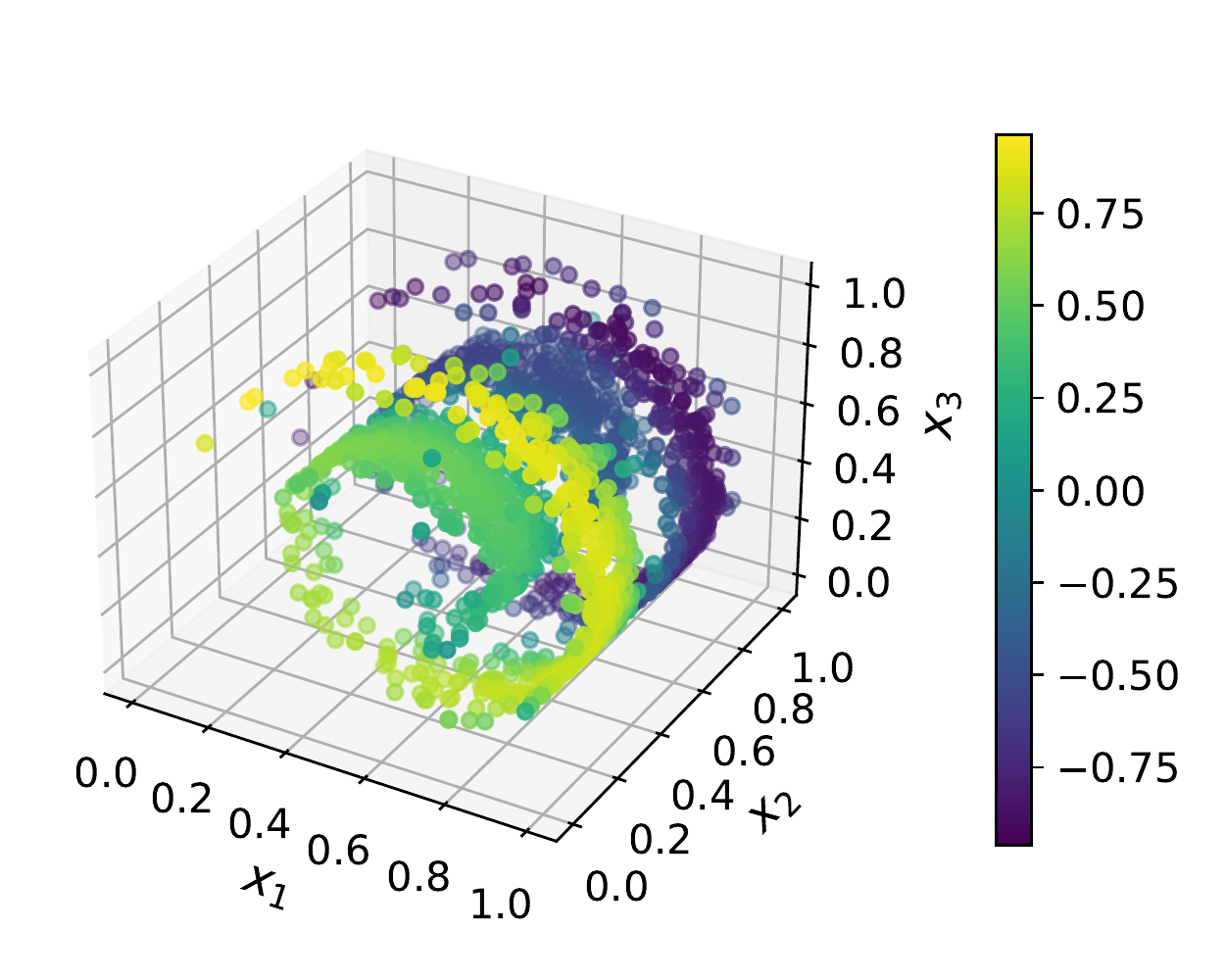}
    \end{minipage}\hspace{0.0cm}
    \begin{minipage}{0.33\linewidth}
    \centering
    \includegraphics[width=1.04\columnwidth]{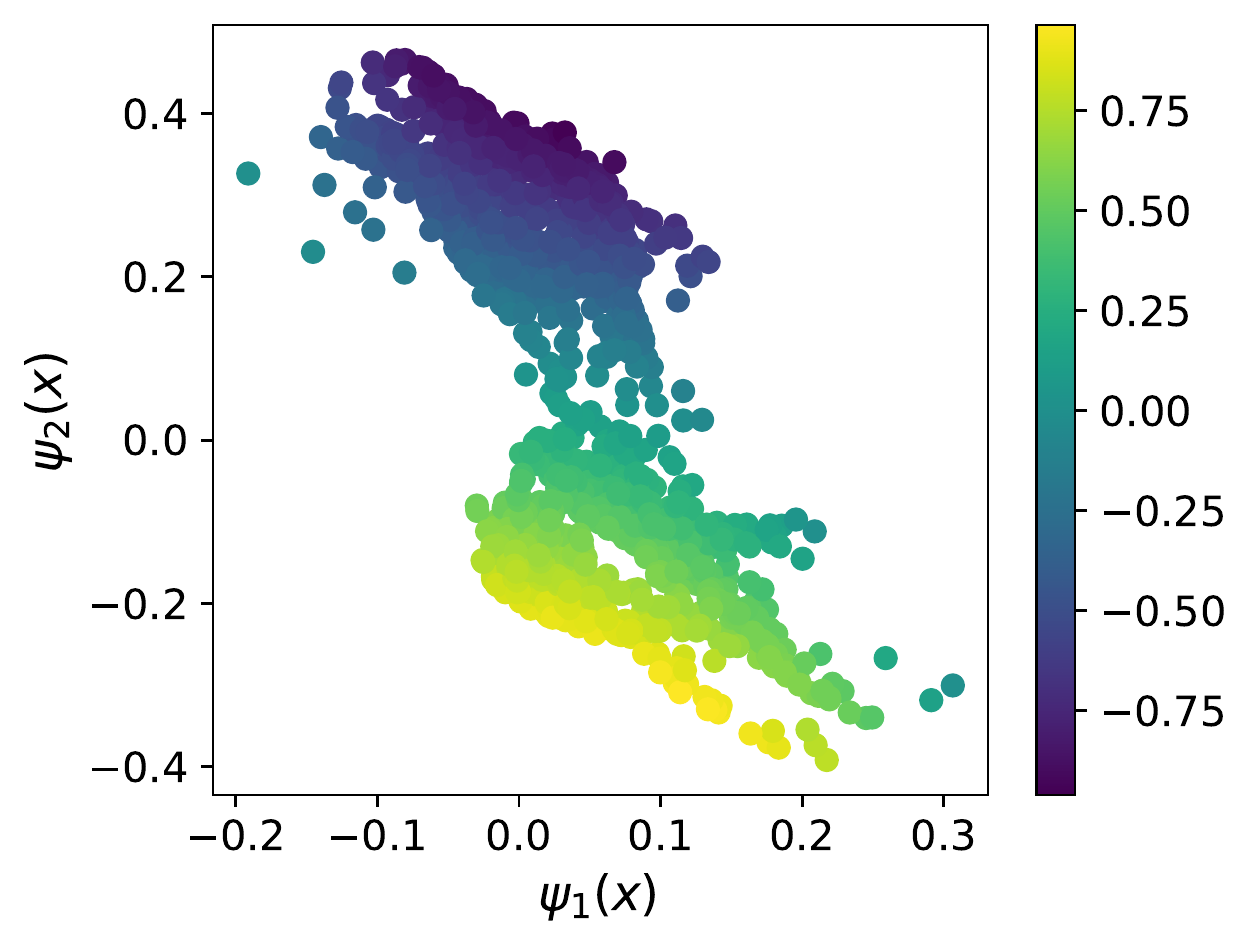}
    \end{minipage}\hspace{0.00cm}
    \begin{minipage}{0.33\linewidth}
    \centering
    ~\includegraphics[width=0.99\columnwidth]{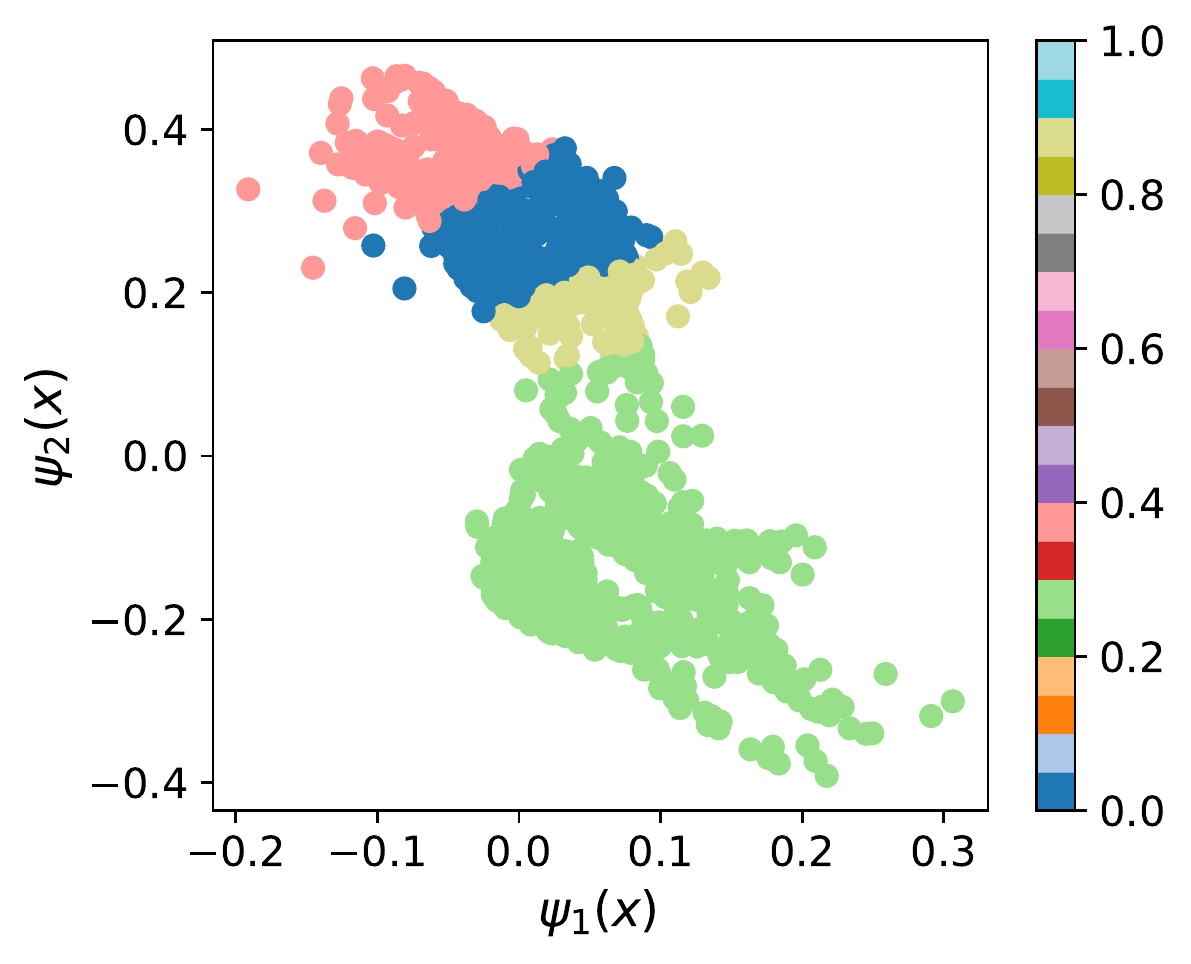}
    \end{minipage}\vspace{0.1cm}
    \begin{minipage}{0.33\linewidth}
    \centering \hspace{3.4cm}
    \end{minipage} \hspace{-0.06cm}
    \begin{minipage}{0.33\linewidth}
    \centering
    approximation
    \end{minipage}
    \begin{minipage}{0.33\linewidth}
    \centering
    error
    \end{minipage}\vspace{0.1cm}
    \begin{minipage}{0.33\linewidth}
    \hspace{3.2cm}
    \end{minipage}
    \begin{minipage}{0.33\linewidth}
    \centering
    ~\includegraphics[width=1.04\columnwidth]{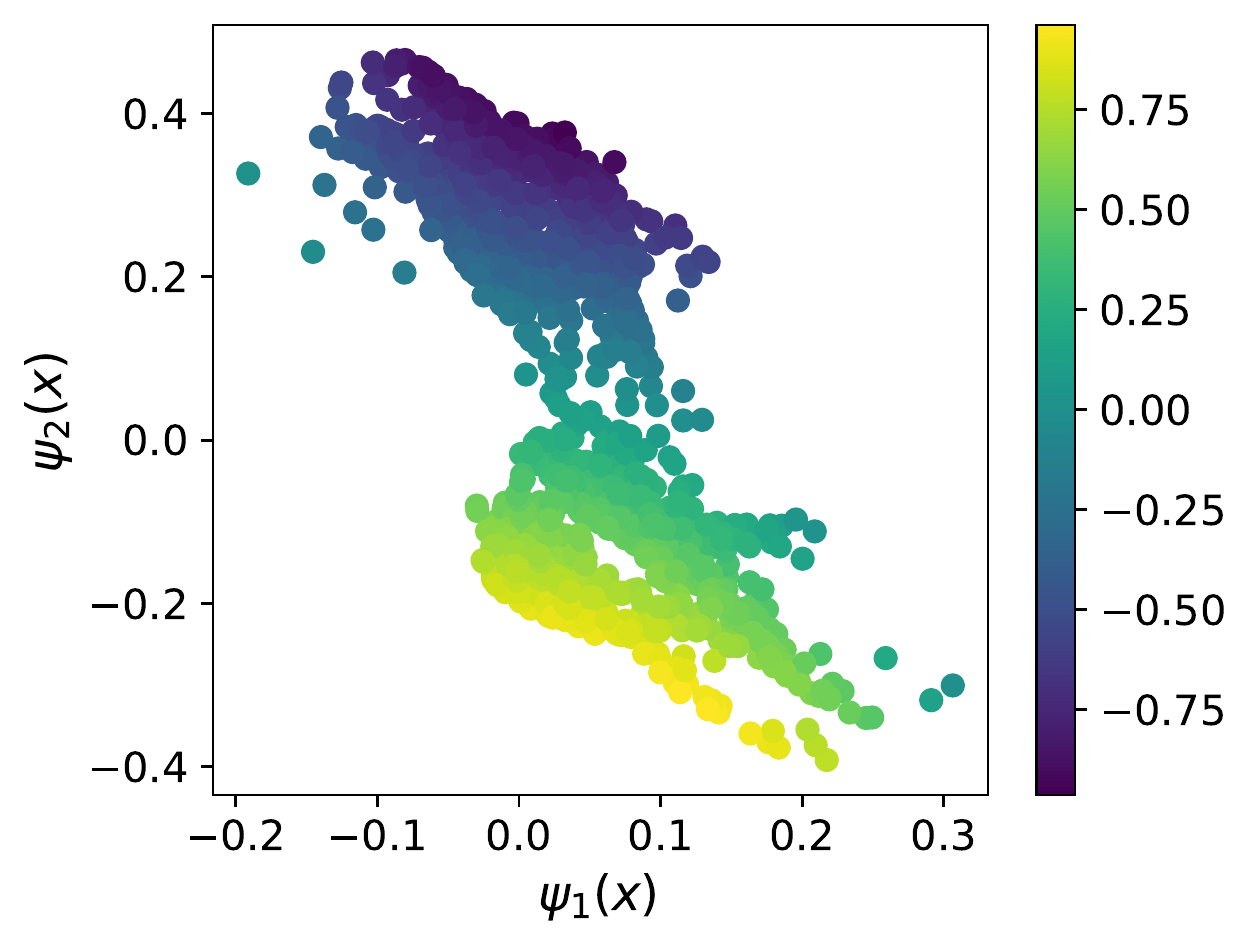}
    \end{minipage}
    \begin{minipage}{0.33\linewidth}
    \centering
    ~~\includegraphics[width=1.04\columnwidth]{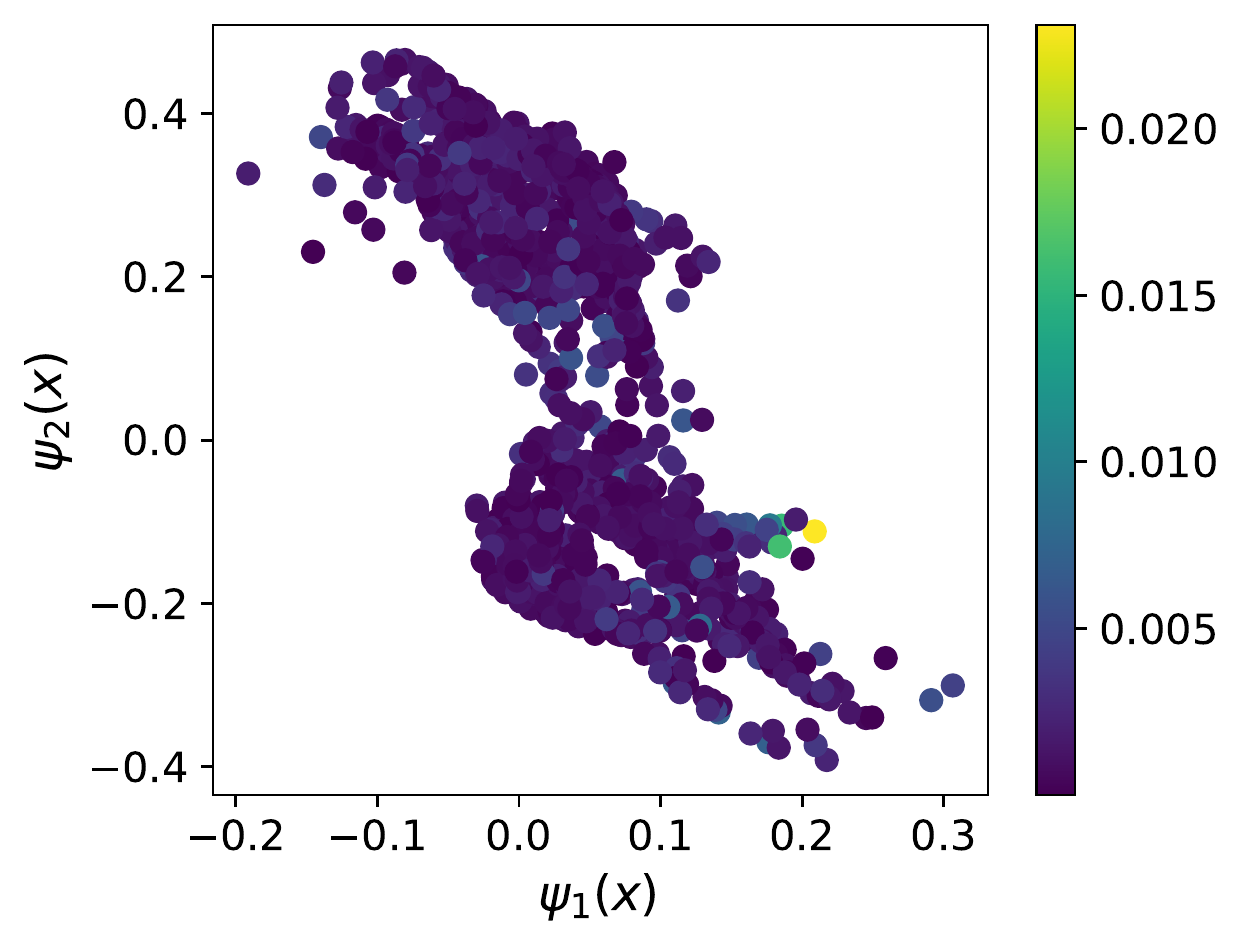}
    \end{minipage}
\caption{\label{fig:swiss} Example of a PPOU-Net trained with data defined on a 3D swiss roll. The target function is non-smooth (top left). We visualize the training results as functions of the encoded data $\psi(\bf{x})\in \mathbb{R}^2$: the dominating partitions (plotted as the dominating partition indices normalized by the total number of partitions so that each color represents a distinct partition, top right), the target function (top center), the PPOU-Net approximation (bottom center), and the absolute error (bottom right). The PPOU-Net projects the complex data manifold onto a 2D plane where training examples are partitioned based on similarity and smoothness. The resulting mixture of 2D quadratic functions provides an accurate approximation of the target function. }
\end{figure}

In this experiment, we consider the non-smooth function defined on a Swiss roll manifold in 3D \cite{surendran2004swiss} \added{(see top left plot in \Cref{fig:swiss})}, parameterized by $t\in \mathbb{R}^2$:
\begin{equation*} \label{eq:swissroll}
\begin{aligned}
x  =\, ( t_1 \cos t_1, \ t_2 , \ t_1 \sin t_1 ) 
, \quad
y  = \, \tilde {t}_1^{\  1/2} \sin (2 \pi \tilde{t}_2 )  
\end{aligned}
\end{equation*}
where $\tilde {t}$ is obtained from normalizing $t$ to the unit square $[0,1]^2$. We train a PPOU-Net that projects the target function onto a 2D latent space and approximates the target function using quadratic functions of the latent variables. \added{The task is to learn the mapping from input $x\in \mathbb{R}^3$ to output $y\in \mathbb{R}$.}  The PPOU-Net contains an encoder with depth 4 and width 32, a Box-initialized ResNet classifier with depth 12 and width 8, and 8 partitions with quadratic functions. 

As shown in \Cref{fig:swiss}, the PPOU-Net model successfully simplifies the 3D geometry by finding a 2D latent space, and separates the data points into four partitions based on similarity and smoothness in the target function values. The PPOU-Net achieves an overall relative $\ell^2$ error on the order of  $10^{-3}$. The error tends to be higher where the model transitions between partitions and on the boundary of the latent manifold.

\begin{figure}[htbp!]
\centering
\includegraphics[width=0.85\columnwidth]{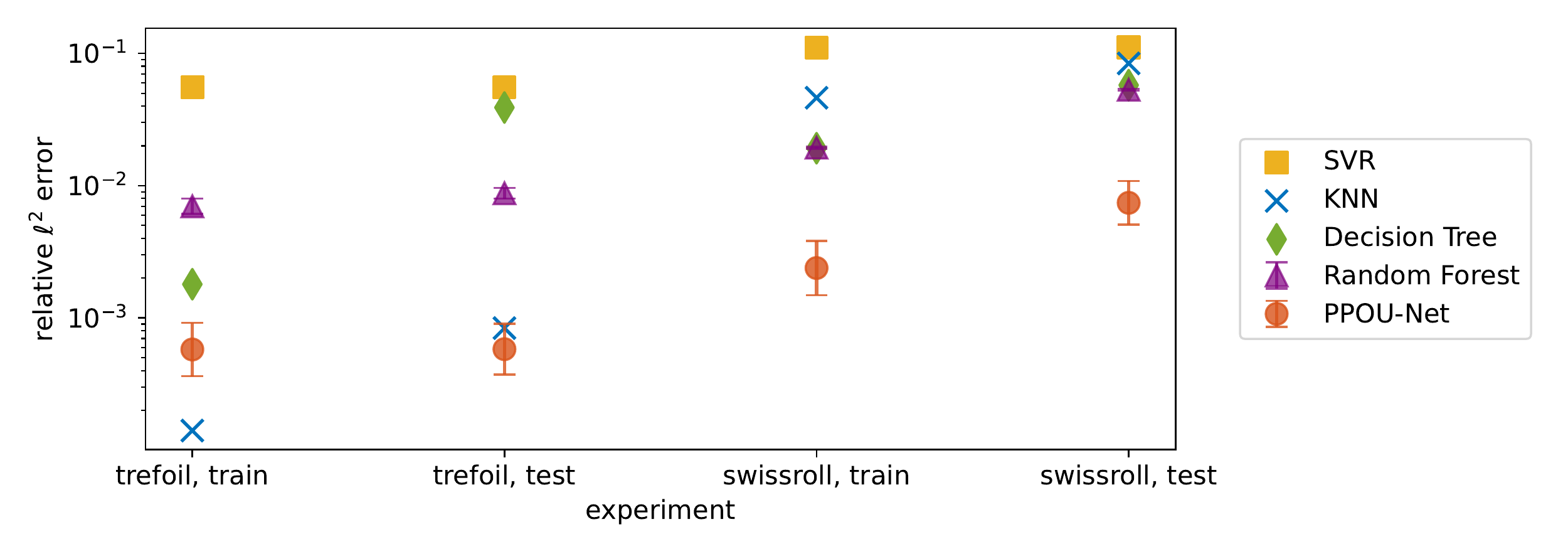}
\caption{\label{fig:baseline_regression} A comparison of training and testing relative $\ell^2$ error for PPOU-Net and baseline regression methods including support vector regression (SVR), K-nearest neighbors (KNN), decision tree, and random forest. The PPOU-Net has robust performance in simplifying complex geometries and achieves the lowest testing error for both datasets.}
\end{figure}

We compare the PPOU-Net to several common regression methods in machine learning: support vector regression (SVR), K-nearest neighbors (KNN), decision tree, \added{and random forest}. The training and testing errors for both datasets are summarized in \Cref{fig:baseline_regression}. The PPOU-Net consistently outperforms the baseline regression methods and exhibits high reliability.

\subsection{Encoding of high-dimensional structured data}\label{subsec:numerical_rings}

\begin{figure}[htbp]
\centering
\begin{minipage}{0.36\linewidth}\centering \vspace{0.28cm}
\includegraphics[width=\columnwidth,rviewport=0.05 0 1 0.92, clip]{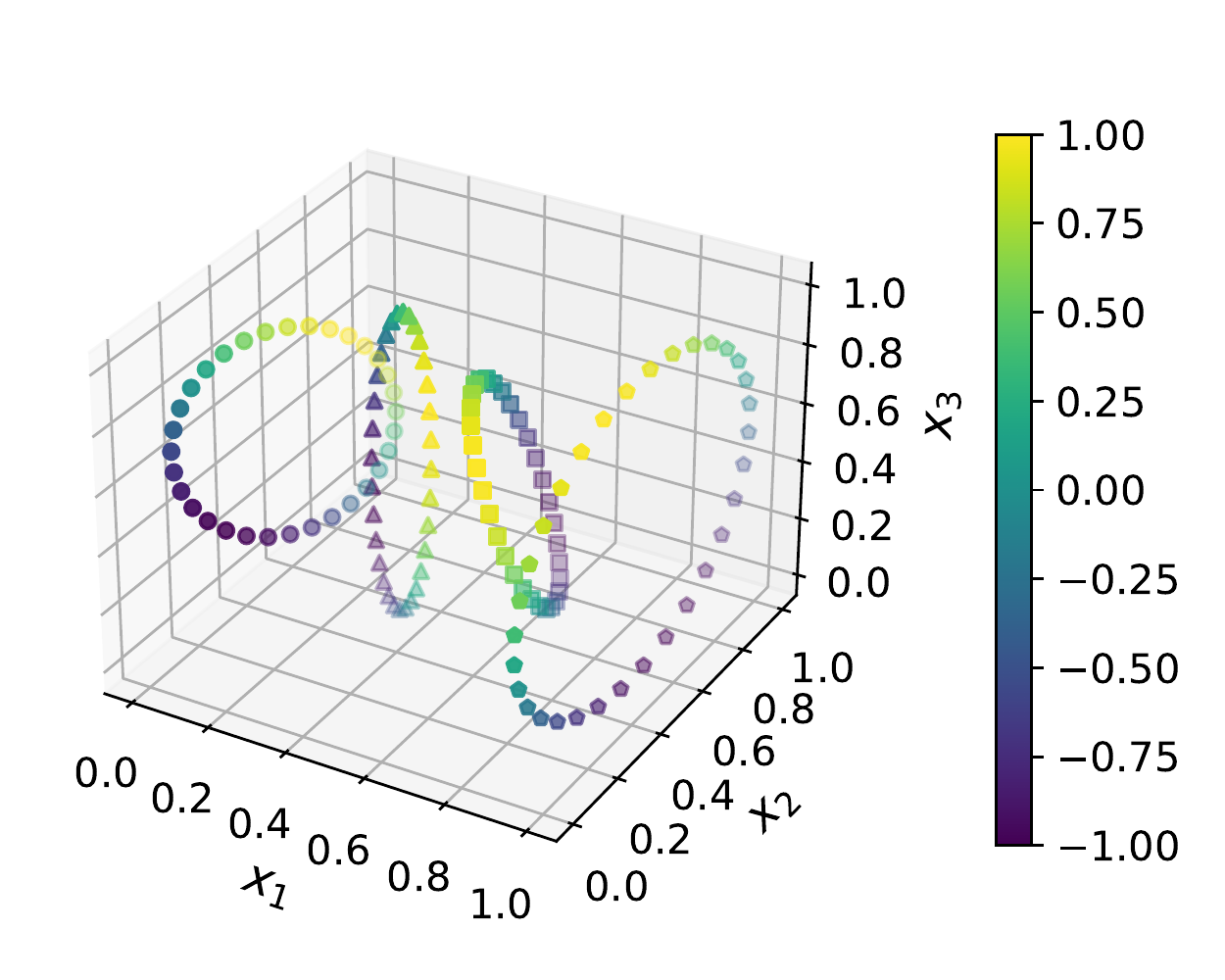}
\caption{\label{fig:rings_target} Scattered data simulated on four rings with random orientations in $\mathbf{R}^3$. The target function along each ring is sinusoidal with a random phase shift.
}
\end{minipage} 
~~~~~~
\begin{minipage}{0.61\linewidth}~~~~~~~~\vspace{0.1cm}
\includegraphics[width=1\columnwidth]{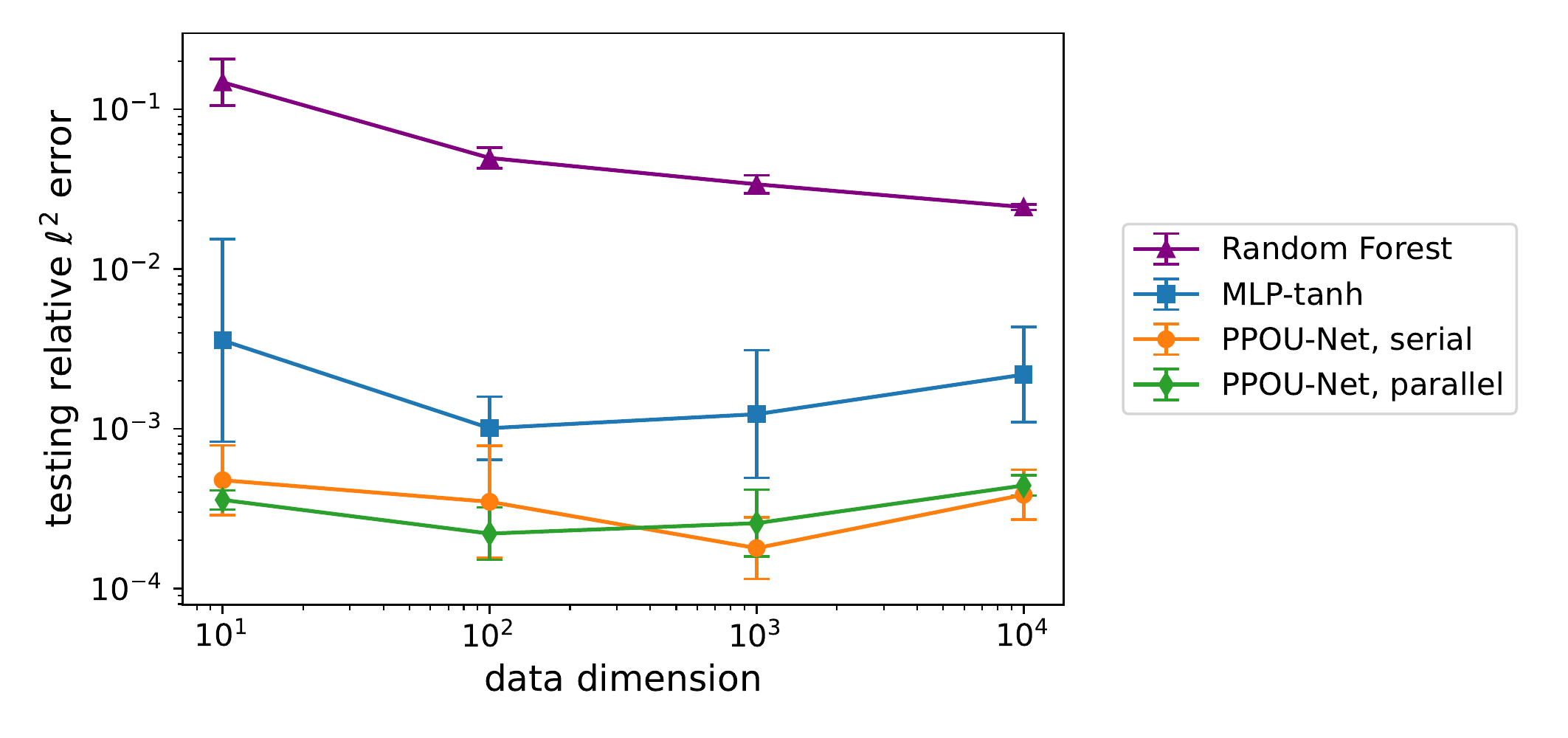}
\caption{\label{fig:rings_convergence} Relative $\ell^2$ errors achieved by PPOU-Nets for \added{input} dimensions $10$ to $10^4$. The error bars cover a 95\% confidence interval of the testing errors. Both PPOU-Net models consistently outperform the \added{two baseline methods}. The performance of the PPOU-Nets does not degrade as dimension increases. }
\end{minipage}
\end{figure}

For $d\in \{10, 10^2, 10^3, 10^4\}$, we generate a collection of $N_c=4$ rings with random orientations in a $d$-dimensional space. \added{We denote the spatial coordinates of the high-dimensional rings as $\mathbf{x}=(x_1, \ldots, x_d)\in \mathbb{R}^d$.} On each ring, we sample a periodic function (e.g., $y=\sin 2\pi t$ 
with $t$ evenly spaced on $[0,1]$) with a random phase shift as the target outputs. The centers of the rings are evenly spaced on the $x_1$-axis in $d$ dimensions (see \Cref{fig:rings_target}). 

\begin{figure}[htbp!]
\centering
\begin{minipage}[c]{0.245\textwidth}
\centering 
target~~
\end{minipage}
\begin{minipage}[c]{0.245\textwidth}
\centering
approximation~~
\end{minipage} 
\begin{minipage}[c]{0.245\textwidth}
\centering
error~~~~
\end{minipage}
\begin{minipage}[c]{0.245\textwidth}
\centering
partitions~~
\end{minipage}
\begin{minipage}[c]{0.245\textwidth}
\centering 
\includegraphics[width=1.01\columnwidth]{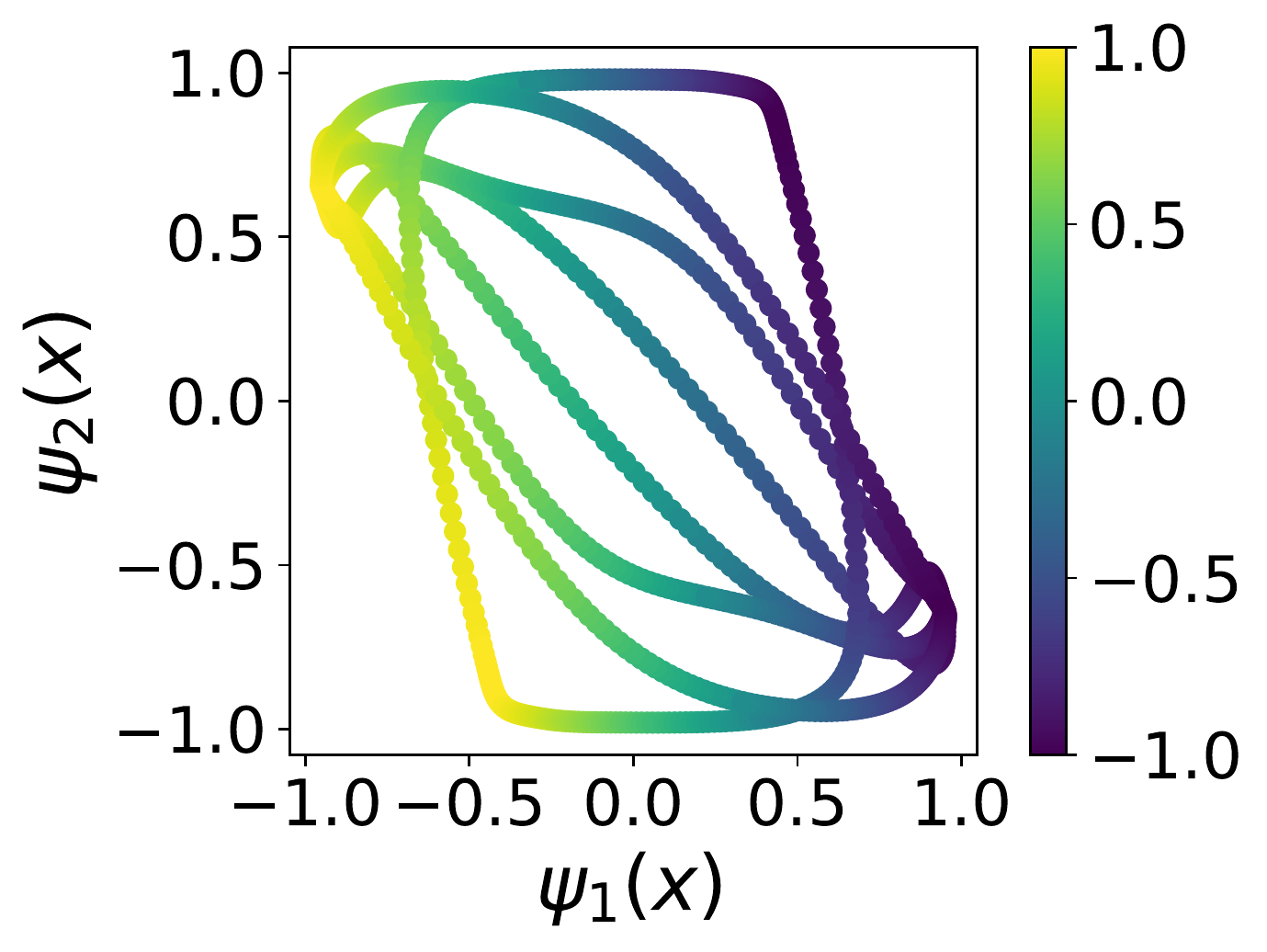}
\end{minipage}
\begin{minipage}[c]{0.245\textwidth}
\centering
\includegraphics[width=1.01\columnwidth]{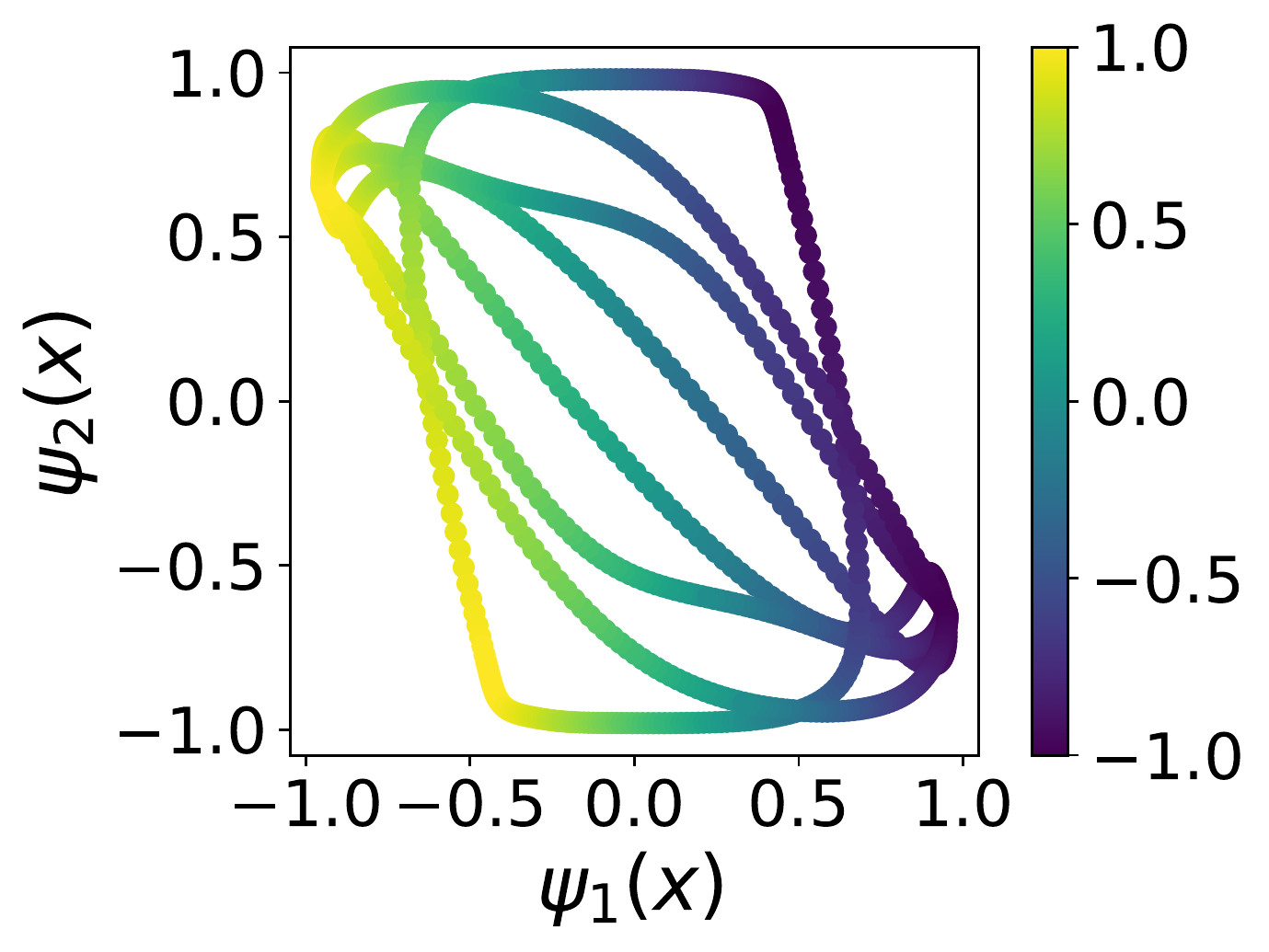}
\end{minipage} 
\begin{minipage}[c]{0.253 \textwidth}
\centering
\includegraphics[width=1.06\columnwidth]{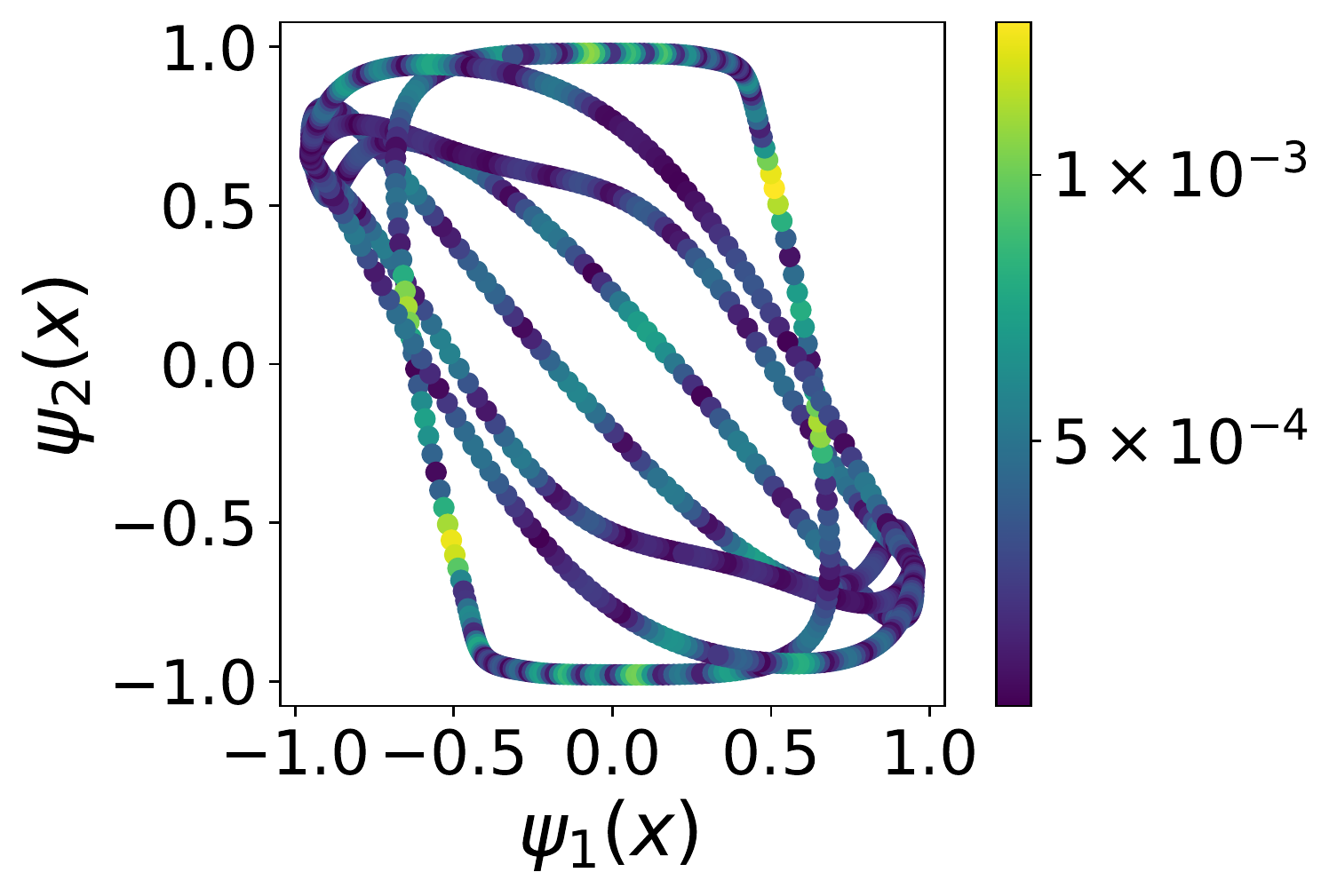}
\end{minipage}
\begin{minipage}[c]{0.245\textwidth}
\centering
\includegraphics[width=0.94\columnwidth]{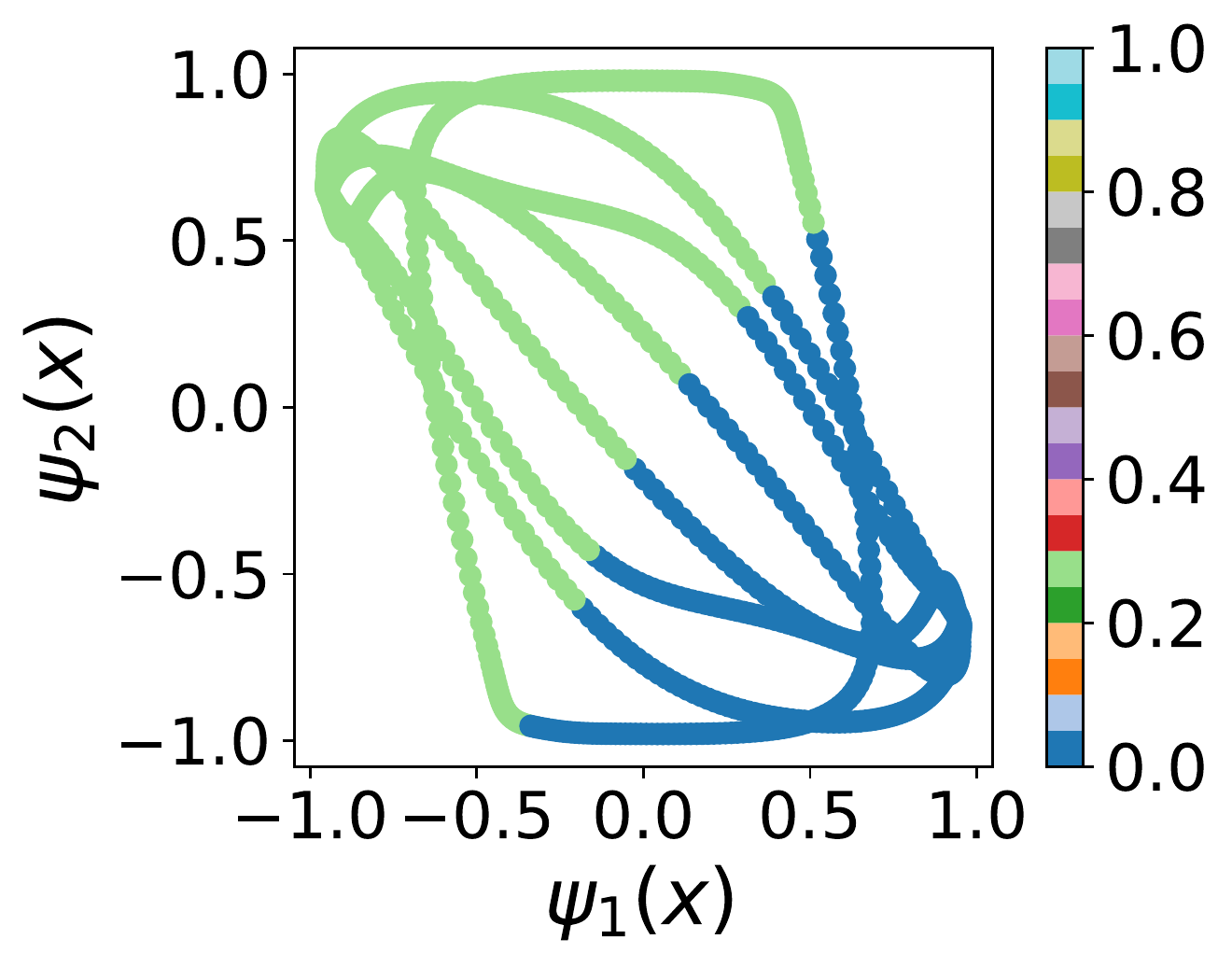}
\end{minipage}
\caption{\label{fig:rings_selected_visual} Training results for data residing on randomly oriented rings in $d=10^4$ dimensions. As functions of the 2D latent variable learned by the PPOU-Net, we plot the target function, the model approximation, the absolute error, and dominating partitions. The PPOU-Net successfully identifies a 2D latent space for the high-dimensional data where linear functions suffice to approximate the target function accurately.
}
\end{figure}

We use a PPOU-Net consisting of an encoder (with depth 4, width 16, and 2 latent dimensions), a Box-initialized ResNet classifier (with depth 4, width 8) and 4 partitions with linear functions to approximate the \added{mapping from $\mathbf{x}\in \mathbb{R}^d$ to $y\in \mathbb{R}$}. The magnitude of error by PPOU-Nets is similar across the input dimensions of $10$, $10^2$, $10^3$, and $10^4$ (see \Cref{fig:rings_convergence}) and consistently lower than that of the baseline MLP model. This is consistent with the theoretical results on the convergence of PPOU-Net, i.e., the convergence of error depends on the dimension of the latent space \cite{trask2021probabilistic}. For all data dimensions, the PPOU-Net is able to warp the high-dimensional input space to ``align'' the rings in a 2D latent space so that a simple combination of polynomials provides an accurate approximation of the target function (see \Cref{fig:rings_selected_visual} for an example when $d=10^4$).

\subsection{Application: learning high-dimensional cost functions of quantum approximate optimization}\label{subsec:numerical_quantum}

In this benchmark, we use the PPOU-Net model to approximate cost functions in the application of the quantum approximate optimization algorithm (QAOA). The general idea of a QAOA is to encode the cost function of an optimization problem as a Hamiltonian, so that the ground state of the Hamiltonian corresponds to the optimal solution of the original optimization problem. We use a variational method to find the ground state of the Hamiltonian, where the variational form of the QAOA is a layerized quantum circuit. \cite{Qiskit-Textbook} The input dimension of the cost function increases linearly with the number of layers. A cost function of higher input dimension corresponds to a quantum circuit with a larger space of design parameters, and allows for more accurate solutions to the optimization problem of interest. Thus, efficient representation of high-dimensional cost functions is highly desirable for the application.

\begin{figure}[htbp!]
    \centering
    \begin{minipage}[c]{0.32\textwidth}
    \centering
    2D target ~~~~~~ \\
    ~~\includegraphics[width=1.02\columnwidth, rviewport=0 0 1 0.95, clip]{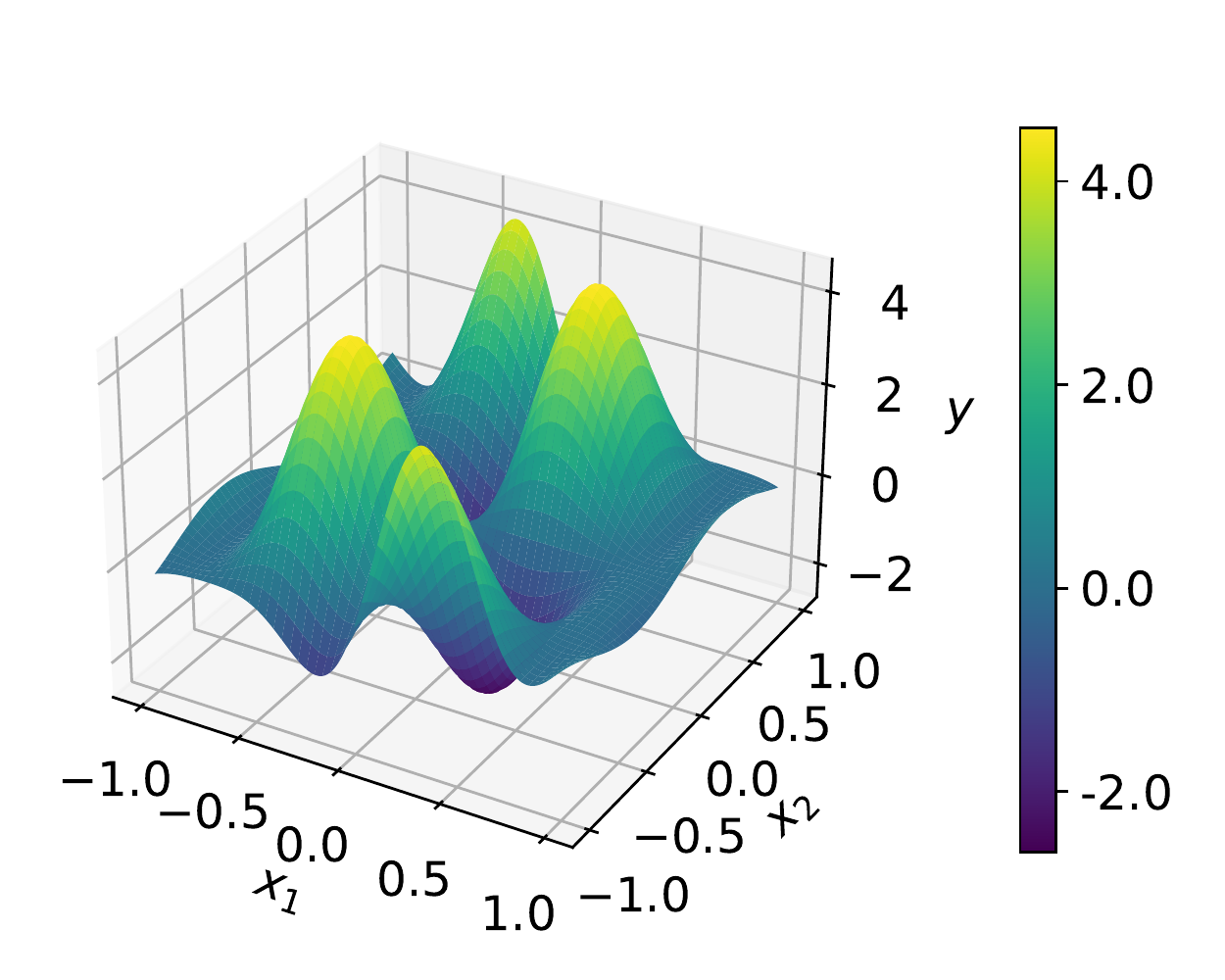}
    \end{minipage}
    \begin{minipage}[c]{0.32\textwidth}
    \centering
    PPOU-Net mean\\ \vspace{0.2cm}
    \includegraphics[width=\columnwidth]{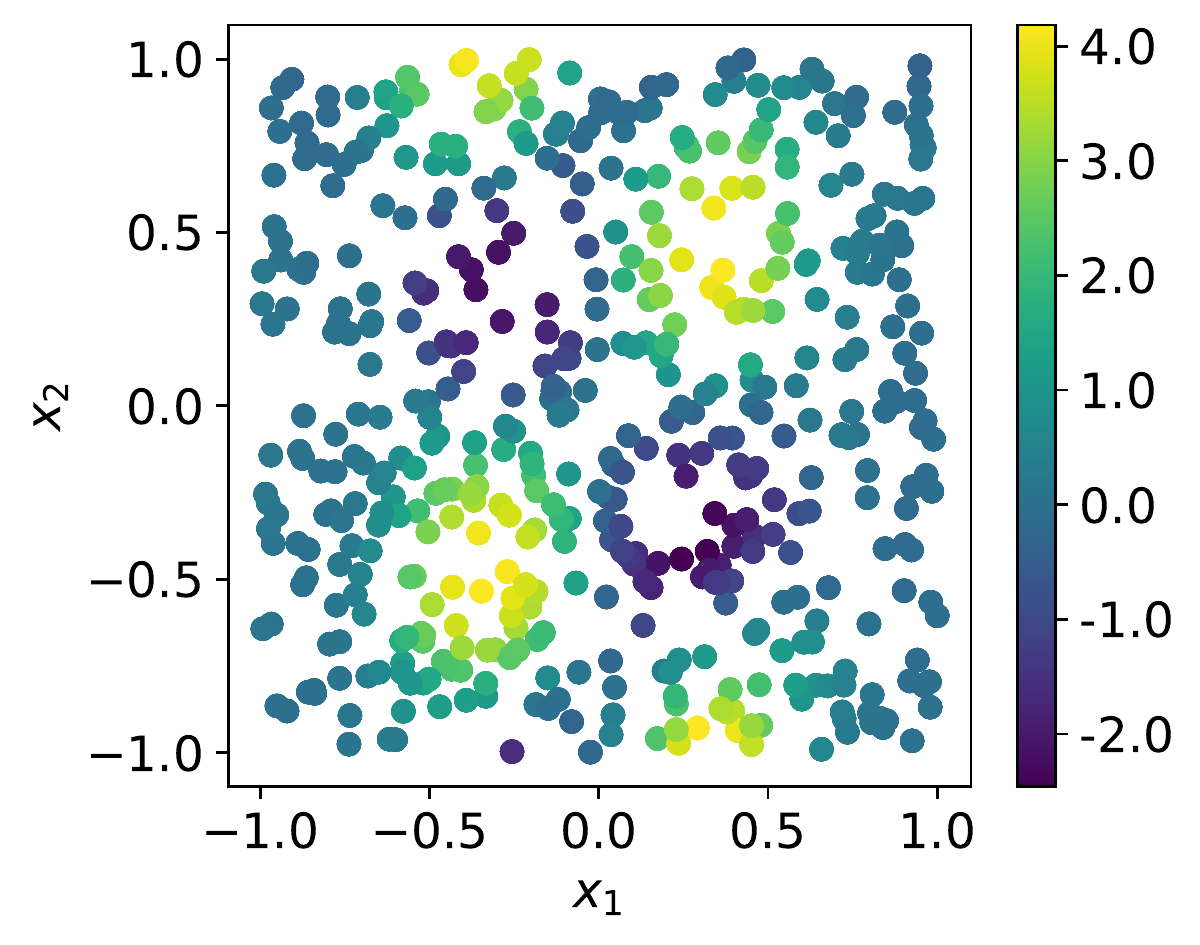}
    \end{minipage}
    \begin{minipage}[c]{0.32\textwidth}
    \centering
    empirical mean\\ \vspace{0.2cm}
    \includegraphics[width=\columnwidth]{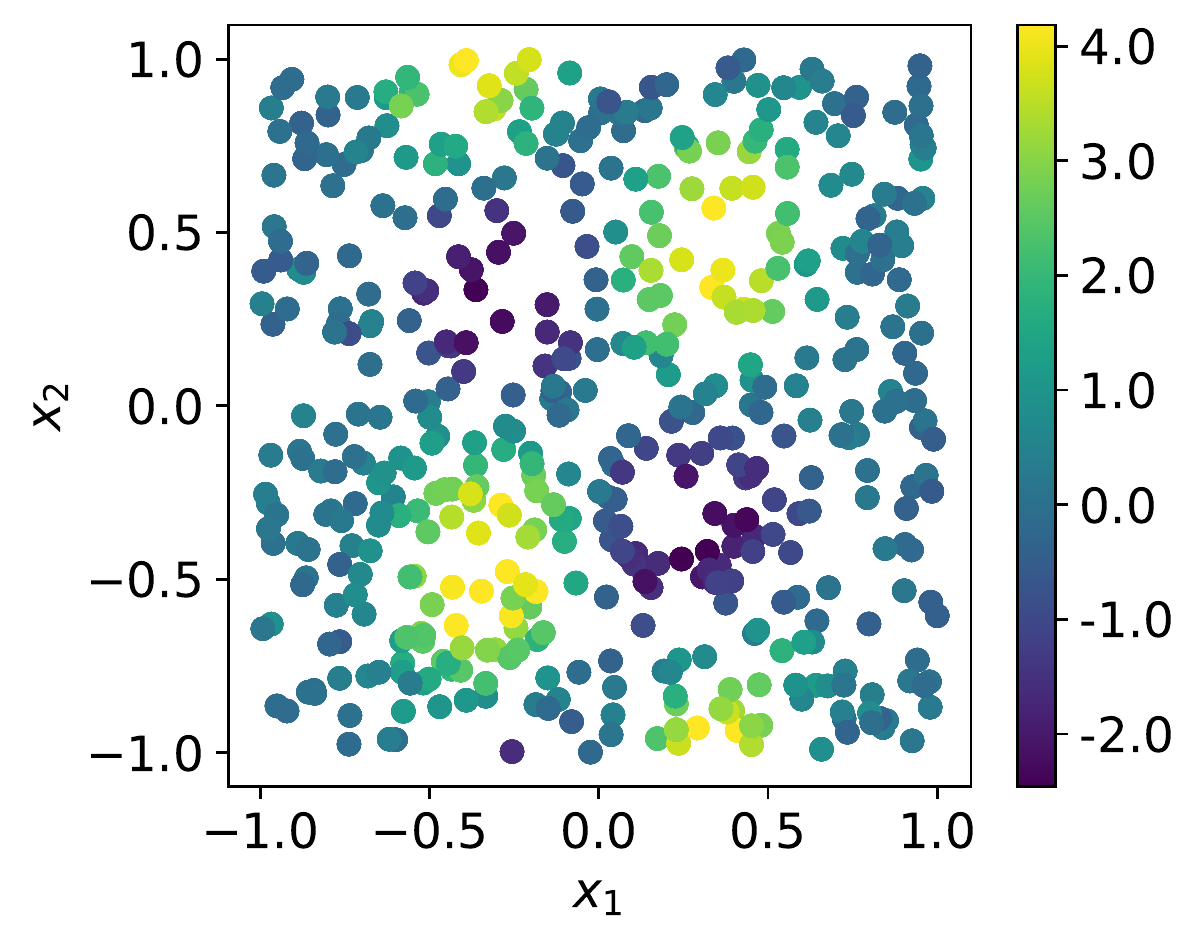}
    \end{minipage}\\ \vspace{0.2cm}
    
    \begin{minipage}[c]{0.32\textwidth}
    \centering
    partitions \\  \vspace{0.02cm}
    \includegraphics[width=\columnwidth]{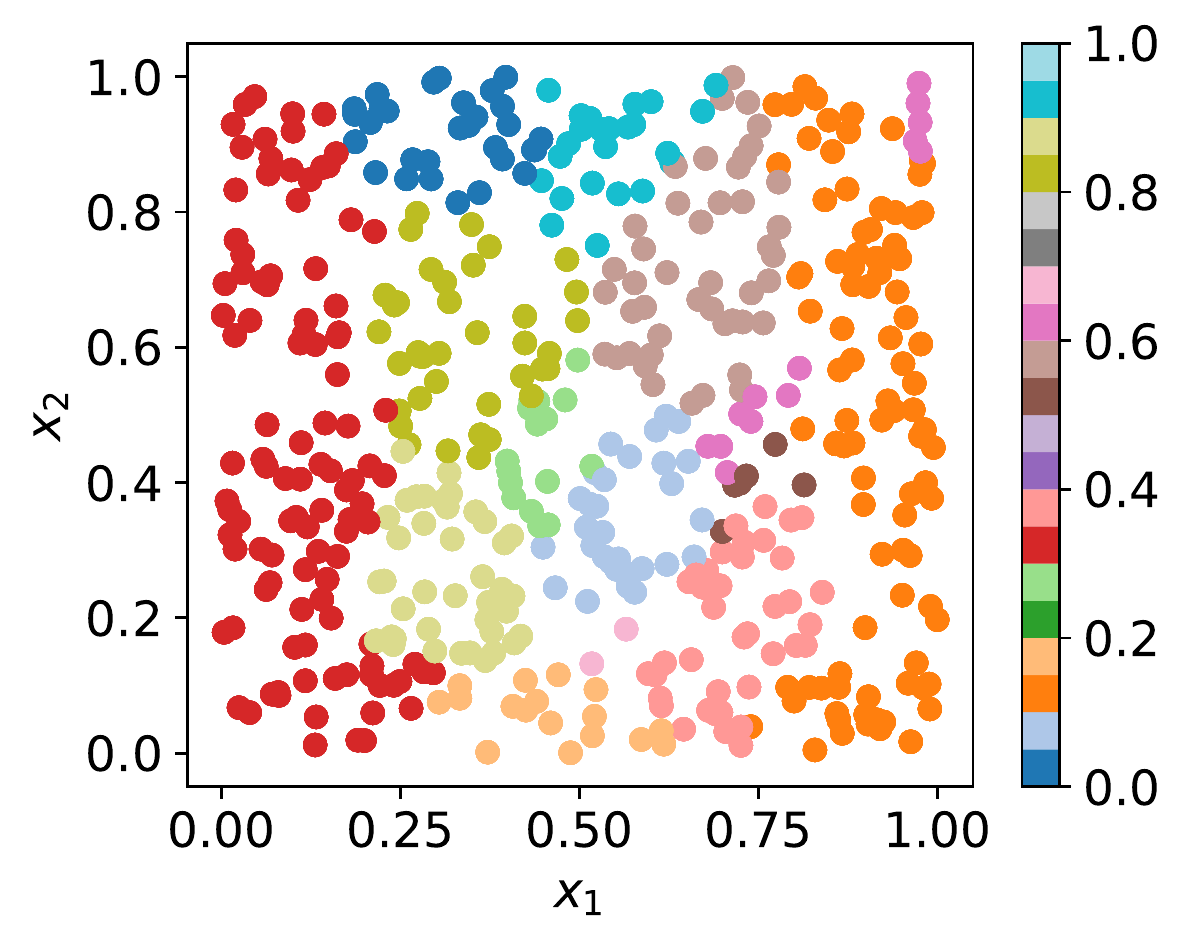}
    \end{minipage}
    \begin{minipage}[c]{0.32\textwidth}
    \centering
    PPOU-Net standard deviation \\ \vspace{0.2cm}
    \includegraphics[width=\columnwidth]{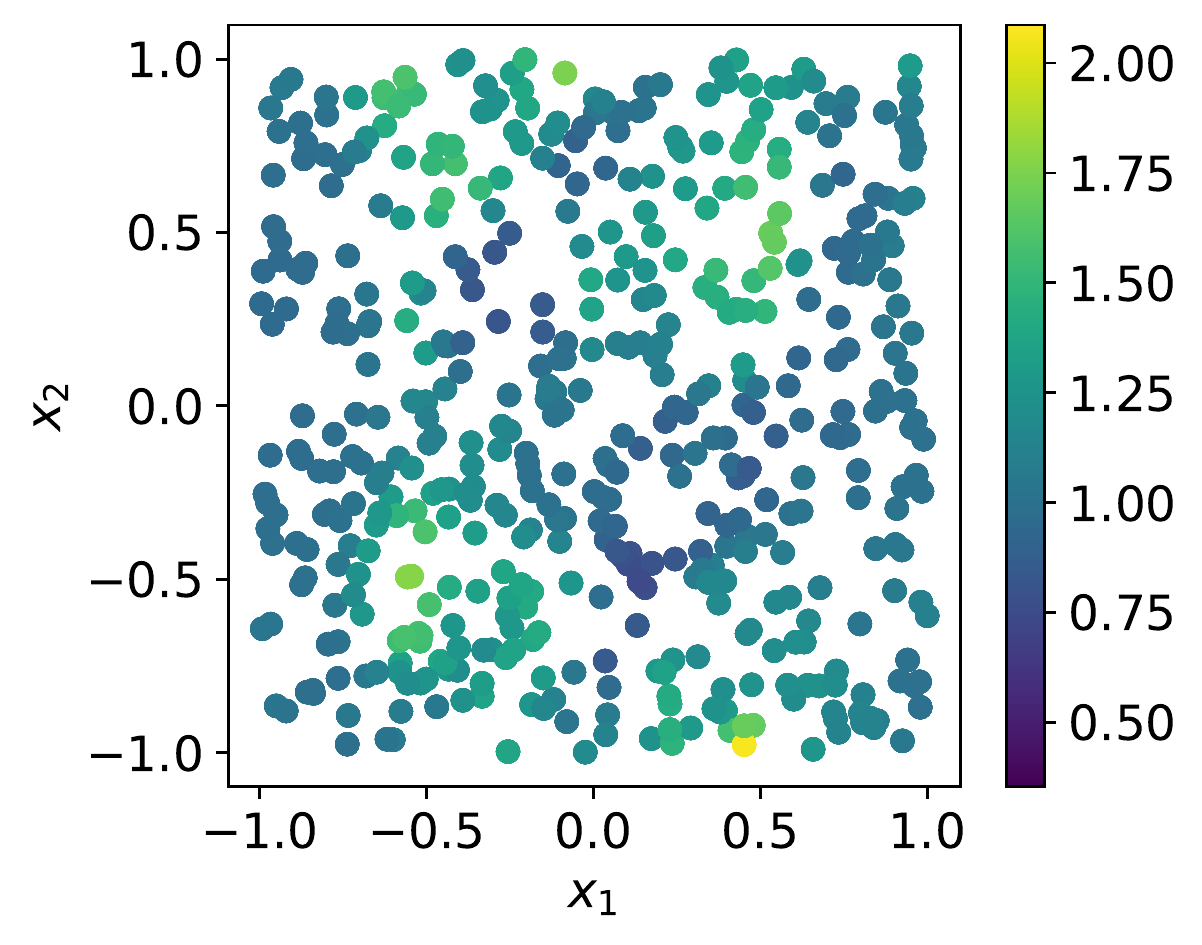}
    \end{minipage}
    \begin{minipage}[c]{0.32\textwidth}
    \centering
    empirical standard deviation \\ \vspace{0.2cm}
    \includegraphics[width=\columnwidth]{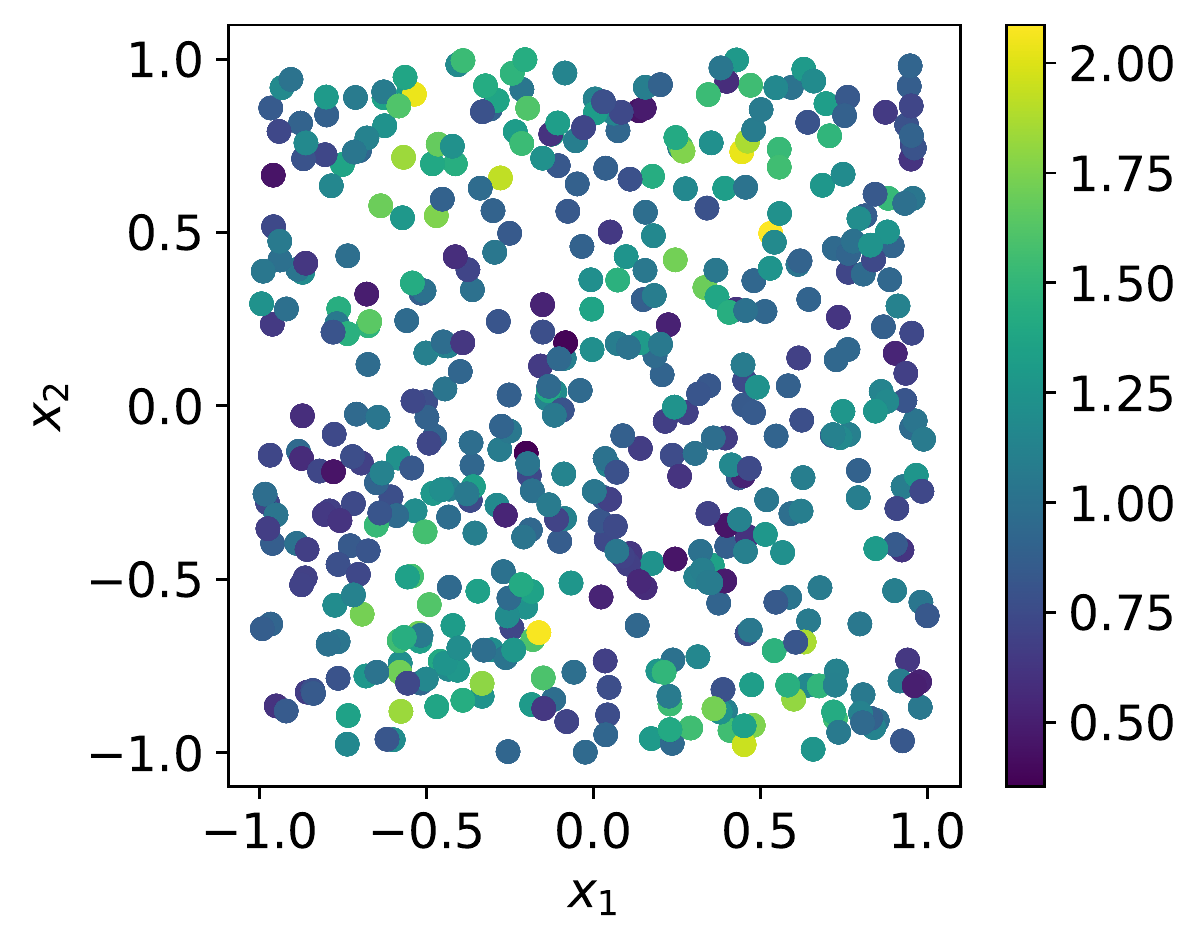}
    \end{minipage}
    \caption{Example of a PPOU-Net trained to the 2D cost function subject to experimental noise. The top left plot illustrates the ideal cost function (without noise). We visualize the dominating partitions in the PPOU-Net model (bottom left), and compare both the predicted mean (top center) and the predicted standard deviation (top right) to the empirical values (bottom center and bottom right), which we compute from the raw data. The PPOU-Net model partitions the complex data domain into regions where quadratic functions can accurately approximate the target function on each partition. Both the mean and the standard deviation predicted by the PPOU-Net closely match the empirical values of the noisy data. \label{fig:quantum_example}}
\end{figure}

The cost function can be measured from the quantum circuit as a sample mean, where the sample size is specified by the $N_{\text{shots}}$ parameter. A larger $N_{\text{shots}}$ corresponds to a lower uncertainty in the measurement. In the following experiments, the cost functions are generated for an NP-hard graph max-cut problem. As an example, the cost function (without machine error) of input dimension $2$ is plotted in \Cref{fig:quantum_example} (top left).

\begin{figure}[htbp]
    \centering
    \begin{minipage}[c]{0.36\textwidth}
    \centering
    ~~~$d=2$ \\
    \includegraphics[width=1\columnwidth]{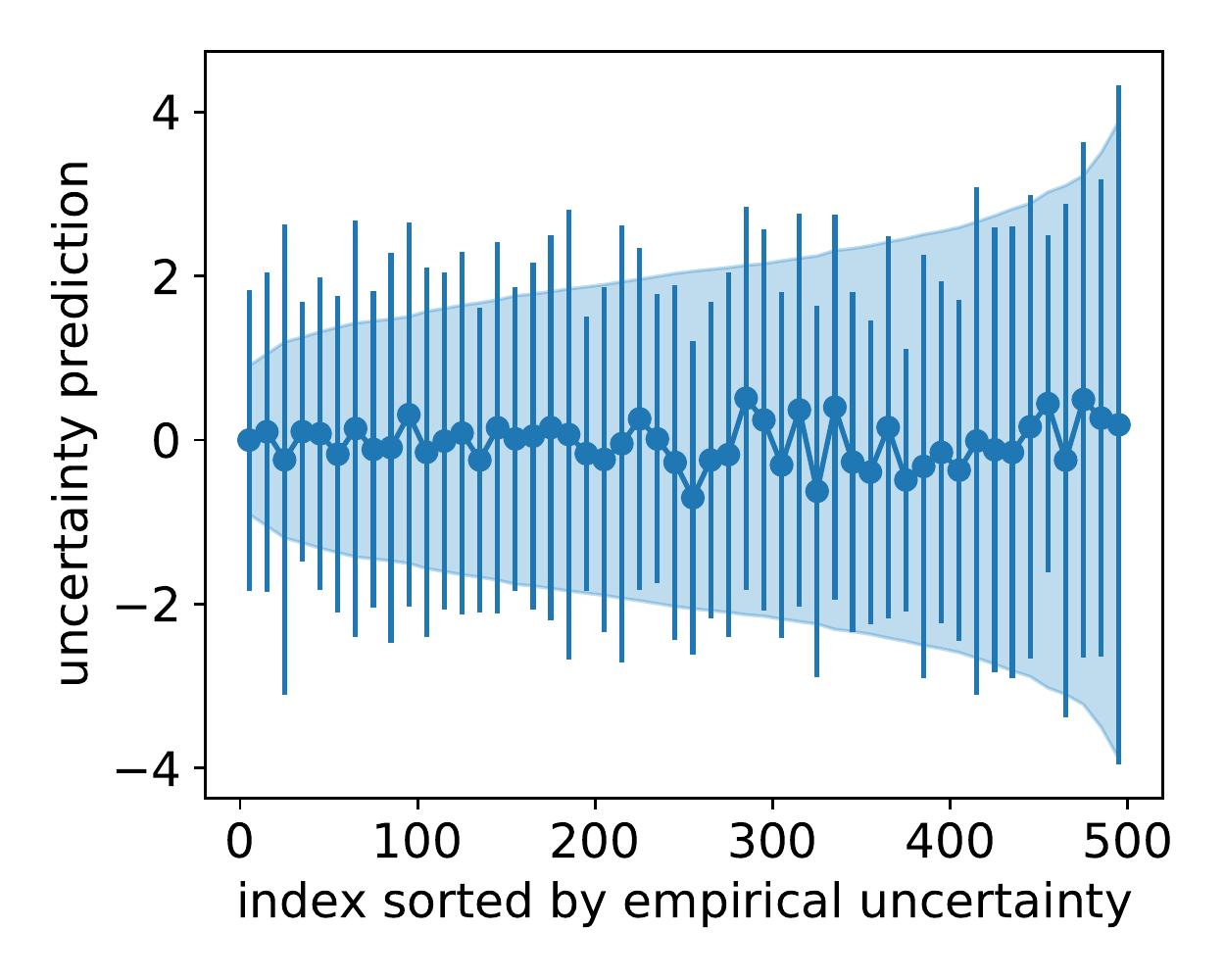}
    \end{minipage}
    \begin{minipage}[c]{0.585\textwidth}
    $d=4$ ~~~~~~~~~~\\
    \centering
    ~~~~~~\includegraphics[width=1\columnwidth]{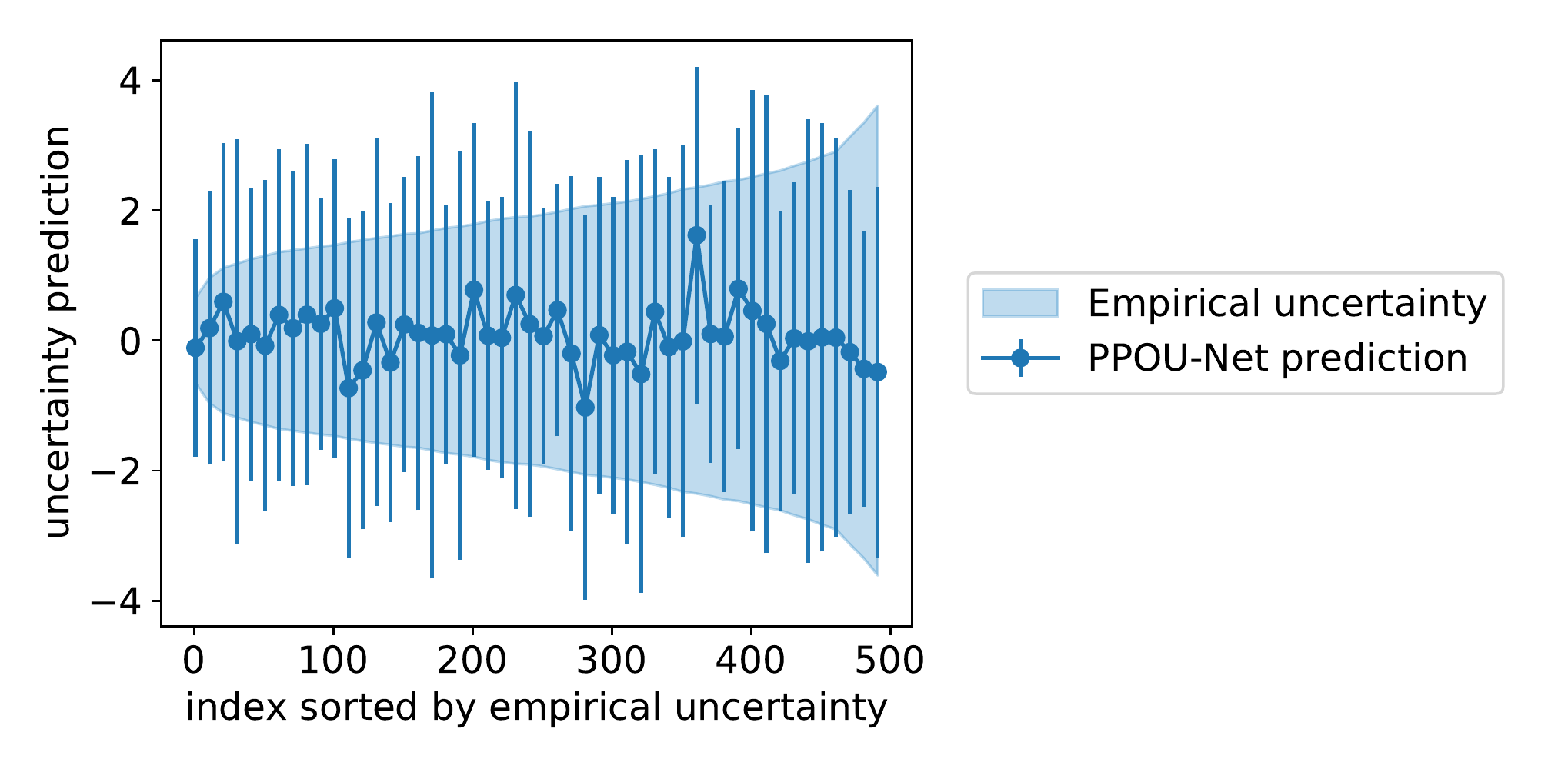}
    \end{minipage}\\
    \begin{minipage}[c]{0.36\textwidth}
    \centering
    ~~$d=8$ \\ \vspace{0.05cm}
    \includegraphics[width=1\columnwidth]{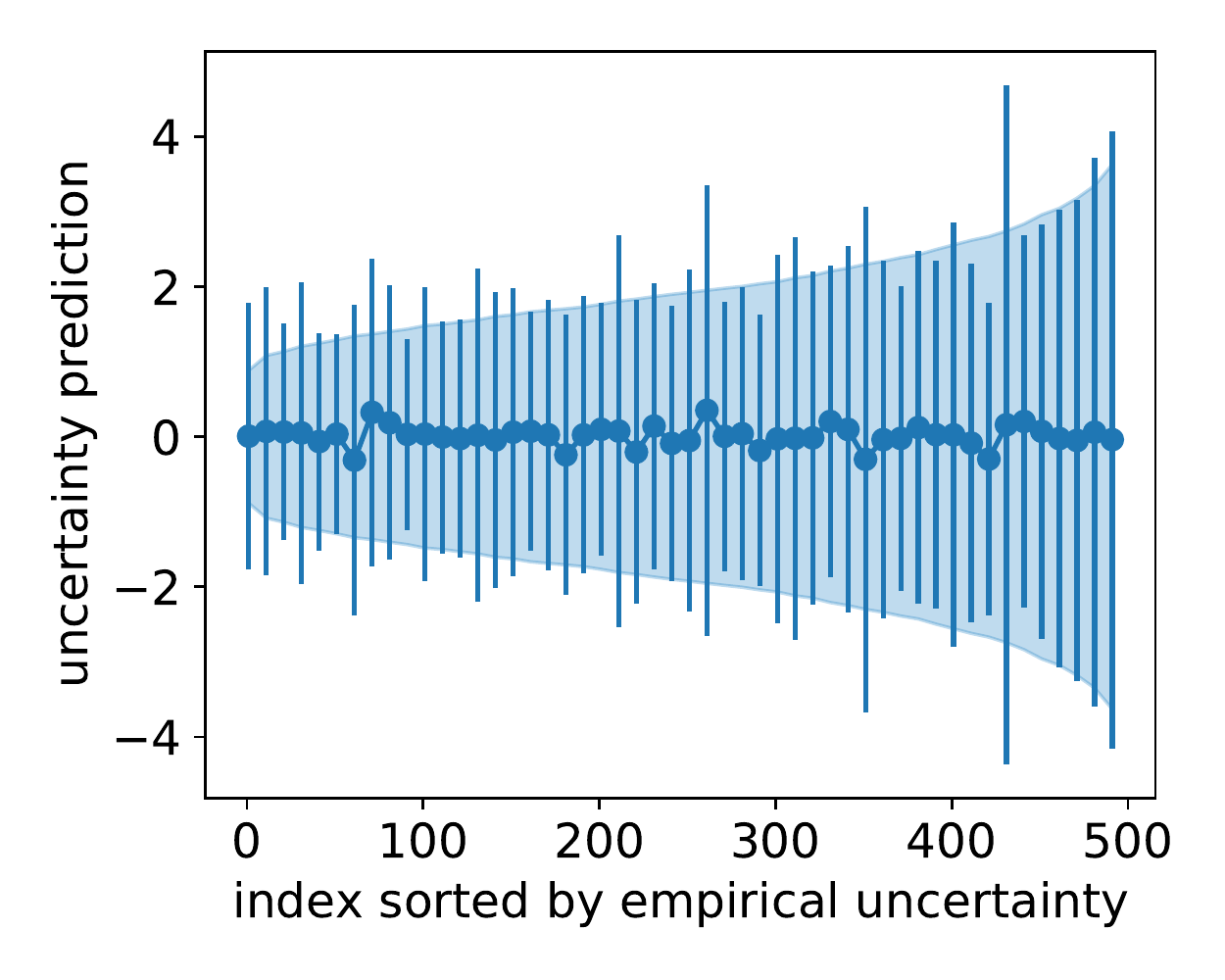}
    \end{minipage}
    \begin{minipage}[c]{0.36\textwidth}
    \centering
    ~~$d=16$ \\ \vspace{0.05cm}
    \includegraphics[width=1\columnwidth]{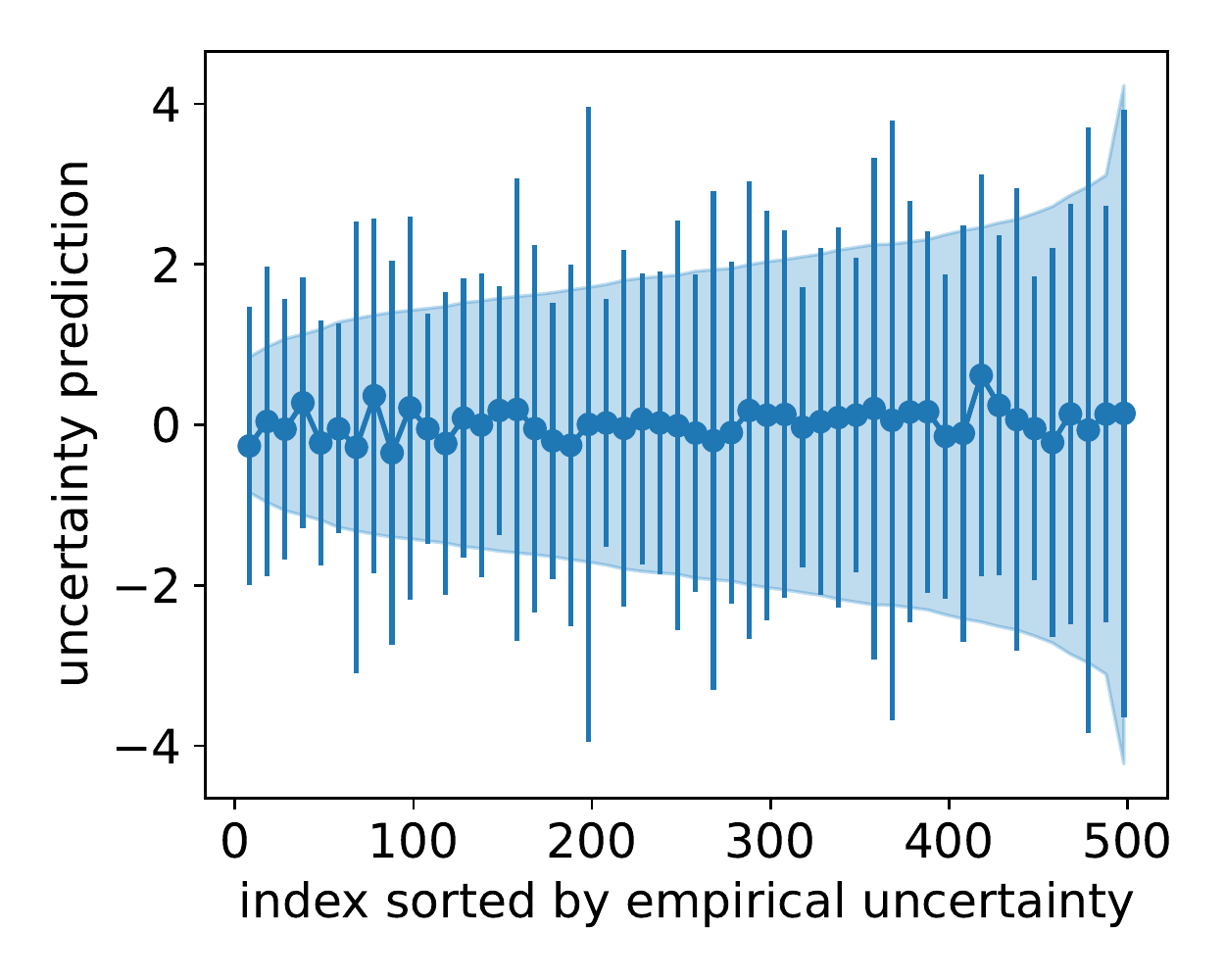}
    \end{minipage}
    \hspace{3.9cm}
    \caption{A quantitative evaluation of the uncertainty prediction by a PPOU-Net, trained to cost functions of various data dimensions. For each plot, we sample $10\%$ of the input-output pairs and sort them in the order of increasing empirical uncertainty. The violin plots span a 95\% confidence interval computed from the raw training data. The error bars span a 95\% confidence interval computed from the predicted standard deviation and are centered at $\mathbb{E}[Y(\mathbf{x})]- (1/|\mathcal{I}(\mathbf{x})| ) \sum_{s\in \mathcal{I}(\mathbf{x})} y^{(s)}$, where $\mathcal{I}(\mathbf{x})$ denotes the set of indices of the multiple experimental outputs for input $\mathbf{x}$. The PPOU-Net predictions for the machine error match consistently with the empirical uncertainty region of the raw data for all data dimensions. }
    \label{fig:quantum_violin}
\end{figure}

In the first task, we use a PPOU-Net to quantify the uncertainty in the experimental data. For data dimensions $d\in \{ 2, 4, 8, 16 
\}$, 
we sample $500$ uniformly random \added{spatial coordinates} from $[-1, 1]^d$. For each input value, we \added{sample} the quantum circuit 10 times under the precision level $N_{\text{shots}}=10$. We use the $N=5000$ \added{samples} in total as the training set.  \added{We use PPOU-Nets to learn the mapping from the spatial coordinates $\mathbf{x}\in \mathbb{R}^d$ to the noisy measurements of the quantum circuit $y = \tilde f_d ( \mathbf{x} ) \in \mathbb{R}$, where $\tilde f_d$ is the sampling function of the quantum circuit of input data dimension $d$}. The PPOU-Net includes an explicit model for background noise as described in \Cref{subsec:background noise} and has depth 6, width 12, and 16 partitions with quadratic functions. As illustrated in \Cref{fig:quantum_example}, the PPOU-Net predictions for both the mean and the standard deviation match closely with the empirical values.  We obtain a 95\% confidence region for the PPOU-Net prediction in each experiment and compare that to the empirical confidence intervals in \Cref{fig:quantum_violin}. For all data dimensions, the PPOU-Net prediction lies consistently in the empirical uncertainty region computed from the raw data, and the sizes of the confidence intervals predicted by PPOU-Nets are consistent with the empirical confidence intervals.

Next, we use a PPOU-Net to approximate the cost functions of dimensions between 6 and 32, by encoding a uniformly random sample in high dimensions to a 4D latent space. For dimensions $d\in \{6, 8, \ldots, 32\}$, we sample $N=10000$ spatial coordinates from a random 4D subspace of side length $0.5$ from $[0, 1]^d$ as the training set inputs; we independently sample $N_{\text{test}}=10000$ spatial coordinates from the same subspace as the testing set inputs.  We use a PPOU-Net consisting of an encoder (with depth 3, width 32, and 4 latent dimensions), a classifier (with depth 10, width 8) and 32 partitions with cubic functions. \added{The PPOU-Net is trained with the spatial coordinates $\mathbf{x}\in \mathbb{R}^d$ as input data and the  cost function value $f_d(\mathbf{x}) \in \mathbb{R}$ as output data, where $f_d$ is the ideal cost function of input data dimension $d$ without measurement error.} In order to efficiently populate the large number of partitions, we pre-train the PPOU-Nets to match the K-means classification results with 32 clusters, before applying the EM-motivated training strategy. 

\begin{figure}[tpbh!]
\centering
\begin{minipage}[c]{0.7\textwidth}
    \centering
    \includegraphics[width=\columnwidth]{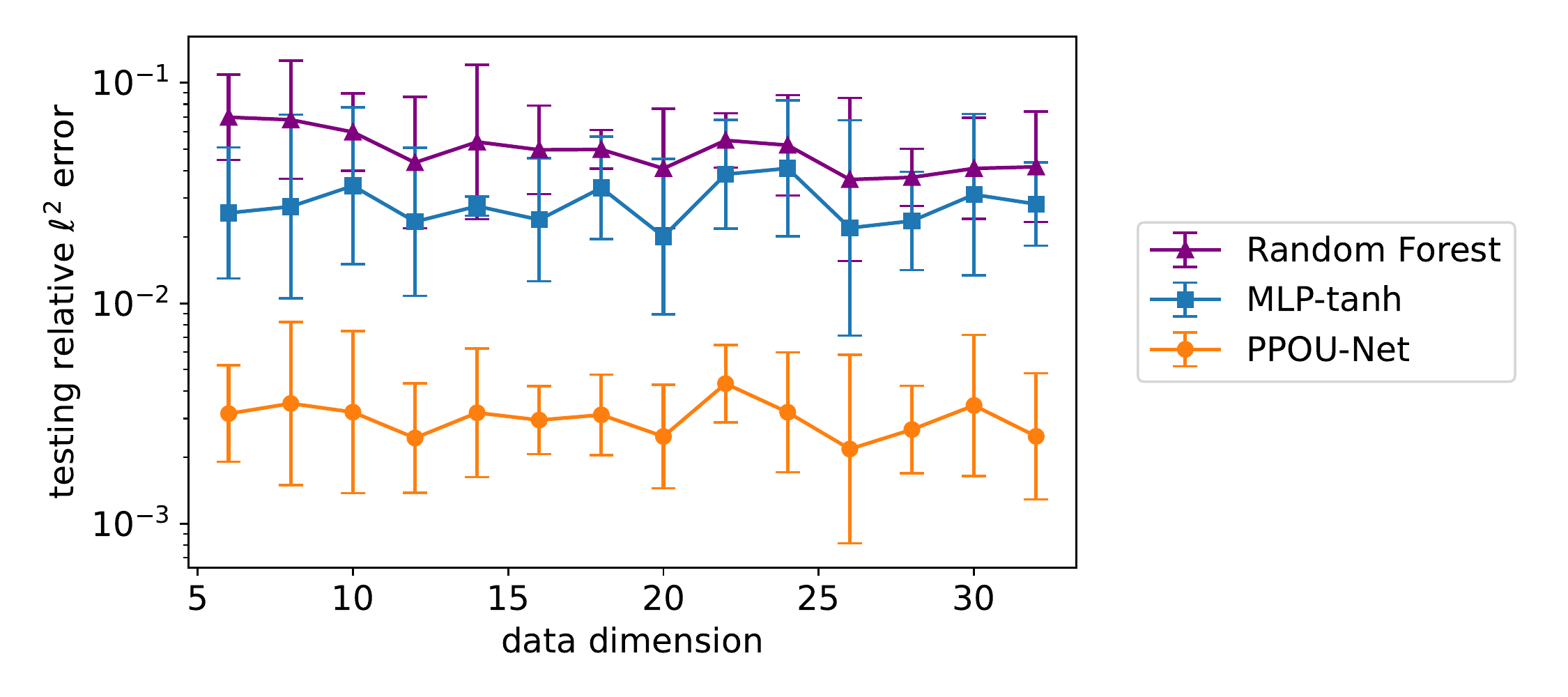}
\end{minipage}
\caption{\label{fig:quantum_encode_convergence} A comparison of relative $\ell^2$ errors for target functions of various data dimensions by PPOU-Nets, \added{Random Forest regressors, and tanh-based MLPs}. The error bars cover a 95\% confidence interval of the testing errors. The PPOU-Nets consistently achieve an order of magnitude improvement in accuracy over \added{the two baseline methods} at all dimensions. The performance of PPOU-Nets does not degrade as the data dimension increases.}
\end{figure}

We compare the PPOU-Nets \added{to random forest regression models and} tanh-based MLP models of matching degrees of freedom. \added{We repeat the experiments with different neural network initializations and random ensembles of data.} As shown in \Cref{fig:quantum_encode_convergence}, \added{the fact that the 95\% confidence intervals of the PPOU-Net and of the two baseline methods do not overlap suggests that the PPOU-Net outperforms the baseline methods with statistical significance at the level of $\alpha=0.05$.} The PPOU-Nets lead to a consistent improvement of model accuracy by an order of magnitude, and its robust performance does not deteriorate with the increase of ambient dimensions.

\section{Conclusion}\label{sec:conclusion}

The PPOU-Net is a general probabilistic framework for high-dimensional regression problems that naturally provides uncertainty quantification. \added{Unlike many existing sequential machine learning pipelines for high-dimensional scientific data, the proposed PPOU-Net performs adaptive dimension reduction such that the latent space is designed with the information of the target output. At each step of training, we obtain the optimal least squares fit of data for a given set of partitions. Therefore the partitions evolve along a manifold of optimal representations of data, contributing to significant improvement in accuracy.} The PPOU-Net framework provides accurate point estimations and confidence regions; performs efficient dimensionality reduction of data domains; and achieves robust convergence across various examples and applications, including noisy observations from smooth and non-smooth functions. The model functions as a low-dimensional surrogate where the gradients can be easily accessed via automatic differentiation. We show that PPOU-Nets achieve more robust approximations compared to baseline classical and machine learning regression methods for a range of numerical experiments.

As the expectation-maximization training strategy allows for the reduction of a global least-squares problem into a number of local least-squares solves, we will explore parallel and distributed training of the model in the future. In addition, we will use ideas from Bayesian statistics to further improve the model robustness in the limit of small data size.

\added{We conclude this manuscript by stressing the fact that the objective of Gaussian mixture models is to approximate more complex distributions using a piecewise Gaussian decomposition of the space. In previous work \cite{trask2022hierarchical} we have explored settings under which a piecewise decomposition of the space approximates well heteroskedastic unimodal distributions. While this approximation becomes more accurate with increasing numbers of partitions, one may overfit in the limit of very large numbers. For datasets following binomial or multimodal distributions, the Gaussian distribution would also be restrictive. Thus, in general, one may adopt other distributions in this framework, provided they lead to a computationally tractable ELBO and their parameters admit closed-form expressions in the expectation-maximization algorithm.}


\section*{Acknowledgments}

The authors would like to thank Dr. Mohan Sarovar and Ryan Shaffer at Sandia National Laboratories for introducing us to the application in quantum computing and generously providing the simulation code for the quantum circuit experiments. We acknowledge the PhILMs Center (PhILMs: Collaboratory on Mathematics and Physics Informed Learning Machines for Multiscale and Multiphysics, Department of Energy, no.\ DE-SC0019453) for funding support. N. Trask also acknowledges funding from the U.S. Department of Energy, Office of Advanced Scientific Computing Research Early Career Research Program. 


Sandia National Laboratories is a multimission laboratory managed and operated by National Technology \& Engineering Solutions of Sandia, LLC, a wholly owned subsidiary of Honeywell International Inc., for the U.S. Department of Energy’s National Nuclear Security Administration under contract DE-NA0003525. This paper describes objective technical results and analysis. Any subjective views or opinions that might be expressed in the paper do not necessarily represent the views of the U.S. Department of Energy or the United States Government. This article has been co-authored by an employee of National Technology \& Engineering Solutions of Sandia, LLC under Contract No. DE-NA0003525 with the U.S. Department of Energy (DOE). The employee owns all right, title and interest in and to the article and is solely responsible for its contents. The United States Government retains and the publisher, by accepting the article for publication, acknowledges that the United States Government retains a non-exclusive, paid-up, irrevocable, world-wide license to publish or reproduce the published form of this article or allow others to do so, for United States Government purposes. The DOE will provide public access to these results of federally sponsored research in accordance with the DOE Public Access Plan https://www.energy.gov/downloads/doe-public-access-plan.

SAND Number: SAND2022-12013 O

\subsection*{Author contributions}
\textbf{Tiffany Fan}: Conceptualization (equal); writing – original draft (lead); methodology (equal); software (lead); writing – review and editing (equal). \textbf{Nathaniel Trask}: Conceptualization (equal);  methodology (equal); software (supporting); writing – review and editing (equal). \textbf{Marta D'Elia}: Conceptualization (equal);  methodology (equal); writing – review and editing (equal). \textbf{Eric Darve}: Conceptualization (equal);  methodology (equal); writing – original draft (supporting); writing – review and editing (equal).

\subsection*{Financial disclosure}

None reported.

\subsection*{Conflict of interest}

The authors declare no potential conflict of interests.

\appendix

\section{derivation of the basic PPOU-Net training strategy}\label{sec:basic_ppou_derivation}

We consider a general regression problem with $N$ independent training examples $\{\mathbf{x}^{(n)}, y^{(n)} \}_{n=1}^N$, where $\mathbf{x}^{(n)} \in \Omega \subset \mathbb{R}^d$ and  $ y^{(n)}\in \mathbb{R}$. We choose a set of basis polynomials $\{p_k(\mathbf{x})\}_{k=1}^K$ that are supported on $\Omega$ and span a Banach space $V$. Under the model assumptions in \Cref{eq:assumptions1},
and leveraging the ideas of EM \cite{dempster1977maximum,blei2017variational}, Trask et al.\ \cite{trask2021probabilistic} proposed the following training strategy for PPOU-Nets. 

The log-likelihood of the data $\{\mathbf{x}^{(n)}, y^{(n)}\}_{n=1}^N$ can be expressed in terms of the latent variable model
\begin{equation} \label{eq:log_likelihood}
\log p( \{ \mathbf{x}^{(n)}, y^{(n)} \}_n ; \theta, c, \sigma^2) = \sum_{n} \log p( \mathbf{x}^{(n)}, y^{(n)} ; \theta, c, \sigma^2)
=  \sum_{n} \log  \sum_{z^{(n)}} p( \mathbf{x}^{(n)}, y^{(n)}, z^{(n)} ; \theta, c, \sigma^2) .
\end{equation}
For any valid distributions $ \{ w^{(n)} (z^{(n)}) \}_{n=1}^N$ where we denote the positive probability mass values $w_j^{(n)}:= p_{w^{(n)}} (z^{(n)}=j) > 0$, by Jensen's Inequality,  the log-likelihood has an evidence lower bound (ELBO)
\begin{equation}
\begin{aligned}
 \eqref{eq:log_likelihood} \geq \, &  \ell_{\text{ELBO}} 
 (  \theta, c, \sigma^2; \{\mathbf{x}^{(n)}, y^{(n)}, w^{(n)} \}_n) \\ &=   \sum_{n} \sum_{z^{(n)}} w^{(n)}(z^{(n)})   \log \frac{ p( \mathbf{x}^{(n)}, y^{(n)}, z^{(n)} ; \theta, c, \sigma^2) }{  w^{(n)}(z^{(n)})  }\\
 &= \sum_{n} \sum_{j} w^{(n)}_j   \log  p( \mathbf{x}^{(n)}, y^{(n)}, z^{(n)} ; \theta, c, \sigma^2)  + C \\
 &= \sum_{n} \sum_{j} w^{(n)}_j  \Big( \log \phi_j( \mathbf{x}^{(n)};\theta ) + \log \mathcal{N} \big( y^{(n)} \mid \mu_j ( \mathbf{x}^{(n)}; c) , \sigma^2_j \big)\Big) +C ,
 \label{eq:elbo}
 \end{aligned}
\end{equation}
and the equality holds if and only if
$ w^{(n)} (z^{(n)}) = p(z^{(n)} 
\mid \mathbf{x}^{(n)}, y^{(n)} ; \theta, c, \sigma^2) $ for all $n$.

The training strategy alternates between E-step and M-step. 
\subparagraph{E-step.} The goal of the E-step is to obtain a tight lower bound (i.e., the ELBO) for the local log-probability while fixing $\{\theta, c, \sigma^2\}$. From \eqref{eq:elbo}, this is achieved by setting $w^{(n)}(z^{(n)}) = p(z^{(n)} 
\mid \mathbf{x}^{(n)}, y^{(n)} ; \theta, c, \sigma^2)$.

\subparagraph{M-step.} 
The M-step maximizes the ELBO with respect to $\{ \theta, c, \sigma^2\}$ while fixing the distributions $\{ w^{(n)} \}_n$. We leverage the LSGD algorithm \cite{cyr2020robust} in the M-step. Specifically, we perform coordinate ascent of the ELBO in $c$, $\sigma^2$, and $\theta$ sequentially, where the updates of $\{c, \sigma^2\}$ are computed in analytical forms and the update of $\theta$ is computed using numerical gradient ascent.

Taking the partial derivative of the ELBO \eqref{eq:elbo} with respect to $c_{j,k}$ and setting it to zero, 
\begin{align*}
    \frac{\partial \ell_{\text{ELBO}}}{\partial c_{j,k}}  =  \sum_{n }  w_j^{(n)} \frac{y^{(n)} - \mu_j(x^{(n)})}{  \sigma_j^2} \  p_k(x^{(n)})  = \sum_{n } \frac{w_j^{(n)}}{\sigma_j^2}
    \Big( y^{(n)} p_k(x^{(n)})
    - \sum_{l} c_{j, l} p_l(x^{(n)})p_k(x^{(n)}) \Big) =0
\end{align*}
we derive that $\mathbf{c}_j= \{ c_{j,k}\}_ {k}$ satisfies the normal equation 
$$ P_j^\top W_j P_j \mathbf{c}_j = P_j^\top W_j\, \mathbf{y}, $$  where $P_j$ has entries $[P_j]_{n, k} = p_k( \mathbf{x}^{(n)} ) ,  W_j =\operatorname{diag}( \{w^{(n)}_{j} \}_{n=1}^N  )$, and $\mathbf{y}=\{y^{{(n)}} \}_{n=1}^N$. Hence $\mathbf{c}_j$ is the optimal coefficients to the weighted least squares problem
\begin{align*}
    \mathbf{c}_j = \text{argmin}_{\{c_{j, l}\}} \sum_{n} w_j^{(n)} \Big( y^{(n)} - \sum_{l} c_{j,l} p_l(x^{(n)}) \Big)^2.
\end{align*}
Maximizing the ELBO \eqref{eq:elbo} with respect to $\sigma^2_j$ results in a closed-form expression similar to Gaussian mixture models
$$
    \sigma_j^2 = \frac{\sum_{n=1}^N w_j^{(n)} \left( y^{(n)} - \mu_j   \left(  \mathbf{x}^{(n)} \right)  \right)^2 
    }{\sum_{n=1}^N w_j^{(n)}}.
$$
Next, we update the weights and biases $\theta$ using gradient ascent with respect to the ELBO \eqref{eq:elbo}, or equivalently, gradient descent with respect to the loss
\begin{align*}
        L(\theta) = - \sum_{n} \sum_{j} w_j^{(n)} \log \phi_j(\mathbf{x}^{(n)}; \theta) 
\end{align*} or the alternative loss
\begin{align*}
        \tilde{L}(\theta) = - \sum_{n} \sum_{j} w_j^{(n)} \log \phi_j(\mathbf{x}^{(n)}; \theta) +   \sum_{n}   \Big(  y^{(n)} - \sum_{j} \phi_j(\mathbf{x}^{(n)}; \theta)  \sum_{k} c_{j, k} p_k(\mathbf{x}^{(n)}) \Big) ^2.
\end{align*} The motivation of the alternative loss follows from the discussions in \Cref{sec:methods}.

We summarize the training process in \Cref{alg:em}.
\begin{algorithm}[tbh!]
\caption{EM-inspired training loop for basic PPOU-Nets}\label{alg:em}
\begin{algorithmic}
\State Repeat until convergence \{ \\
\begin{enumerate}
    \item (E-step) For $n=1,\ldots,N$ and $j=1,\ldots,J$, compute
    \begin{align*}
        w_j^{(n)} \gets &\   \frac{1}{\sigma_j} \exp \Big( -\frac{ \big(y^{(n)} - \mu_j(\mathbf{x}^{(n)} ) \big)^2 }{2\sigma_j^2} \Big)  \phi_j(\mathbf{x}^{(n)};\theta)\\
        w_j^{(n)} \gets & \  \frac{w_j^{(n)}}{\sum_{l=1}^J w_l^{(n)} } 
    \end{align*}
    \item (M-step) Update the DNN parameters $\theta$ using gradient descent, under the loss function
    \begin{equation*}
        L(\theta) = - \sum_{n=1}^N \sum_{j=1}^J w_j^{(n)} \log \phi_j(\mathbf{x}^{(n)}; \theta) +   \sum_{n=1}^{N}   \Big(  y^{(n)} - \sum_{j=1}^J \phi_j(\mathbf{x}^{(n)}; \theta)\, \mu_j(\mathbf{x}^{(n)} )  \Big) ^2
\end{equation*}
    For $j=1,\ldots,J$, solve the weighted least-squares problem
    \begin{align*}
        \mu_j \gets \operatorname{argmin}_{g \in \text{span} \{p_1, \ldots, p_K \}} \sum_{n=1}^N w_j^{(n)} \big( y^{(n)} - g (\mathbf{x}^{(n)})   \big)^2
    \end{align*}

    and update the variance term \begin{align*}
    \sigma_j^2 \gets \  \frac{\sum_{n=1}^N w_j^{(n)} \big( y^{(n)} - \mu_j( \mathbf{x}^{(n)})  \big)^2 
    }{\sum_{n=1}^N w_j^{(n)}}
    \end{align*}
    
\end{enumerate}
\State \}
\end{algorithmic}
\end{algorithm}

\section{cross-validation experiments}\label{sec:cross_val}

\begin{figure}[tbh!]
    \centering
    \includegraphics[width=0.52\columnwidth]{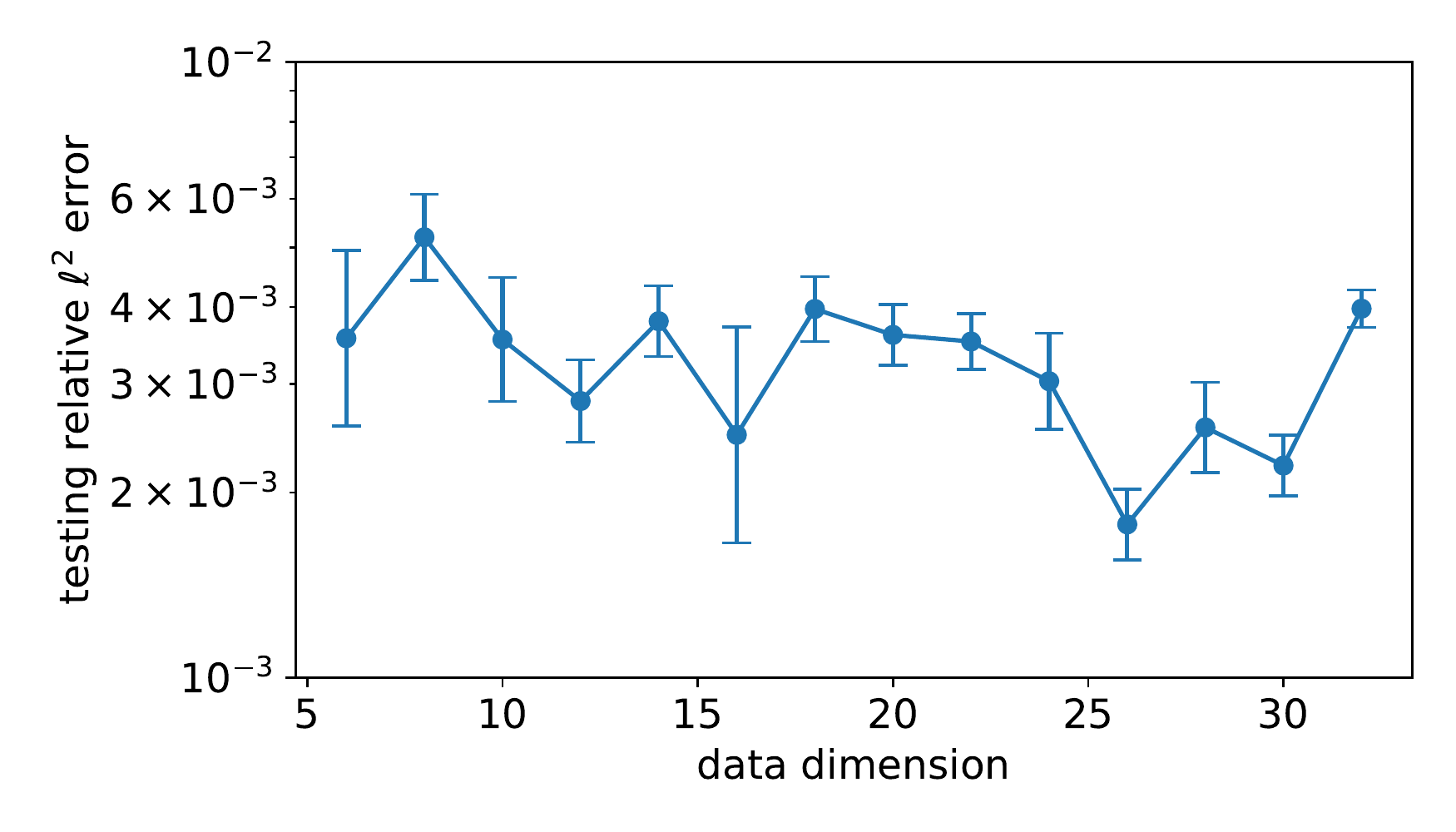}
    \caption{\label{fig:qaoa_cross_val} \added{The testing $\ell^2$ relative errors achieved by PPOU-Nets in $k$-fold cross-validation experiments. The error bars indicate the average and the 95\% confidence interval of the testing errors. The PPOU-Net exhibits consistent performance in the cross-validation experiments. }  }
\end{figure}

\added{In this section, we report cross-validation experimental results for the second task of \Cref{subsec:numerical_quantum}. With a fixed dataset of size $20000$ for each input dimension of the quantum circuit, we conduct $k$-fold cross-validation experiments for PPOU-Nets with $k=4$. The PPOU-Nets are built with the same model architecture as in \Cref{subsec:numerical_quantum}, consisting of an encoder (with depth 3, width 32, and 4 latent dimensions), a classifier (with depth 10, width 8) and 32 partitions with cubic functions. As we illustrate in \Cref{fig:qaoa_cross_val}, PPOU-Nets exhibit consistent performance in the cross-validation experiments and achieve testing errors of the same order of magnitude as observed in \Cref{subsec:numerical_quantum}. }

\bibliography{main}
\end{document}